%% file: neurips_2026.tex
\documentclass{article}

     \PassOptionsToPackage{numbers, compress}{natbib}

 \usepackage[main, final]{neurips_2026_ArXiv}

\usepackage[utf8]{inputenc} 
\usepackage[T1]{fontenc}    
\usepackage{hyperref}       
\usepackage{url}            
\usepackage{booktabs}       
\usepackage{amsfonts}       
\usepackage{nicefrac}       
\usepackage{microtype}      
\usepackage{xcolor}         

\usepackage{amsmath}
\usepackage{amssymb}
\usepackage{mathtools}
\usepackage{amsthm}

\usepackage[capitalize,noabbrev]{cleveref}

\theoremstyle{plain}
\newtheorem{theorem}{Theorem}[section]
\newtheorem{proposition}[theorem]{Proposition}
\newtheorem{lemma}[theorem]{Lemma}

\theoremstyle{definition}
\newtheorem{definition}[theorem]{Definition}

\theoremstyle{remark}
\newtheorem{remark}[theorem]{Remark}

\usepackage[textsize=tiny]{todonotes}

\usepackage{subcaption}
\usepackage{epsf}
\usepackage{fancyhdr}
\usepackage{graphics}
\usepackage{psfrag}

\usepackage{wrapfig}
\usepackage{float}

\usepackage{algorithmic}

\usepackage{color}

\usepackage{bm,bbm}
\usepackage{multirow}
\usepackage{balance}

\def\calGS{\mathcal{GS}}

\def\calGST{\mathcal{GST}}
\def\calST{\mathcal{ST}}
\def\calOW{\mathcal{OW}}

\def\OrliczSobolevPhi{W\!L_{\Phi}^1}
\def\OrliczSobolevPhiZero{W\!L_{\Phi, 0}^1}

\def\OrliczSobolevPsi{W\!L_{\Psi}^1}
\def\OrliczSobolevPsiZero{W\!L_{\Psi, 0}^1}

\usepackage{verbatim}

\input{comments}

\title{Orlicz–Sobolev with Musielak: An Efficient Regularization Approach for Graph-based IPM}

\author{%
  Tam Le\thanks{: equal contribution, $^\dagger$: co-last author. Correspondence to: Tam Le <\texttt{tam@ism.ac.jp}>.} \\
  The Institute of Statistical Mathematics \\
   \& Tohoku University \\
  \texttt{tam@ism.ac.jp} \\
  \And
  Truyen Nguyen$^*$ \\
  The University of Akron \\
  \texttt{truyennguyen3@gmail.com} \\ 
  \And
  Hideitsu Hino$^\dagger$ \\
  The Institute of Statistical Mathematics \\
  \texttt{hino@ism.ac.jp} \\
  \And
  Kenji Fukumizu$^\dagger$ \\
  The Institute of Statistical Mathematics \\
  \texttt{fukumizu@ism.ac.jp} \\
}

\begin{document}

\maketitle

\begin{abstract}

We study the Sobolev IPM problem for probability measures supported on a graph metric space, where critic function is constrained to lie within the unit ball defined by Sobolev norm. Sobolev IPM is intrinsically coupled with $L^p$ geometric structure within its definition, limiting its ability to incorporate other prior geometry beyond the $L^p$ paradigm. Conversely, the classic optimal transport is flexible, easy to adapt to various geometric structures by simply changing its ground cost. An important example is Orlicz-Wasserstein (OW) which utilizes \emph{Orlicz} geometric structure to generalize the $L^p$ within the standard $p$-order Wasserstein, and remarkably play a vital role to advance machine learning methodologies. Inspired by recent advantages of OW, in this work, we leverage a specific class of convex functions for Orlicz geometry to mitigate such limitation for Sobolev IPM, and propose the generalized Sobolev IPM (GSI). Our GSI approach encompasses Sobolev IPM as a special case while accommodating diverse geometric priors beyond $L^p$. It however brings up significant computational hurdles that compound those already notoriously inherent in Sobolev IPM. To address these challenges, we theoretically establish a novel connection between \emph{Orlicz-Sobolev} norm and \emph{Musielak} norm which facilitates a novel efficient regularization for GSI. By further exploiting the underlying graph structure, we show that the regularized GSI reduces to a simple univariate optimization problem, achieving notably computational efficiency, enabling its usage in practical applications. We empirically illustrate that the regularized GSI is several-order faster than the popular OW in computation, and its performances compare favorably to other transport baselines for comparing graph-based measures in document classification and topological data analysis.

\end{abstract}

\section{Introduction}\label{sec:introduction}

Probability measures serve as canonical mathematical representations for diverse objects of interest across various research fields, e.g., documents in natural language processing~\citep{kusner2015word, yurochkin2019hierarchical}, persistence diagrams in topological data analysis (TDA)~\citep{edelsbrunner2008persistent, le2025scalable}, point clouds in computer vision and graphics~\citep{hua2018pointwise, wang2019dynamic, nguyen2021point, choromanski2023efficient}. To compare such measures, integral probability metrics (IPM) offer a versatile and principled class of metric functions ~\citep{muller1997integral, Sriperumbudur2012OnTE}. Conceptually, IPM operate by determining an optimal critic function that achieves maximal discrimination between probability measures. This mathematical elegance and versatility has facilitated the widespread adoption of IPM throughout statistics and machine learning~\citep{sriperumbudur2009integral, gretton2012kernel, peyre2019computational, liang2019estimating, uppal2019nonparametric, uppal2020robust, nadjahi2020statistical, Kolouri2020Sliced, le2025scalable, tran2026treesliced}.


In this work, we study the Sobolev IPM problem for measures supported on a graph metric space, where critic function is constrained within the unit ball induced by Sobolev norm~\citep{adams2003sobolev, liang2017well_arXiv, liang2021well_jmlr, singh2018nonparametric, le2025scalable, tran2026treesliced}. Sobolev IPM has proven fundamental to numerous theoretical analyses, including convergence rates in density estimation and approximation theory for deep architectures~\citep{liang2017well_arXiv, liang2021well_jmlr, singh2018nonparametric}. Additionally, by exploiting the underlying graph structure,~\citet{le2025scalable} pioneered a tractable algorithmic approach for the computation of Sobolev IPM, overcoming long-standing limitations in practical use. Moreover,~\citet{tran2026treesliced} leveraged a tree system~\citep{tran2025distancebased, tran2025treesliced, tran2025nonlinear, tran2026revisiting}, and propose injective Radon transforms to further extend the utilization of Sobolev IPM for applications with probability measures in the popular Euclidean spaces and hyperspheres, including gradient flows, generative modeling with diffusion models, and topic modeling. However, Sobolev IPM is intrinsically coupled with $L^p$ geometric structure, thereby limiting its ability to incorporate other prior geometry beyond the $L^p$~\citep{le2025scalable, tran2026treesliced}. Furthermore, we note that one should distinguish Sobolev IPM problem from Sobolev transport~\citep{le2022st} and its generalized approach (i.e., generalized Sobolev transport (GST)~\citep{le2024generalized}) which only constraint on gradient of critic function, while the constraints of Sobolev IPM problem involve both critic function and its gradient within the Sobolev norm.

Unlike Sobolev IPM, the classic optimal transport (OT) has a flexibility to accommodate prior structures by simply modifying the underlying cost function. In a seminal work,~\citet{sturm2011generalized} utilized Orlicz geometric structure to propose Orlicz-Wasserstein (OW) for moving beyond the $L^p$ structure in OT. Comparing to the standard $p$-order Wasserstein, OW~\citep{sturm2011generalized, kell2017interpolation, GuhaHN23, altschuler2023faster} has demonstrated remarkable effectiveness in advancing machine learning methodologies. Notably,~\citet{altschuler2023faster} leverage OW to facilitate the development of differential-privacy-inspired methodologies that address long-lasting convergence challenges in hypocoercive differential equations. Moreover,~\citet{GuhaHN23} demonstrate that OW substantially enhances Bayesian contraction rates, effectively circumventing limitations inherent to traditional OT with Euclidean ground cost. Furthermore, Orlicz structure has found broad applicability across diverse machine learning paradigms~\citep{le2024generalized, le2026efficient, andoni2018subspace, song2019efficient, deng2022fast, chamakh2020orlicz, lorenz2022orlicz}. For comprehensive studies of Orlicz geometry, see~\citep{adams2003sobolev, rao1991theory}.

Inspired by advantages of OW, and to move beyond the $L^p$ paradigm for Sobolev IPM, we advocate to employ a specific class of convex functions for Orlicz geometric structure, and propose the generalized Sobolev IPM (GSI). Our approach includes Sobolev IPM as a special case while enabling diverse prior geometric structures beyond $L^p$ for graph-based IPM. Analogous to the historically computational challenges notoriously inherent in Sobolev IPM, the GSI approach poses significant computational obstacles, which are beyond the recent advancements developed for OW~\citep{le2024generalized} and Sobolev IPM~\citep{le2025scalable}. To address these challenges, we theoretically establish a novel connection between \emph{Orlicz-Sobolev} norm and \emph{Musielak} norm which in turn motivates a novel regularization scheme for the GSI problem.\footnote{As a by-product of the theoretical finding of such connection, one can draw a connection between GST and the proposed GSI problem, as well as its regularization approach, detailed in \S\ref{sec:Properties_GenSobolevIPM}.} Exploiting the underlying graph structure, we further show that the regularized GSI reduces to a simple univariate optimization problem, yielding substantial computational efficiency and enabling practical deployment at scale. Therefore, our approach not only helps to accommodate geometric structures beyond $L^p$ regime for Sobolev IPM, but also overcomes the arisen computational barriers of GSI, and paves the way for its applications, even at scale, in practice.


\paragraph{Contribution.}  Our contributions are three-fold as follows:
\begin{itemize}

\item  We leverage a certain class of convex functions, enabling diverse geometric structures beyond $L^p$ paradigm for Sobolev IPM for the first time, to our knowledge. Additionally, we propose a novel regularization approach for the resulting GSI problem that yields an efficient computation by simply solving a univariate optimization problem.

\item We prove that the regularized GSI is a metric, and it is equivalent to the original GSI. Moreover, we theoretically establish its connections to Sobolev IPM, and other transport distances. 

\item Empirically, we illustrate that the regularized GSI is several-order faster than the popular OW, and its performances compare favorably with other transport baselines for comparing graph-based measures under the same settings in document classification and TDA.


\end{itemize}

\paragraph{Organization.} In \S\ref{sec:preliminaries}, we review related backgrounds and notations. We describe the GSI and its novel regularization in~\S\ref{sec:GenSobolevIPM}. In \S\ref{sec:Properties_GenSobolevIPM}, we provide theoretical properties for the regularized GSI. Then, in \S\ref{sec:experiments}, we empirically show that the regularized GSI is fast in computation, and its performances compare favorably to other transport baselines for comparing graph-based measures under the same settings in document classification and TDA, following by concluding remarks in \S\ref{sec:conclusion}. Supplementary materials including further theoretical and empirical results, detailed discussions, reviews on related notions, and full proofs are available in the Appendix. We also release code for the proposed approach.\footnote{The code repository is on \url{https://github.com/lttam/GenSobolevIPM-Musielak}.}


\section{Backgrounds and Notations}\label{sec:preliminaries}


In this section, we introduce notations, and briefly review graph, Orlicz functions, and Sobolev IPM.



\paragraph{Graph.} We follow the problem setting for probability measures on a given graph, as in~\citep{le2022st, le2024generalized, le2025scalable}. We consider a connected, undirected, finite, and physical graph $\G$ with sets of nodes and edges $V, E$ respectively, and positive edge lengths $\{w_e\}_{e\in E}$. For continuous graph setting, $\G$ is regarded as the set of all nodes in $V$ and all points forming the edges in $E$. Additionally, we equip $\G$ with graph metric $d_{\G}(x,z)$ which equals to the length of the shortest path $[x, z]$ between $x, z \in \G$, and assume that there exists a root node $z_0 \in V$ such that for any considered point $x \in \G$, then $[z_0, x]$ is unique. For $x \in \G, e \in E$, define the sets $\Lambda(x)$ and $\gamma_e$ as follows: 
\begin{align}\label{sub-graph}
 \Lambda(x) := \big\{y\in \G: \, x\in [z_0,y]\big\}, \qquad \gamma_e := \big\{y\in \G: \, e\subset  [z_0,y]\big\}.
\end{align}
Let $\calP(\G)$ (resp.$\,\calP(\G \times\G)$) be the set of all nonnegative Borel measures on $\G$ (resp.$\,\G\times\G$) with a finite mass. By a continuous function $f$ on $\G$, we mean that  $f: \G\to \R$ is continuous w.r.t.~the topology on $\G$ induced by the Euclidean distance. Similar notation is used for continuous functions on $\G \times \G$. 




\paragraph{A family of convex functions.} We consider the set of $N$-functions~\citep[\S8.2]{adams2003sobolev}, where a strictly increasing and convex function $\Phi: [0, \infty)\to [0, \infty)$ is called an $N$-function if  $\lim_{t \to 0} \frac{\Phi(t)}{t} = 0$ and $\lim_{t \to +\infty} \frac{\Phi(t)}{t} = +\infty$.


\paragraph{Orlicz functional space.} For $N$-function $\Phi$, and a nonnegative Borel measure $\lambda$ on $\G$, let $L_{\Phi}(\G, \lambda)$ be the linear hull of the collection of all Borel measurable functions $f: \G \to \R$ satisfying $\int_{\G} \Phi(|f(x)|) \lambda(\text{d}x) < \infty$. Then, $L_{\Phi}(\G, \lambda)$ is a normed space with the \emph{Luxemburg} norm, defined as 
\begin{equation}\label{eq:Luxemburg_norm}
\norm{f}_{L_\Phi}  \coloneqq  \inf \left\{t > 0 \mid \int_{\G} \Phi\!\left(\frac{|f(x)|}{t}\right)\lambda(\text{d}x) \le 1 \right\}.
\end{equation}
For $\Phi(t) = t^p$ with $1 < p < \infty$, then $\norm{f}_{L_\Phi}$ is equal to the $L^p$ norm $\norm{f}_{L^p}$~\citep{adams2003sobolev}. For positive weight function $\hat w$ on $\G$, consider the weighted $L_{\Phi}^{\hat w}(\G, \lambda)$ as the linear hull of the set of all Borel measurable functions $f: \G \to \R$ satisfying $\int_{\G} \hat w(x) \Phi(|f(x)|)  \lambda(\dd x) < \infty$. Then, $L_{\Phi}^{\hat w}(\G, \lambda)$ is a normed space with \emph{Musielak} norm~\citep[\S10.2]{musielak2006orlicz}\footnote{The weighted $L_{\Phi}^{\hat w}(\G, \lambda)$ is a specific instance of the \emph{Musielak-Orlicz} space, where the generalized $N$-function $\bar{\Phi}(x, t) = \hat w(x) \Phi(t)$ for all $t > 0$ and $x \in \G$, and also see~\citep[Definition 3.2.1]{harjulehto2019generalized} for Musielak norm.}, defined as
\begin{equation}\label{eq:weighted_Orlicz_norm}
 \norm{f}_{L_{\Phi}^{\hat w}}  \coloneqq  \inf  \left\{ t > 0 \mid \int_{\G} \hat w(x) \, \Phi\!\left(\frac{|f(x)|}{t}\right) \lambda(\dd x) \le 1 \right\}. \hspace{-0.4em}
\end{equation}
\paragraph{Sobolev IPM.} For an exponent
$1 \le p \le \infty$ and its conjugate $p'$, i.e., $p' \in [1, \infty]$ satisfying $\frac{1}{p} + \frac{1}{p'} = 1$ and if $p=1$, then $p' = \infty$, let $W^{1, p}_0(\G, \lambda)$ be the subspace consisting of all functions $f$ in the graph-based Sobolev space $W^{1, p}(\G, \lambda)$~\citep[Definition 3.1]{le2022st} satisfying $f(z_0) = 0$, $\norm{f}_{W^{1, p}}$ is the Sobolev norm~\citep[\S3.1]{adams2003sobolev}, defined as
\begin{equation}\label{eq:SobolevNorm}
    \norm{f}_{W^{1,p}} := \left( \norm{f}_{L^p}^p + \norm{f'}_{L^p}^p \right)^{\frac{1}{p}}.
\end{equation} 
Let $\calB_{p} := \left\{ f \in W^{1, p}_0(\G, \lambda), \norm{f}_{W^{1, p}} \le 1 \right\}$, then Sobolev IPM between $\mu, \nu \in \calP(\G)$ is defined as
\begin{equation}\label{eq:SobolevIPM}
\calS_{p}(\mu, \nu) := \sup_{f \in \calB_{p}}  \left| \int_{\G} f(x) \mu(\dd x) - \int_{\G} f(y) \nu(\dd y)\right|.
\end{equation}

From Equation~\eqref{eq:SobolevIPM}, Sobolev IPM is intrinsically coupled with the $L^p$ geometric structure in its definition. Consequently, unlike optimal transport, where one can easily adapt to various prior geometric structures by simply tailoring its cost, it is nontrivial for Sobolev IPM to accommodate other structures beyond the $L^p$ regime. Inspired by the seminal work of~\citet{sturm2011generalized}, and advantages of Orlicz-Wasserstein, we will leverage the set of convex $N$-functions for Orlicz geometry for Sobolev IPM to go beyond the coupled $L^p$ structures, and propose the generalized Sobolev IPM in \S\ref{sec:GenSobolevIPM}.\footnote{To ease the readers, in \S\ref{sec:app:brief_reviews}, we further give reviews covering related notions and approaches in the literature.}

\section{Generalized Sobolev IPM and the Regularization Approach}\label{sec:GenSobolevIPM}

Sobolev IPM provides a powerful, yet rigid framework, essentially coupled with $L^p$ within its definition. Thus, it is nontrivial to utilize Sobolev IPM with other structures beyond $L^p$, which is in stark contrast to the flexibility of OT for its adaptivity to diverse geometric structures by simply modifying the underlying cost function. In this section, we leverage convex $N$-functions to generalize Sobolev IPM beyond its coupled $L^p$ structure. We first introduce the \emph{graph-based Orlicz-Sobolev} space~\citep{le2024generalized} and its \emph{Orlicz-Sobolev} norm~\citep[\S 9.3]{rao1991theory},~\citep[\S 3.1, \S 8.30]{adams2003sobolev}. Based on these components, we then describe the definition of the \emph{generalized Sobolev IPM} (GSI) for graph-based measures.

\begin{definition}[Graph-based Orlicz-Sobolev space~\citep{le2024generalized}] \label{def:OrliczSobolev}
Let $\Phi$ be an  $N$-function and $\lambda$ be a nonnegative Borel measure on $\G$. A continuous function $f: \G \to \R$ is said to belong to the graph-based Orlicz-Sobolev space $\OrliczSobolevPhi(\G, \lambda)$ if there exists a function $h\in L_{\Phi}( \G, \lambda) $ satisfying 
\begin{equation}\label{eq:OrliczSobolevFunction}
f(x) -f(z_0) =\int_{[z_0,x]} h(y) \lambda(\mathrm{d}y),  \quad \forall x\in \G.
\end{equation}
Such function $h$ is unique in $L_{\Phi}(\G, \lambda)$ and is called the generalized graph derivative of $f$ w.r.t.~$\lambda$, and this generalized graph derivative of $f$ is denoted $f'$.
\end{definition}
\paragraph{Orlicz-Sobolev norm.} $\OrliczSobolevPhi(\G, \lambda)$ is a normed space with Orlicz-Sobolev norm~\citep{rao1991theory, adams2003sobolev}, defined as
\begin{equation}\label{eq:OrliczSobolevNorm}
    \norm{f}_{\OrliczSobolevPhi} := \norm{f}_{L_{\Phi}} + \norm{f'}_{L_{\Phi}}.
\end{equation}
Let ${\OrliczSobolevPhiZero}(\G, \lambda)$ be the subspace consisting of all functions $f$ in $\OrliczSobolevPhi(\G, \lambda)$ satisfying $f(z_0) = 0$. For $N$-function $\Phi(t) = t^p$ with $1 < p < \infty$, following~\citep[Proposition 4.3]{le2024generalized}, we have $W\!L_{\Phi}^1(\G, \lambda) = W^{1, p}(\G, \lambda)$ where $W^{1, p}$ is the graph-based Sobolev space.\footnote{See a review in \S\ref{app:subsec:review-ST}.} 


\paragraph{Generalized Sobolev IPM (GSI).} Similar to Sobolev IPM, the GSI is an instance of IPM where its critic function belongs to the graph-based Orlicz-Sobolev space, and is constrained within the unit ball of that space. More concretely, given a nonnegative Borel measure $\lambda$ on $\G$, a pair of complementary $N$-functions $\Phi, \Psi$,\footnote{See a review in \S\ref{appsubsec:Orlicz_functions}.} the GSI problem between probability measures $\mu, \nu \in \calP(\G)$ is defined as
\begin{equation}\label{eq:GenSobolevIPM}
\calGS_{\Phi}(\mu, \nu) :=  \sup_{f \in \calB_{\Psi}} \left| \int_{\G} f(x) \mu(\text{d}x) - \int_{\G} f(y) \nu(\text{d}y)\right|,
\end{equation}
where $\calB_{\Psi} := \left\{ f \in {\OrliczSobolevPsiZero}(\G, \lambda),  \norm{f}_{{\OrliczSobolevPsi}} \le 1 \right\}$ is the unit ball defined by Orlicz-Sobolev norm.\footnote{GSI may be regarded as \emph{Orlicz-Sobolev IPM}.} For $N$-function $\Psi(t) = t^p$ with $1 < p < \infty$, the GSI turns into Sobolev IPM. Furthermore, notice that the quantity inside the absolute signs is unchanged if $f$ is replaced by $f - f(z_0)$. Thus, we can assume without loss of generality that $f(z_0) = 0$. This is the motivation for the introduction of the Orlicz-Sobolev space ${\OrliczSobolevPsiZero}(\G, \lambda)$, which shares the same sense to approaches for Sobolev GAN~\citep{Mroueh-2018-Sobolev}, and for Sobolev IPM~\citep{le2025scalable}. 

Analogous to the historically long-standing computational challenges notoriously inherent in Sobolev IPM, the GSI poses significant obstacles in computation for practical use, which are beyond advancements developed for OW and Sobolev IPM in~\citep{le2024generalized, le2025scalable}. 

\paragraph{The regularization approach for the GSI problem.} To address these aforementioned challenges, we theoretically establish a novel relation between the \emph{Orlicz-Sobolev} norm and \emph{Musielak}, ground-based for our novel efficient regularization scheme approach for GSI.

\begin{theorem}[Equivalence]\label{thrm:equivalence_OrliczSobolev_Musielak}
For the length measure $\lambda$ on $\G$, and function $f \in {\OrliczSobolevPhiZero}(\G, \lambda)$, then
    \begin{equation}
        \frac{1}{2} \norm{f'}_{L_{\Phi}^{\hat w}} \le \norm{f}_{\OrliczSobolevPhi} \le \left( 1 + \lambda(\G) \right) \norm{f'}_{L_{\Phi}^{\hat w}},
    \end{equation}
    where the weight function $\hat w$ is defined as
    \begin{equation}\label{eq:weighting_func_wOrlicz}
 	\hat w(x) := 1 + \frac{\lambda(\Lambda(x))}{\lambda(\G)}, \quad \forall x \in \G.
   \end{equation}
\end{theorem}
Theorem~\ref{thrm:equivalence_OrliczSobolev_Musielak} implies that the \emph{Orlicz-Sobolev} norm of a critic function $f \in {\OrliczSobolevPhiZero}(\G, \lambda)$ is equivalent to the \emph{Musielak} norm of its gradient $f'$, where weight function $\hat w$ is given explicitly in Equation~\eqref{eq:weighting_func_wOrlicz}.

\begin{remark}
The novel theoretical finding result in Theorem~\ref{thrm:equivalence_OrliczSobolev_Musielak} plays the key role for the development of our regularization scheme approach. More precisely, we relax the constraint of critic function $f$ within the unit ball induced by Orlicz-Sobolev norm by the Musielak norm constraint of its gradient $f'$ with a specified weight function $\hat w$ (Equation~\eqref{eq:weighting_func_wOrlicz}).
\end{remark}

\paragraph{Discussions.} In~\citep{le2025scalable},~\citeauthor{le2025scalable} established an equivalence between the \emph{Sobolev} norm of a critic function and \emph{weighted $L^p$} norm of its gradient for their weight function ${\hat w}_{\hat{\calS}}(x) := 1 + \lambda(\Lambda(x))$ for all $x \in \G$ by relying on closed-form expressions of (weighted) $L^p$-norm. Such developments are \emph{not} applicable for general convex $N$-functions, beyond $L^p$ regime, in \emph{Orlicz-Sobolev} norm (Equations~\eqref{eq:OrliczSobolevNorm},~\eqref{eq:Luxemburg_norm}) and \emph{Musielak} norm (Equation~\eqref{eq:weighted_Orlicz_norm}). Indeed, in this work, we employ different weight function $\hat w$ (Equation~\eqref{eq:weighting_func_wOrlicz}), formed from our novel theoretical findings,\footnote{See Theorem~\ref{thrm:bound_f_gradf} in Appendix~\ref{sec:app_new_results}, and its important role for the proof of Theorem~\ref{thrm:equivalence_OrliczSobolev_Musielak} in Appendix~\ref{sec:proof:thrm:equivalence_OrliczSobolev_Musielak}.} to establish an equivalence between the \emph{Orlicz-Sobolev} norm of a critic function and \emph{Musielak} norm of its gradient for the theoretical finding result in Theorem~\ref{thrm:equivalence_OrliczSobolev_Musielak}, which is novel, and in fact nontrivial. Unlike the (weighted) $L^p$ geometric structures exploited for Sobolev IPM in~\citep{le2025scalable}, there are no closed-form expressions for \emph{Orlicz-Sobolev} norm and \emph{Musielak} norm for general convex $N$-functions in GSI. In fact, to establish the novel equivalence in Theorem~\ref{thrm:equivalence_OrliczSobolev_Musielak}, we have developed several novel auxiliary theoretical results (Appendix~\ref{sec:app_new_results}), beyond developments in~\citep{le2025scalable, le2024generalized} for Sobolev IPM and OW. Moreover, we emphasize that Theorem~\ref{thrm:equivalence_OrliczSobolev_Musielak} holds for our weight function $\hat w$ (Equation~\eqref{eq:weighting_func_wOrlicz}), and such result may not still hold for any given general weight functions. 

\paragraph{Generalized Sobolev IPM with Musielak regularization (GSI-M).} Hereafter, we employ the weight function $\hat w$ in Theorem~\ref{thrm:equivalence_OrliczSobolev_Musielak}, and define $\calB_{\Psi}^{\hat w} := \left\{f \in {\OrliczSobolevPsiZero} (\G, \lambda), \norm{f'}_{L_{\Psi}^{\hat{w}}} \leq 1 \right\}$. Based on the equivalent relation given by Theorem~\ref{thrm:equivalence_OrliczSobolev_Musielak}, we propose to regularize GSI (Equation~\eqref{eq:GenSobolevIPM}) by relaxing the constraint on the critic function $f$ in the graph-based Orlicz-Sobolev space ${\OrliczSobolevPsiZero}$. More precisely, instead of critic $f$ belonging to the unit ball $\calB_{\Psi}$ of the \emph{Orlicz-Sobolev} space, we propose to constraint critic $f$ within the unit ball $\calB_{\Psi}^{\hat w}$ of the \emph{Musielak} norm of its gradient $f'$. We now formally define the  \emph{generalized Sobolev IPM with Musielak regularization} (GSI-M).\footnote{Unlike a traditional convention for regularization by adding constraints into a problem, we relax the constraint with Orlicz-Sobolev norm by its corresponding Musielak norm, following a similar sense in~\citep{le2025scalable} for Sobolev IPM for simplicity.}
\begin{definition}[Generalized Sobolev IPM with Musielak regularization]\label{def:regSobolevIPM}
Let $\lambda$ be a nonnegative Borel measure on $\G$ and a pair of complementary $N$-functions $\Phi, \Psi$. Then, for probability measures $\mu, \nu \in \calP(\G)$, the generalized Sobolev IPM with Musielak regularization (GSI-M) is defined as
\begin{equation}\label{eq:regGenSobolevIPM}
\widehat{\calGS}_{\Phi}(\mu, \nu) :=  \sup_{f \in \calB_{\Psi}^{\hat w}} \left| \int_{\G} f(x) \mu(\text{d}x) - \int_{\G} f(y) \nu(\text{d}y)\right|. 
\end{equation}
\end{definition}

\paragraph{Computation.} We next show that one can compute the regularized GSI $\widehat{\calGS}_{\Phi}$ by simply solving a univariate optimization problem, paving ways for its practical use in applications.

\begin{theorem}[GSI-M as univariate optimization problem]\label{thrm:RGSI_1d_optimization}
The generalized Sobolev IPM with Musielak regularization $\widehat{\calGS}_{\Phi}(\mu,\nu)$ can be computed as follows:
\begin{align}\label{Amemiya_RGSI}
\widehat{\calGS}_{\Phi}(\mu,\nu ) = \inf_{k > 0} \frac{1}{k}\Bigg\{ 1 +  \int_{\G} \hat w(x) \, \Phi\!\left( \frac{k}{\hat{w}(x)} \left| \mu(\Lambda(x)) -  \nu(\Lambda(x)) \right|\right) \lambda(\text{d}x) \Bigg\}.
\end{align}
\end{theorem}
Notice that in Equation~\eqref{Amemiya_RGSI}, both the subgraph $\Lambda(x)$ (Equation~\eqref{sub-graph}) and the weight function $\hat{w}(x)$ (Equation~\eqref{eq:weighting_func_wOrlicz}) depend on input point $x$ under the integral over $\G$. For practical applications, we next derive an explicit  formula for the integral over graph $\G$ in Equation~\eqref{Amemiya_RGSI} 
when the input probability measures are supported on nodes  $V$ of graph $\G$. This gives an efficient method for computing the GSI-M $\widehat{\calGS}_{\Phi}$. Note that to achieve this result, we use the length measure on graph $\G$~\citep{le2022st} for the nonnegative Borel measure $\lambda$, i.e., we have $\lambda([x, z]) = d_{\G}(x, z), \forall x, z \in \G$. We summarize the result in the following theorem.

\begin{theorem}[Discrete case]\label{thrm:RGSI_1d_optimization_discrete}
Let $\lambda$ be the length measure on $\G$, and $\Phi$ be an $N$-function. Suppose that $\mu,\nu\in \calP(\G)$ are supported on nodes $V$ of graph $\G$.\footnote{An extension for measures supported in graph $\G$ is discussed in Appendix \S\ref{sec:app:further_discussions}.} 
Then we have
\begin{align}\label{equ:Amemiya_RGSI_1d_optimization_discrete}
\widehat{\calGS}_{\Phi}(\mu,\nu ) = \inf_{k > 0} \frac{1}{k}\Bigg\{ 1 + \sum_{e \in E} \left[ \int_{0}^{1} \bar{w}_t(e) \, \Phi \! \left(\frac{k | \bar{h}(e) |}{\bar{w}_t(e)} \right) w_e \dd t \right] \Bigg\},
\end{align}
where $\bar{h}(e) := \mu(\gamma_e) - \nu(\gamma_e)$, and $\bar{w}_t(e) := \frac{w_e}{\lambda(\G)}t + 1 + \frac{\lambda(\gamma_e)}{\lambda(\G)}$ for all edge $e \in E$ and $t \in [0, 1]$.
\end{theorem}
From Theorem~\ref{thrm:RGSI_1d_optimization_discrete}, by exploiting the graph structure for the integral over graph $\G$ in Equation~\eqref{Amemiya_RGSI}, one only needs to simply solve the univariate optimization problem to compute the regularized GSI $\widehat \calGS_{\Phi}$.\footnote{\citet[Theorem 13]{rao1991theory} derived the necessary and sufficient conditions to obtain the infimum for problem~\eqref{equ:Amemiya_RGSI_1d_optimization_discrete}.} We further note that for each edge $e$, given a specific popular $N$-function, then the integral w.r.t.~scalar $t$ has an explicit form for efficient computation, see Appendix \S\ref{app:subsec:thrm:RGSI_1d_optimization_discrete} for details.\footnote{See \S\ref{subsec:app:implementation_notes_GSIM} for further implementation notes for $\gamma_e$ in GSI-M including practical preprocessing procedure, exploiting its sparsity to improve computational complexity, and further discussion on its computational complexity.}

\paragraph{Special case with closed-form expression.} We illustrate that for a specific $N$-function, the regularized GSI $\widehat \calGS_{\Phi}$ can yield a closed-form expression for fast computation. 
\begin{proposition}[Closed-form discrete case]\label{prop:RGSI_closed_form_discrete}
    Suppose that $\mu,\nu\in \calP(\G)$ are supported on nodes $V$ of graph $\G$. For $N$-function $\Phi(t) = \frac{(p-1)^{p-1}}{p^p} t^p$ with $1 < p < \infty$, length measure $\lambda$ on $\G$, then
    \begin{equation}\label{eq:regGenSobolevIPM_discrete}
    \widehat{\calGS}_{\Phi}(\mu, \nu) = \left( \sum_{e \in E} \beta_e \left| \mu(\gamma_e) - \nu(\gamma_e) \right|^p
    \right)^{\frac{1}{p}},
    \end{equation}
    where for each edge $e \in E$ of graph $\G$, the scalar number $\beta_e$ is given by
    \begin{equation}\label{eq:closed_form_edge_weight_regSobolevIPM}
        \beta_e := \left\{
        \begin{array}{ll}
              \lambda(\G)\log{\left(1 + \frac{w_e}{\lambda(\G) + \lambda(\gamma_e)} \right)} & \mbox{if } p = 2,\\
              \frac{\left(\lambda(\G) + \lambda(\gamma_e)+ w_e \right)^{2-p} - \left(\lambda(\G) + \lambda(\gamma_e) \right)^{2-p} }{(2-p)\lambda(\G)^{1-p}} & \text{otherwise.}
        \end{array} 
    \right. 
    \end{equation}
\end{proposition}

The proofs for these theoretical results in \S\ref{sec:GenSobolevIPM} are respectively placed in \S\ref{sec:proof:thrm:equivalence_OrliczSobolev_Musielak} -- \S\ref{sec:proof:prop:RGSI_closed_form_discrete}.


\paragraph{Related work.} We study Sobolev IPM problem for \emph{two probability measures} supported on the \emph{same} graph, as in~\citep{le2022st, le2024generalized, le2025scalable, choromanski2026near}. We distinguish the considered problem with the research lines on computing either kernels~\citep{borgwardt2020graph, sando2025tree, choromanski2023taming} or distances/discrepancies~\citep{petric2019got, xu2019gromov, dong2020copt, brogat2022learning, bai2025fused} between \emph{two input graphs} which are possibly \emph{different}. See Appendix \S\ref{sec:app:further_discussions} for a further discussion.

\textbf{$\bullet$ Relation with regularized Sobolev IPM~\citep{le2025scalable}.} In~\citep{le2025scalable},~\citeauthor{le2025scalable} pioneered a tractable computational approach for Sobolev IPM, addressing historically long-lasting limitations in practical use. More concretely, by exploiting closed-form expressions of $L^p$ norms,~\citet{le2025scalable} developed an efficient regularization for Sobolev IPM by relying on the connections between $L^p$-norm of a critic function and weighted $L^p$-norm of its gradient for the specific weight function $\hat w_{\hat \calS}$. Therefore, such developments in~\citep{le2025scalable} are neither applicable nor directly extendable for the GSI problem~\eqref{eq:GenSobolevIPM}, involving \emph{Luxemburg} norm (Equation~\eqref{eq:Luxemburg_norm}) without closed-form expressions for general convex $N$-functions, for accommodating geometric structures beyond $L^p$. 


\textbf{$\bullet$ Relation with generalized Sobolev transport (GST)~\citep{le2024generalized}.} In~\citep{le2024generalized},~\citeauthor{le2024generalized} developed the generalized Sobolev transport (GST), which is a scalable variant of OW~\citep{sturm2011generalized} for graph-based measures. More concretely,~\citet{le2024generalized} adopted Orlicz geometry for \emph{Sobolev transport} (ST)~\citep{le2022st}, which relaxed the Lipschitz constraint for critic functions as in $1$-order Wasserstein by considering such constraint in a \emph{Sobolev} space. It is worth noting that similar to Sobolev transport, GST only constraints on gradient of critic function. Therefore, the GST problem is essentially different to the considered Sobolev IPM problem where its constraint involves both critic function and its gradient. Consequently, although GST also utilizes Orlicz geometry, such developments in~\citep{le2024generalized} are neither suitable nor directly relevant for the GSI problem~\eqref{eq:GenSobolevIPM}. In the next section, by leveraging our novel finding results in Theorem~\ref{thrm:equivalence_OrliczSobolev_Musielak}, we derive a by-product result on the relation between GST and the GSI problem as well as the proposed regularization approach.

\section{Theoretical Properties of the Regularized GSI}\label{sec:Properties_GenSobolevIPM}

In this section, we derive the metric property for the GSI with Musielak regularization (GSI-M) and establish a relationship for the GSI-M with different $N$-functions. Additionally, we draw connections of the regularized GSI to the original GSI, and other related IPM and transport distances including the regularized Sobolev IPM~\citep{le2025scalable}, generalized Sobolev transport (GST)~\citep{le2024generalized} and OW~\citep{sturm2011generalized}.\footnote{See \S\ref{sec:app:brief_reviews} for a review on these IPM and transport distances.}


\begin{theorem}[Metrization]\label{thrm:metrization}
The generalized Sobolev IPM with Musielak regularization $\widehat \calGS_{\Phi}$ is a metric on the space $\calP(\G)$ of probability measures on graph $\G$.
\end{theorem}
The GSI-M is monotone with respect to the $N$-function $\Phi$ as shown in the next result. Consequently, it may enclose a stronger notion of metrics than Sobolev IPM for comparing graph-based measures.

\begin{proposition}[GSI-M with different $N$-functions]\label{prop:strong_metric}
For any two $N$-functions $\Phi_1, \Phi_2$ satisfying  $\Phi_1(t) \le \Phi_2(t)$ for all $t \in \R_+$, and $\mu, \nu \in \calP(\G)$, then we have
\[
\widehat \calGS_{\Phi_1}(\mu,\nu ) \le \widehat \calGS_{\Phi_2}(\mu,\nu ).
\]
\end{proposition}

\paragraph{Connections to other IPM and transport distances.} We then theoretically establish relations between the regularized GSI \eqref{eq:regGenSobolevIPM} and other related IPM and transport approaches. Consequently, we can theoretically position the regularized GSI within the research picture of the literature.


\textbf{$\bullet$ Original GSI.} We show that the regularized GSI~\eqref{eq:regGenSobolevIPM} is equivalent to the original GSI~\eqref{eq:GenSobolevIPM}.
\begin{theorem}[Connection between GSI-M and original GSI]\label{thrm:relationGenSobolevIPM}
For a nonnegative Borel measure $\lambda$ on $\G$, $N$-function $\Phi$, $\mu, \nu \in \calP(\G)$, then
\begin{equation}\label{eq:equivalence_SobolevIPM}
\hspace{-0.1em} \frac{1}{2} \, \calGS_{\Phi}(\mu, \nu) \le \widehat \calGS_{\Phi}(\mu, \nu) \le (1 + \lambda(\G)) \, \calGS_{\Phi}(\mu, \nu), \hspace{-1em}
\end{equation}
\end{theorem}


\textbf{$\bullet$ Regularized Sobolev IPM~\citep{le2025scalable}.} Let $\hat \calS_p$ be the regularized $p$-order Sobolev IPM.
\begin{proposition}[Connection between GSI-M and regularized Sobolev IPM]\label{prop:Connection_GSIMPMR_RSIPM}
Let $\Phi(t) := \frac{(p-1)^{p-1}}{p^p} t^p$ with $1 < p < \infty$, $\hat c_1 := \max(1, \lambda(\G)^{-1})^{\frac{1-p}{p}}$, and $\hat c_2 := \min(1, \lambda(\G)^{-1})^{\frac{1-p}{p}}$. Then, for a nonnegative Borel measure $\lambda$ on graph $\G$, and measures $\mu, \nu \in \calP(\G)$, we have
\begin{equation}\label{eq:GSIMR_RSIPM}
    \hat c_1 \, \hat \calS_p(\mu, \nu) \le \widehat \calGS_{\Phi}(\mu, \nu) \le \hat c_2 \, \hat \calS_p(\mu, \nu).
\end{equation}
\end{proposition}

Proposition~\ref{prop:Connection_GSIMPMR_RSIPM} illustrates that for a certain given $N$-function, the regularized GSI $\widehat \calGS$~\eqref{eq:regGenSobolevIPM} is provably equivalent to the regularized Sobolev IPM $\hat \calS_p$~\citep{le2025scalable}. Additionally, $\widehat \calGS$ can be considered as a variational generalization for $\hat \calS_p$ to incorporate more general geometric priors, beyond $L^p$ paradigm.

\textbf{$\bullet$ Generalized Sobolev transport (GST)~\citep{le2024generalized}.} Denote $\calGST_{\Phi}$ for the GST with $N$-function $\Phi$.
\begin{proposition}[Connection between GSI-M and generalized Sobolev transport]\label{prop:Connection_GSIMPMR_GST}
For a nonnegative Borel measure $\lambda$ on $\G$, then for all measures $\mu, \nu \in \calP(\G)$, we have
\begin{equation}\label{eq:GSIMR_GST}
    \frac{1}{2} \, \calGST_{\Phi}(\mu, \nu) \le \widehat \calGS_{\Phi}(\mu, \nu) \le \calGST_{\Phi}(\mu, \nu).
\end{equation}
\end{proposition}

Propositions~\ref{prop:Connection_GSIMPMR_GST} illustrates that GSI-M is provably equivalent to GST for the same convex $N$-functions. Similar to GST, GSI-M constraints on gradient of a critic function, thereby GSI-M may be regarded as a weighted variant of GST. However, we emphasize that the weight function $\hat w$ (Equation~\eqref{eq:weighting_func_wOrlicz}) plays the key role to establish the equivalence in Theorem~\ref{thrm:equivalence_OrliczSobolev_Musielak} between \emph{Orlicz-Sobolev} norm of a critic function in original GSI problem~\eqref{eq:GenSobolevIPM} and \emph{Musielak} norm of its gradient in the regularized GSI problem~\eqref{eq:regGenSobolevIPM}. Therefore, GSI-M with the weight function $\hat w$ provides an efficient regularization approach for the GSI problem, theoretically ground-based by novel results in Theorem~\ref{thrm:equivalence_OrliczSobolev_Musielak}. 



\textbf{$\bullet$ Orlicz-Wasserstein (OW)~\citep{sturm2011generalized}.} Let $\calOW$ denote the Orlicz-Wasserstein.
\begin{proposition}[Connection between GSI-M and Orlicz-Wasserstein]\label{prop:Connection_GSIMPMR_OW}
Consider the limit case $\Phi(t) := t$, and graph $\G$ is a tree.\footnote{Although $\Phi(t) = t$ is not an $N$-function due to its linear growth, it can be regarded as the limit $p \to 1^+$ for $\Phi(t) = t^p$.} For a nonnegative Borel measure $\lambda$ on $\G$, $\mu, \nu \in \calP(\G)$, then
\begin{equation}\label{eq:GSIMR_OW}
   \frac{1}{2} \, \calOW(\mu, \nu) \le \widehat \calGS_{\Phi}(\mu, \nu) \le \calOW(\mu, \nu).
\end{equation}
\end{proposition}

Propositions~\ref{prop:Connection_GSIMPMR_OW} shows that GSI-M $\widehat \calGS$ is provably equivalent to OW when graph $\G$ is a tree, and $\Phi(t) = t$. Both GSI-M and OW are capable to employ Orlicz structures with general convex $N$-functions. However, OW is challenging for computation, hinders its practical applications while GSI-M is computationally efficient, by simply solving a univariate optimization problem.

The proofs for the theoretical results in \S\ref{sec:Properties_GenSobolevIPM} are respectively placed in \S\ref{app:proof:thrm:metrization} -- \S\ref{app:proof:prop:Connection_GSIMPMR_OW}. See Appendix \S\ref{app:subsec:further_theory} for further theoretical results of the regularized GSI.



\section{Experiments}
\label{sec:experiments}

In this section, we illustrate that GSI-M is fast for computation, which is several-order faster than the popular OW. Following the problem setups in~\citep{le2025scalable}, we evaluate GSI-M for probability measures on a given graph. We show that performances of GSI-M compare favorably to other transport distances to compare graph-based measures under the same settings for document classification and TDA. Note that there are different approaches for document classification and TDA. However, our goals are to compare different transport distances for graph-based probability measures under the same settings as considered in~\citep{le2025scalable}.

\paragraph{Datasets.} For document classification, we consider $4$ real-world datasets: \texttt{TWITTER, RECIPE, CLASSIC, AMAZON} where their properties are given in Figure~\ref{fg:DOC_10KLog_main}. For TDA, we consider $2$ tasks: (i) orbit recognition on the synthesized \texttt{Orbit} dataset~\citep{adams2017persistence} for linked twist map, a discrete dynamical system modeling flow, which is used to model flows in DNA microarrays~\citep{hertzsch2007dna}, and (ii) object shape image classification on a $10$-class subset of \texttt{MPEG7} dataset~\citep{latecki2000shape}. We summarize the dataset properties in Figure~\ref{fg:TDA_mix10KLog_main}. The objects of interests for TDA are represented by persistence diagrams (PD)~\citep{edelsbrunner2008persistent}. 

Following~\citep[\S5]{le2025scalable}, we regard a document as a distribution over its words, and PD as a distribution over its topological features. We employ the graphs $\G_{\text{Log}}$ and $\G_{\text{Sqrt}}$ for words and topological features in document classification and TDA respectively.\footnote{See a review in Appendix~\S\ref{sec:app:further_discussions}.} Thus, document and PD are represented as probability measures supported on these given graphs $\G_{\text{Log}}$ and $\G_{\text{Sqrt}}$.

\paragraph{Set up.} We employ kernelized support vector machine (SVM) for both document classification and TDA. We use kernel $\exp(-\tilde{t} \tilde{d})$ with hyperparameter $\tilde{t} > 0$ and $\tilde{d}$ is the considered IPM/transport distances for graph-based measures. Following~\citep{Cuturi-2013-Sinkhorn}, we regularize the Gram matrices of indefinite kernels by adding a sufficiently large diagonal term, and use 1-vs-1 strategy for multi-class classification with SVM. We randomly split each dataset into $70\%/30\%$ for training and test with $10$ repeats. We typically utilize cross validation for hyper-parameters. For SVM regularization hyperparameter, we choose it from $\left\{0.01, 0.1, 1, 10\right\}$. For the kernel hyperparameter $\tilde{t}$, we choose it from $\{1/q_{s}, 1/(2q_{s}), 1/(5q_{s})\}$ where $q_s$ is the $s\%$ quantile of a random subset of distances observed on a training set and $s = 10, 20, \dotsc, 90$. For the root node $z_0$ in graph $\G$, we choose it from a random $10$-root-node subset of $V$ in $\G$. Note that we include preprocessing procedures, e.g., computing shortest paths, into reported time consumptions. Additionally, we use a second-order solver (i.e., \emph{fmincon} with Trust Region Reflective solver) for solving the \emph{univariate} optimization problem for GSI-M. Following~\citep{GuhaHN23, altschuler2023faster, le2024generalized, andoni2018subspace, song2019efficient, chamakh2020orlicz, adams2003sobolev, rao1991theory, chewi2023optimization, vershynin2018high}, we evaluate two most popular $N$-functions widely used in the literature: $\Phi_1(t) = \exp(t) - t - 1$, and $\Phi_2(t) = \exp(t^2) - 1$. We also examine the limit case: $\Phi_0(t) = t$ as in~\citep{GuhaHN23, le2024generalized}. To illustrate the scale, e.g., in \texttt{AMAZON}, we need to evaluate distances for more than $29$ million pairs of probability measures for each SVM run.\footnote{See further details in \S\ref{sec:app:further_exp_details_empirical_results}.}

\subsection{Computation}\label{subsec:computation}
 

\begin{wrapfigure}{r}{0.45\textwidth}
  \vspace{-18pt}
  \begin{center}
    \includegraphics[width=0.45\textwidth]{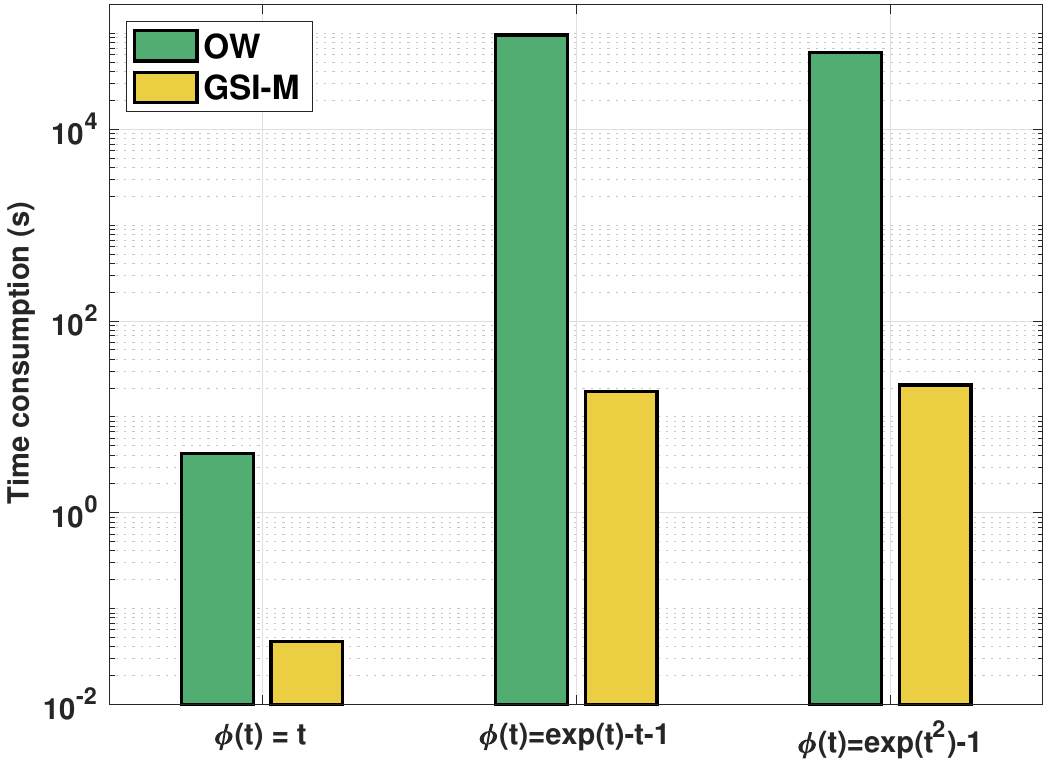}
  \end{center}
  \vspace{-6pt}
  \caption{Time consumption on $\G_{\text{Log}}$.}
  \label{fg:Time_GSIM_GST_OW_10K_LLE}
 \vspace{-10pt}
\end{wrapfigure}

We compare the time consumption of GSI-M and OW with $\Phi_0, \Phi_1, \Phi_2$ on random 10K pairs of measures in \texttt{AMAZON} with graph $\G_{\text{Log}}$ having 1K nodes and about 7K edges. Figure~\ref{fg:Time_GSIM_GST_OW_10K_LLE} illustrates that the computation of GSI-M is several-order faster than OW. More concretely, GSI-M is $90\times, 5100\times, 2900\times$ faster than OW for $\Phi_0, \Phi_1, \Phi_2$ respectively. For $N$-functions $\Phi_1$ and $\Phi_2$, GSI-M takes less than \emph{$22$ seconds} while OW needs at least \emph{$17$ hours}, and up to \emph{$27$ hours} for the computation. We remark that following Proposition~\ref{prop:RGSI_closed_form_discrete}, by taking the limit $p \to 1^+$, GSI-M with $\Phi_0$ has closed-form expression for fast computation. Additionally, OW with $\Phi_0$ is equal to the OT with graph metric ground cost~\citep{GuhaHN23, le2024generalized}. Therefore, the computations of GSI-M and OW with $\Phi_0$ are more efficient than with $\Phi_1, \Phi_2$ as expectation.



\subsection{Document classification}\label{subsec:document}

\begin{figure}[h]
  \begin{center}
\includegraphics[width=0.95\textwidth]{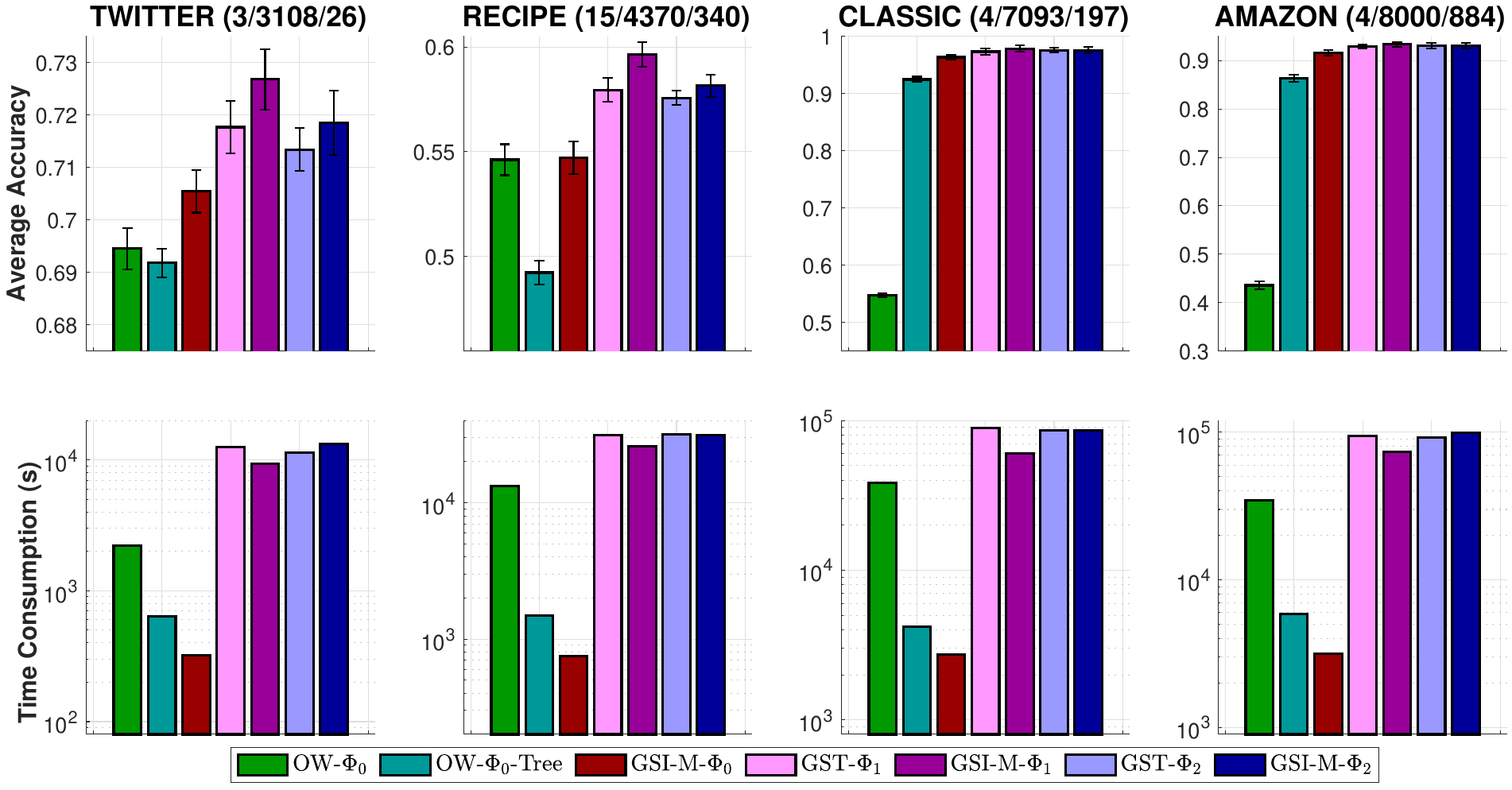}
  \end{center}
  \vspace{-4pt}
  \caption{SVM results and time consumption for kernel matrices with graph $\G_{\text{Log}}$. For each dataset, the numbers in the parenthesis are the number of classes; the number of documents; and the maximum number of unique words for each document respectively.}
  \label{fg:DOC_10KLog_main}
\end{figure}

\paragraph{Set up.} We evaluate GSI-M with $\Phi_0, \Phi_1, \Phi_2$, denoted as GSI-M-$\Phi_i$  for $i=0, 1, 2$ respectively. For OW, we only use $\Phi_0$, denoted as OW-$\Phi_0$, but exclude $\Phi_1, \Phi_2$ due to their prohibited computations (Figure~\ref{fg:Time_GSIM_GST_OW_10K_LLE}). We include GST~\citep{le2024generalized}, which is a scalable alternative for OW in applications, for $\Phi_0, \Phi_1, \Phi_2$, denoted as GST-$\Phi_i$ for $i=0, 1, 2$ respectively.\footnote{Note that for the limit $\Phi_0$, GSI-M is equal to GST, i.e., GSI-M-$\Phi_0$ = GST-$\Phi_0$.} Additionally, we also carry out a special case of OW-$\Phi_0$ where it is computed on a random tree extracted from the given graph $\G$, denoted as OW-$\Phi_0$-Tree. We remark that OW-$\Phi_0$-Tree is equal to tree-Wasserstein~\citep{LYFC}. We use graph $\G_{\text{Log}}$ with 10K nodes and about 100K edges; and graph $\G_{\text{Sqrt}}$ with 10K nodes, about 1M edges~\citep[\S5]{le2025scalable}.


\paragraph{Results.} Figure~\ref{fg:DOC_10KLog_main} illustrates that the performances of GSI-M with all $\Phi$ functions compare favorably to those baselines. GSI-M-$\Phi_1$ and GSI-M-$\Phi_2$ improve performances of GSI-M-$\Phi_0$, especially in \texttt{RECIPE}, but their computational time is slower since GSI-M-$\Phi_0$ has a closed-form expression (\S\ref{subsec:computation}). Therefore, it implies that \emph{Orlicz geometric structure} in GSI-M may be helpful for graph-based document classification, similar to observations in~\citep{le2024generalized}. Additionally, performances of GSI-M compare favorably to those of GST with the same $N$-functions $\Phi_1, \Phi_2$, especially in \texttt{TWITTER} and \texttt{RECIPE}. Performances of OW-$\Phi_0$-Tree are better in \texttt{CLASSIC} and \texttt{AMAZON}, but worse in \texttt{TWITTER} and \texttt{RECIPE} than those of OW-$\Phi_0$. Although OW-$\Phi_0$-Tree only leverages a partial information of $\G$, it forms positive definite kernels while kernels of OW-$\Phi_0$ are indefinite. 




\subsection{Topological Data Analysis}\label{subsec:TDA}

\paragraph{Set up.} We use the same set up as document classification. Except in \texttt{MPEG7}, due to its smaller size, we use $\G_{\text{Log}}$ with 1K nodes and about 7K edges for $\G_{\text{Log}}$; and $\G_{\text{Sqrt}}$ with 1K nodes and about 32K edges as in~\citep{le2025scalable}.

\begin{figure}[h]
  \begin{center}
    \includegraphics[width=0.55\textwidth]{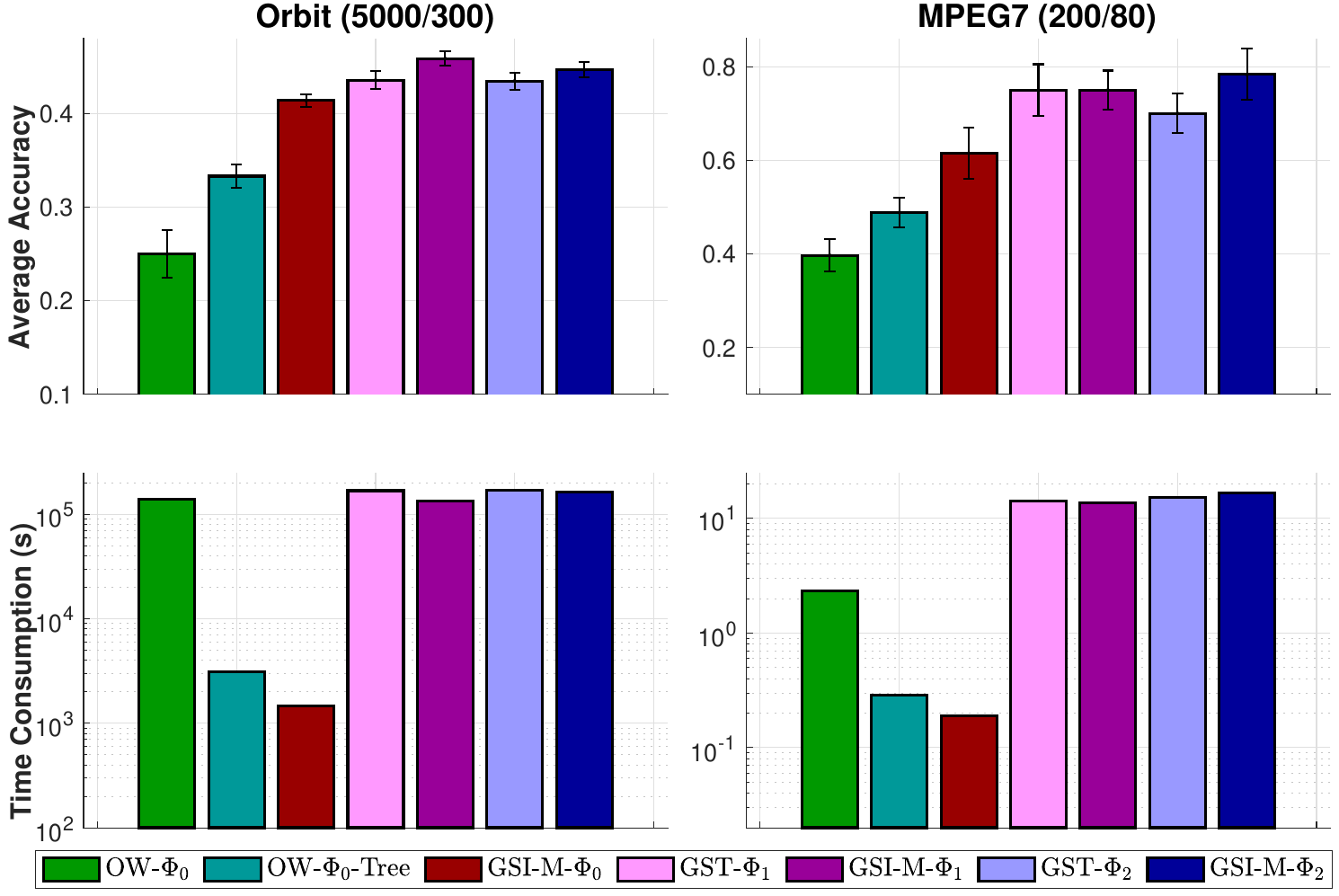}
  \end{center}
  \vspace{-4pt}
  \caption{SVM results and time consumption for kernel matrices with graph $\G_{\text{Log}}$. For each dataset, the numbers in the parenthesis are respectively the number of PD; and the maximum number of points in PD.}
  \label{fg:TDA_mix10KLog_main}
\end{figure}

\paragraph{Results.} Figure~\ref{fg:TDA_mix10KLog_main} shows that we obtain similar observations as for document classification. For the same $\Phi$ function, GSI-M is several-order faster OW. Performances of GSI-M-$\Phi_1$ GSI-M-$\Phi_2$ compare favorably to those of GSI-M-$\Phi_0$, especially in \texttt{MPEG7}, but they are also slower. Therefore, similar to observations in~\citep{le2024generalized}, \emph{Orlicz geometric structure} in GSI-M may be also useful for graph-based TDA. Additionally, performances of GSI-M compare favorably to those of GST with the same $N$-functions $\Phi_1, \Phi_2$, especially in \texttt{MPEG7} for $N$-function $\Phi_2$. Although OW-$\Phi_0$-Tree use a partial information of $\G$, the positive definiteness of its corresponding kernel may help to improve performances of OW-$\Phi_0$. 

For further empirical results for both document classification and TDA, e.g., ablation studies on different graph structures (e.g., number of vertices, number of edges), different $N$-functions, different root nodes, effects of minor perturbation on edge weights of a given graph, empirical gaps between GSI-M and GST/OW, we refer the readers in Appendix~\ref{sec:app:further_exp_details_empirical_results} for rigorous details due to the limited space.

\section{Conclusion}
\label{sec:conclusion}

In this work, we leverage convex $N$-functions to adopt Orlicz geometric structure for Sobolev IPM to go beyond $L^p$. Moreover, we propose a novel regularization for GSI, which is efficient for computation by simply solving a univariate optimization problem. For future works, it is interesting to consider whether it is possible to derive efficient algorithmic approaches for the original Sobolev IPM and its generalization (without regularization), and go beyond the Sobolev geometric structures, e.g., Besov IPM, for applications. Additionally, automatically learning $N$-functions for GSI beyond the traditional cross-validation will be also an impactful direction for further investigations in depth. Moreover, in this work, we investigate GSI for measures on a given graph. Therefore, constructing the best graph for a given task is also an interesting open problem for research on graph-based IPM. Also, see Appendix \S\ref{sec:app:further_discussions} for further discussions and limitations.


\begin{ack}



We thank the area chairs and anonymous reviewers for their comments. KF has been supported in part by Grant-in-Aid for Transformative Research Areas (A) JP22H05106 and JSPS KAKENHI JP26K02990. HH acknowledges the support of JSPS KAKENHI JP25H01494, 26K02989, and 26K23861. TL gratefully acknowledges the support of JST BOOST Grant number JPMJBY25F8.

\end{ack}

\bibliography{bibSobolev25, bibSobolev22, bibEPT21}
\bibliographystyle{icml2026}

\newpage
\input{checklist.tex}

\newpage
\appendix

\begin{center}
{\bf{\Large{\textit{Supplement to}  ``Orlicz–Sobolev with Musielak: An Efficient Regularization Approach for Graph-based IPM''}}}
\end{center}

In this appendix, we provide further theoretical results in~\S\ref{app:sec:further_theoretical_results}, further experiment experimental details and empirical results in~\S\ref{sec:app:further_exp_details_empirical_results}, and further discussions in \S\ref{sec:app:further_discussions}. Then, we provide reviews for related notions and approaches in the literature in \S\ref{sec:app:brief_reviews}. Detailed proofs for theoretical results in the main manuscript and additional theoretical results in~\S\ref{app:sec:further_theoretical_results} are placed in~\S\ref{app:sec:proofs}.


\section{Further Theoretical Results}\label{app:sec:further_theoretical_results}

\subsection{Auxiliary Results}\label{sec:app_new_results}

In this section, we describe auxiliary theoretical finding results, which play important roles to derive our theoretical results presented in the main text. 

\begin{lemma}\label{lem:ineq_f_gradf}
Let $\Phi$ be an $N$-function and $f \in {\OrliczSobolevPhiZero}(\G, \lambda)$. Then for any nonnegative weight function $\hat w$ and $t > 0$, we have
\begin{eqnarray}
\int_{\G} \hat w(x) \, \Phi\!\left( \frac{1}{t} \left|f(x)\right| \right) \lambda(\dd x) 
\le \frac{1}{\lambda(\G)} \int_{\G} \Phi \! \left( \frac{\lambda(\G)}{t} \left|f'(y)\right| \right) \bar{\lambda}_{\hat w}(\Lambda(y)) \lambda(\dd y),
\end{eqnarray}
where we write $\bar{\lambda}_{\hat w}$ for measure $\lambda$ weighted by function $\hat w$, i.e., $\int \bar{\lambda}_{\hat w}(\dd x) := \int \hat w(x) \lambda(\dd x)$.
\end{lemma}
The proof is placed in Appendix~\S\ref{app:proof:lem:ineq_f_gradf}.

\begin{lemma}\label{lem:bound_fnorm}
    Let $w_0$ be any positive weight function such that $w_0(x) \ge \lambda(\Lambda(x))$ for all $x \in \G$. Then if function $f \in {\OrliczSobolevPhiZero}(\G, \lambda)$ and scalar $t > 0$ satisfying $\int_{\G} w_0(x) \, \Phi\!\left(\frac{|f'(x)|}{t}\right)\lambda(\dd x) \le \lambda(\G)$, we must have 
    \begin{equation}
    t \ge \frac{\norm{f}_{L_\Phi}}{\lambda(\G)}.
    \end{equation}
\end{lemma}
The proof is placed in Appendix~\S\ref{app:proof:lem:bound_fnorm}.

\begin{theorem}\label{thrm:bound_f_gradf}
   Let $w_0$ be any positive weight function satisfying $w_0(x) \ge \lambda(\Lambda(x))$ for all $x \in \G$. For every function $f \in {\OrliczSobolevPhiZero}(\G, \lambda)$, then
        \begin{equation}
        \norm{f}_{L_{\Phi}} \le \lambda(\G) \norm{f'}_{L_{\Phi}^{{w_0}/{\lambda(\G)}}}.
    \end{equation}
\end{theorem}
The proof is placed in Appendix~\S\ref{app:proof:thrm:bound_f_gradf}.

\begin{lemma}\label{lem:different_weight_func}
Let $\Phi$ be an $N$-function.
For any two positive weight functions $w_1, w_2$ such that $w_1(x) \ge w_2(x)$ for all $x \in \G$, and any Borel measurable function $f$ on $\G$, then
\begin{equation}
\norm{f}_{L_{\Phi}^{w_1}} \ge \norm{f}_{L_{\Phi}^{w_2}}.
\end{equation}
\end{lemma}
The proof is placed in Appendix~\S\ref{app:proof:lem:different_weight_func}.

\begin{lemma}\label{lem:Nfunc_multiplicative}
    For $N$-function $\Phi$, scalar $t > 0$, and scalar $k \ge 1$, then
    \begin{equation}
        \Phi(kt) \ge k \Phi(t).    
    \end{equation}
\end{lemma}
The proof is placed in Appendix~\S\ref{app:proof:lem:Nfunc_multiplicative}.

\begin{lemma}\label{lem:MusielakNorm_kPhi}
For $N$-function $\Phi$, positive weight function $\hat w$ on $\G$, $k > 0$, then we have
\begin{equation}
\norm{kf}_{L_{\Phi}^{\hat w}} = k \norm{f}_{L_{\Phi}^{\hat w}}.
\end{equation}
\end{lemma}
The proof is placed in Appendix~\S\ref{app:proof:lem:MusielakNorm_kPhi}.

\subsection{Further Theoretical Results}\label{app:subsec:further_theory}

In this section, we provide further theoretical results, complemented to our results in the main text.

\paragraph{Special case with closed-form expression.} For specific $N$-function, the generalized Sobolev IPM can yield a closed-form expression for computation.

\begin{proposition}[Closed-form expression]\label{prop:RGSI_closed_form}
    For $N$-function $\Phi(t) = \frac{(p-1)^{p-1}}{p^p} t^p$ with $1 < p < \infty$, the generalized Sobolev IPM with Musielak regularization has a closed-form expression as follows:
    \begin{eqnarray}
        \widehat{\calGS}_{\Phi}(\mu, \nu) = \left[ \int_{\G} \hat w(x)^{1-p} \left| \mu(\Lambda(x)) - \nu(\Lambda(x)) \right|^p \lambda (\dd x) \right]^{\frac{1}{p}}. 
    \end{eqnarray}
\end{proposition}
The proof is placed in Appendix \S\ref{sec:proof:prop:RGSI_closed_form}.

\paragraph{Further connections between the regularized GSI with other IPM/transport distances.} We establish further connections between the regularized GSI with original Sobolev IPM, Sobolev transport~\citep{le2022st}, and optimal transport.

\paragraph{$\bullet$ Original Sobolev IPM.} Denote $\calS_p$ for the original $p$-order Sobolev IPM.
\begin{proposition}[Connection between GSI-M and original Sobolev IPM]\label{prop:Connection_GSIMPMR_SIPM}
Let $\Phi(t) := \frac{(p-1)^{p-1}}{p^p} t^p$ with $1 < p < \infty$. For a nonnegative Borel measure $\lambda$ on $\G$, and measures $\mu, \nu \in \calP(\G)$, then
\begin{equation}\label{eq:GSIMR_RSIPM}
    c_1 \, \calS_p(\mu, \nu) \le \widehat \calGS_{\Phi}(\mu, \nu) \le c_2 \, \calS_p(\mu, \nu),
\end{equation}
where $c_1 := \left[ \frac{\min(1, \lambda(\G)^{p-1}) \max(1, \lambda(\G)^{-1})^{1-p}}{1 + \lambda(\G)^{p}} \right]^{\frac{1}{p}}$, $c_2 := \left[\min(1, \lambda(\G)^{-1})^{1-p} \max(1,  \lambda(\G)^{p-1})\right]^{\frac{1}{p}}$.
\end{proposition}
The proof is placed in Appendix~\S\ref{app:proof:prop:Connection_GSIMPMR_SIPM}.

We note that the original GSI~\eqref{eq:GenSobolevIPM} includes Sobolev IPM as a special case. As discussed in \S\ref{sec:GenSobolevIPM}, for $N$-function $\Psi(t) = t^p$ with $1 < p < \infty$, the GSI turns into Sobolev IPM. Proposition~\ref{prop:Connection_GSIMPMR_SIPM} shows a connection between the regularized GSI~\eqref{eq:regGenSobolevIPM} with the original Sobolev IPM.

\paragraph{$\bullet$ Sobolev transport.} Denote $\calST_{p}$ for the $p$-order Sobolev transport.
\begin{proposition}[Connection between GSI-M and Sobolev transport]\label{prop:Connection_GSIMPMR_ST}
For a nonnegative Borel measure $\lambda$ on $\G$, $N$-function $\Phi(t) := \frac{(p-1)^{p-1}}{p^p} t^p$ with $1 < p < \infty$, and $\mu, \nu \in \calP(\G)$, then we have
\begin{equation}\label{eq:GSIMR_ST}
    \frac{1}{2} \, \calST_{p}(\mu, \nu) \le \widehat \calGS_{\Phi}(\mu, \nu) \le \calST_{p}(\mu, \nu).
\end{equation}
\end{proposition}
The proof is placed in Appendix~\S\ref{app:proof:prop:Connection_GSIMPMR_ST}.

\paragraph{$\bullet$ Optimal transport.} Denote $\calW_1$ for the $1$-order Wasserstein.
\begin{proposition}[Connection between GSI-M and optimal transport]\label{prop:Connection_GSIMPMR_OT}
Under the same assumptions in Proposition~\ref{prop:Connection_GSIMPMR_OW}, then for all measures $\mu, \nu \in \calP(\G)$, we have
\begin{equation}\label{eq:GSIMR_OW}
    \frac{1}{2} \, \calW_1(\mu, \nu) \le \widehat \calGS_{\Phi}(\mu, \nu) \le \calW_1(\mu, \nu).
\end{equation}
\end{proposition}
The proof is placed in Appendix~\S\ref{app:proof:prop:Connection_GSIMPMR_OT}. 

\subsection{Discrete Case for Generalized Sobolev IPM with Musielak Regularization with Popular $N$-Functions}\label{app:subsec:thrm:RGSI_1d_optimization_discrete}

In this section, we provide rigorous details for discrete case for generalized Sobolev IPM with Musielak regularization with popular $N$-functions, which are used for implementations.

Following Theorem~\ref{thrm:RGSI_1d_optimization_discrete}, the discrete case of the regularized generalized Sobolev IPM $\widehat{\calGS}_{\Phi}$ is
\begin{align}
\widehat{\calGS}_{\Phi}(\mu,\nu ) = \inf_{k > 0} \frac{1}{k}\left( 1 + \sum_{e \in E} \left[ \int_{0}^{1} \bar{w}_t(e) \, \Phi \! \left(\frac{k | \bar{h}(e) |}{\bar{w}_t(e)} \right) w_e \dd t \right] \right),
\end{align}
where $\bar{h}(e) := \mu(\gamma_e) - \nu(\gamma_e)$ for every edge $e \in E$, and $\bar{w}_t(e) := \frac{w_e}{\lambda(\G)}t + 1 + \frac{\lambda(\gamma_e)}{\lambda(\G)}$ for all edge $e \in E$ and $t \in [0, 1]$.

We consider popular practical $N$-functions $\Phi$ for the GST and the OW: $\Phi_1(t) = \exp(t) - t - 1$, and $\Phi_2(t) = \exp(t^2) - 1$. We also examine the limit case: $\Phi_0(t) = t$.

For each edge $e \in E$, we want to compute
\begin{equation}
\mathcal{A}_{\Phi}(e) := \int_{0}^{1} \left[ \frac{w_e}{\lambda(\G)}t + 1 + \frac{\lambda(\gamma_e)}{\lambda(\G)} \right] \, \Phi \! \left(\frac{k | \bar{h}(e) |}{\frac{w_e}{\lambda(\G)}t + 1 + \frac{\lambda(\gamma_e)}{\lambda(\G)}} \right) w_e \dd t
\end{equation}

For simplicity, let denote the exponential integral function for $z > 0$ as follow:
\begin{equation}
Ei(z) = \int_{-\infty}^{z} \frac{\exp(t)}{t} \dd t,
\end{equation}
and notice that 
\begin{equation}
\frac{\dd}{\dd z}Ei(z) = \frac{\exp(z)}{z}, \quad z \ne 0,
\end{equation}
and $Ei(z) = - E_1(-z)$, where $E_1(z) = \int_{z}^{\infty} \frac{\exp(-t)}{t} \dd t$.

\paragraph{For the limit case: $\Phi_0(t) = t$.} We have
\begin{equation}
\mathcal{A}_{\Phi_0}(e) := \int_{0}^{1} k | \bar{h}(e) | w_e \dd t = k | \bar{h}(e) | w_e.
\end{equation}

\paragraph{For $N$-function: $\Phi_1(t) = \exp(t) - t - 1$.} For $a > 0, b > 0$, we consider
\begin{equation}
A_1 := \beta \int_{0}^{1} (at + b) \, \Phi_1\!\left(\frac{\alpha}{at + b}\right)\dd t.
\end{equation}
See Appendix~\S\ref{subsec:app:compute_A1} for the detailed computation of $A_1$ where its results are summarized as follow:
\begin{eqnarray}
    A_1 = \frac{\beta}{2a}\Bigg[ -\alpha^2 \left[ Ei\!\left(\frac{\alpha}{a+b}\right) - Ei\!\left(\frac{\alpha}{b}\right) \right] + \alpha\left[(a+b)\exp\!\left(\frac{\alpha}{a+b} \right) - b \exp\!\left(\frac{\alpha}{b} \right)\right] \nonumber \hspace{3em} \\
    + \left[(a+b)^2\exp\!\left(\frac{\alpha}{a+b} \right) - b^2 \exp\!\left(\frac{\alpha}{b} \right)\right] - 2\alpha a - \left[(a+b)^2 - b^2 \right] \Bigg].
\end{eqnarray}
    
Therefore, we obtain
\begin{equation}
    \mathcal{A}_{\Phi_1} = A_1,
\end{equation}
where $\beta = w_e, a = \frac{w_e}{\lambda(\G)}, b = 1 + \frac{\lambda(\gamma_e)}{\lambda(\G)}, \alpha = k|\bar h(e) |$.

Additionally, the first-order derivative of $A_1$ w.r.t. $k$ is as follows:
\begin{eqnarray}
    \frac{\dd}{\dd k}A_1 = \frac{\beta |\bar h(e)|}{a} \left\{ -\alpha \left[ Ei\!\left(\frac{\alpha}{a+b}\right) - Ei\!\left(\frac{\alpha}{b}\right) \right] +(a+b)\exp\!\left(\frac{\alpha}{a+b} \right) - b \exp\!\left(\frac{\alpha}{b} \right) - a \right\}.
\end{eqnarray}
Moreover, its second-order derivative of $A_1$ w.r.t. $k$ is as follows:
\begin{equation}
    \frac{\dd^2}{\dd k^2}A_1 = -\frac{\beta |\bar h(e)|^2}{a}  \left[ Ei\!\left(\frac{\alpha}{a+b}\right) - Ei\!\left(\frac{\alpha}{b}\right) \right].
\end{equation}

\paragraph{For $N$-function: $\Phi_2(t) = \exp(t^2) - 1$.}

For $a > 0, b > 0$, we consider 
\begin{equation}
A_2 := \beta \int_{0}^{1} (at + b) \, \Phi_2\!\left(\frac{\alpha}{at + b}\right)\dd t.
\end{equation}
See Appendix~\S\ref{subsec:app:compute_A2} for the detailed computation of $A_2$ where its result is summarized as follow:
\begin{eqnarray}
    A_2 = \frac{\beta}{2a}\Bigg[ -\alpha^2 \left[ Ei\!\left(\frac{\alpha^2}{(a+b)^2}\right) - Ei\!\left(\frac{\alpha^2}{b^2}\right) \right] \hspace{14em} \nonumber \\
    + \left[(a+b)^2\exp\!\left(\frac{\alpha^2}{(a+b)^2} \right) - b^2 \exp\!\left(\frac{\alpha^2}{b^2} \right)\right] - \left[(a+b)^2 - b^2 \right] \Bigg].
\end{eqnarray}
Therefore, we get
\begin{equation}
    \mathcal{A}_{\Phi_2} = A_2,
\end{equation}
where $\beta = w_e, a = \frac{w_e}{\lambda(\G)}, b = 1 + \frac{\lambda(\gamma_e)}{\lambda(\G)}, \alpha = k|\bar h(e) |$.

Additionally, the first-order derivative of $A_2$ w.r.t. $k$ is as follows:
\begin{eqnarray}
    \frac{\dd A_2}{\dd k} = -\frac{\beta\alpha|\bar{h}(e)|}{a}\left[ Ei\!\left(\frac{\alpha^2}{(a+b)^2}\right) - Ei\!\left(\frac{\alpha^2}{b^2}\right) \right].
\end{eqnarray}
Moreover, its second-order derivative of $A_2$ w.r.t. $k$ is as follows:
\begin{eqnarray}
    \frac{\dd^2}{\dd k^2}A_2 = - \frac{\beta |\bar h(e)|^2}{a}  \left[ Ei\!\left(\frac{\alpha^2}{(a+b)^2}\right) \! - Ei\!\left(\frac{\alpha^2}{b^2}\right) \! + 2 \exp(\alpha^2/(a+b)^2) - 2 \exp(\alpha^2/b^2) \right].
\end{eqnarray}






\section{Further Experimental Details and Empirical Results}\label{sec:app:further_exp_details_empirical_results}

\subsection{Further Experimental Details}\label{subsec:app:implementation_notes_GSIM}

We summarize the number of pairs of probability measures for each dataset which is required to evaluate for kernelized SVM in Table~\ref{tb:numpairs}.

\paragraph{Implementation notes for $\gamma_e$ in GSI-M.}

\paragraph{$\bullet$ Preprocessing procedure for $\gamma_e$.} Following~\citep{le2025scalable}, we precompute $\gamma_e$ for all edge $e$ in the given graph $\G$. It only needs once for such preprocessing procedure since it does not depend on input measures, but only the graph structure itself. More concretely, we apply the Dijkstra algorithm to recompute the shortest paths from root node $z_0$ to all other input supports (or vertices) with complexity $\mathcal{O}(|E| + |V| \log{|V|})$ where we write $|\cdot|$ for the set cardinality. Then, we can evaluate $\gamma_e$ for each edge $e$ in $E$.

\paragraph{$\bullet$ Sparsity of $\gamma_e$.} As observed in~\citep{le2025scalable}, for any support of input measure $\mu$, its mass is contributed to $\mu(\gamma_e)$ if and only if $e \subseteq [z_0, x]$~\citep{le2022st}. Therefore, let $\text{supp}(\mu)$ be the set of supports of measure $\mu$, and define $E_{\mu, \nu} \subseteq E$ as follows
\[
E_{\mu, \nu} \hspace{-0.2em}:=\hspace{-0.2em} \left\{e \! \in \! E \mid \exists z \! \in \! (\text{supp}(\mu) \cup \text{supp}(\nu)), e \subseteq [z_0, z] \right\}.
\]
Additionally, note that $\Phi(0) = 0$ for all $N$-function $\Phi$. Then, in fact, we can remove all edges $e \in E \setminus E_{\mu, \nu}$ in the summation in Equation~\eqref{equ:Amemiya_RGSI_1d_optimization_discrete} for the univariate optimization problem for computing GSI-M.

\begin{table}[ht]
\caption{The number of pairs of probability measures on datasets for kernelized SVM.}
\label{tb:numpairs}
    \centering
\begin{tabular}{|l|r|}
\hline
Datasets & The number of pairs \\ \hline
\texttt{TWITTER}  & 4394432 \\ \hline
\texttt{RECIPE}   & 8687560      \\ \hline
\texttt{CLASSIC}  & 22890777       \\ \hline
\texttt{AMAZON}   & 29117200      \\ \hline
\texttt{Orbit}    & 11373250   \\ \hline
\texttt{MPEG7}    & 18130     \\ \hline
\end{tabular}
\end{table}

\paragraph{Notes on IPM and transport baselines.} We carry out the popular OW as baselines for our experiments. Since OW is notoriously computational heavy in practical applications, we employ generalized Sobolev transport (GST)~\citep{le2024generalized} which is a scalable alternative of OW in applications.

As discussed in~\S\ref{sec:experiments}, we emphasize that for the limit case $\Phi_0$, OW-$\Phi_0$ is equal to the standard optimal transport where its ground cost is the underlying graph metric in $\G$. Additionally, we also have GSI-M-$\Phi_0$ is equal to GST-$\Phi_0$, which is further equal to the $1$-order Sobolev transport~\citep{le2022st}.

Especially, we note that following Proposition~\ref{prop:RGSI_closed_form_discrete}, by taking the limit $p \to 1^+$, GSI-M-$\Phi_0$ has closed-form expression for fast computation. Furthermore, $\Phi_0$ is not a convex $N$-function due to its linear growth. By leveraging convex $N$-functions, one can adopt Orlicz geometric structure for Sobolev IPM, beyond $L^p$ paradigm.

\paragraph{Further discussion on computational complexity.} Following Theorem~\ref{thrm:RGSI_1d_optimization_discrete}, we simply need to solve the univariate optimization problem for the computation of GSI-M. In practice, we use a second-order solver (i.e., \emph{fmincon} with Trust Region Reflective solver) in our experiments.

As in the implementation notes discussed in this section, we apply a preprocessing for $\gamma_e$ with computational complexity $\mathcal{O}(|E| + |V| \log{|V|})$. It only needs once for such preprocessing procedure since it does not depend on input measures, but only the graph structure itself.

Furthermore, the computational complexity for each iteration is linear to the number of edges, i.e., $\mathcal{O}(|E|)$. Moreover, we exploit the sparsity of $\gamma_e$ to reduce the computational complexity into $\mathcal{O}(|E_{\mu, \nu}|)$ since we can remove all edges $e \in E \setminus E_{\mu, \nu}$ in the summation in Equation~\eqref{equ:Amemiya_RGSI_1d_optimization_discrete} for the computation of GSI-M between $\mu$ and $\nu$. Therefore, let $T$ be the number of iterations used in the \emph{fmincon} solver, then the computational complexity of GSI-M between measures $\mu$ and $\nu$ is $\mathcal{O}(T|E_{\mu, \nu}|)$. Empirically, we observe that the solver converges after about 40--50 iterations.

\subsection{Further Empirical Results}

We provide further results, corresponding to the empirical results in \S\ref{sec:experiments} for graph $\G_{\text{Sqrt}}$.

\begin{figure}[h]
  \begin{center}
    \includegraphics[width=0.5\textwidth]{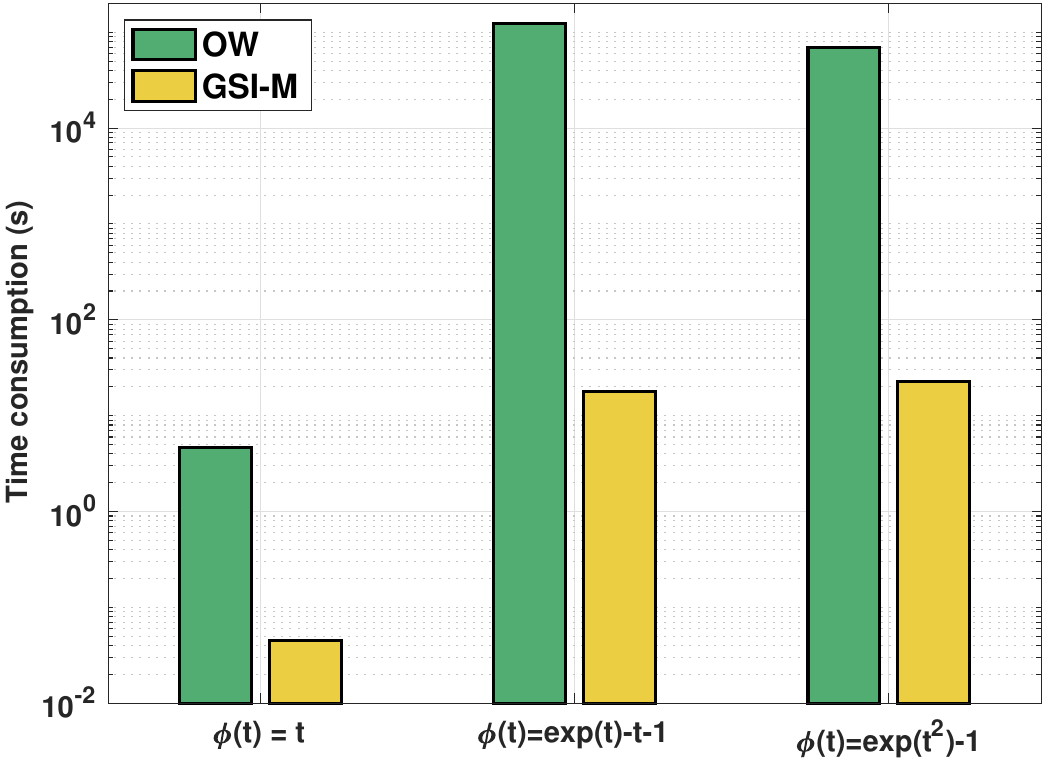}
  \end{center}
  \vspace{-6pt}
  \caption{Time consumption for GSI-M and OW on $\G_{\text{Sqrt}}$ with 1K nodes and about 32K edges.}
  \label{fg:Time_GSIM_GST_OW_10K_SLE}
\end{figure}

\paragraph{Computational comparison.} We illustrate corresponding results for computational comparison on graph $\G_{\text{Sqrt}}$ with 1K nodes and 32K edges in Figure~\ref{fg:Time_GSIM_GST_OW_10K_SLE}. The computation of GSI-M is also several-order faster than OW, and comparable to GST. More concretely, $\widehat \calGS$ is $100\times, 6800\times, 2900\times$ faster than OW for $\Phi_0, \Phi_1, \Phi_2$ respectively. For $N$-functions $\Phi_1$ and $\Phi_2$, GSI-M takes less than \emph{$23$ seconds} while OW needs at least \emph{$19$ hours}, and up to \emph{$34$ hours} for the computation.

\begin{figure}[h]
  \begin{center}
    \includegraphics[width=0.5\textwidth]{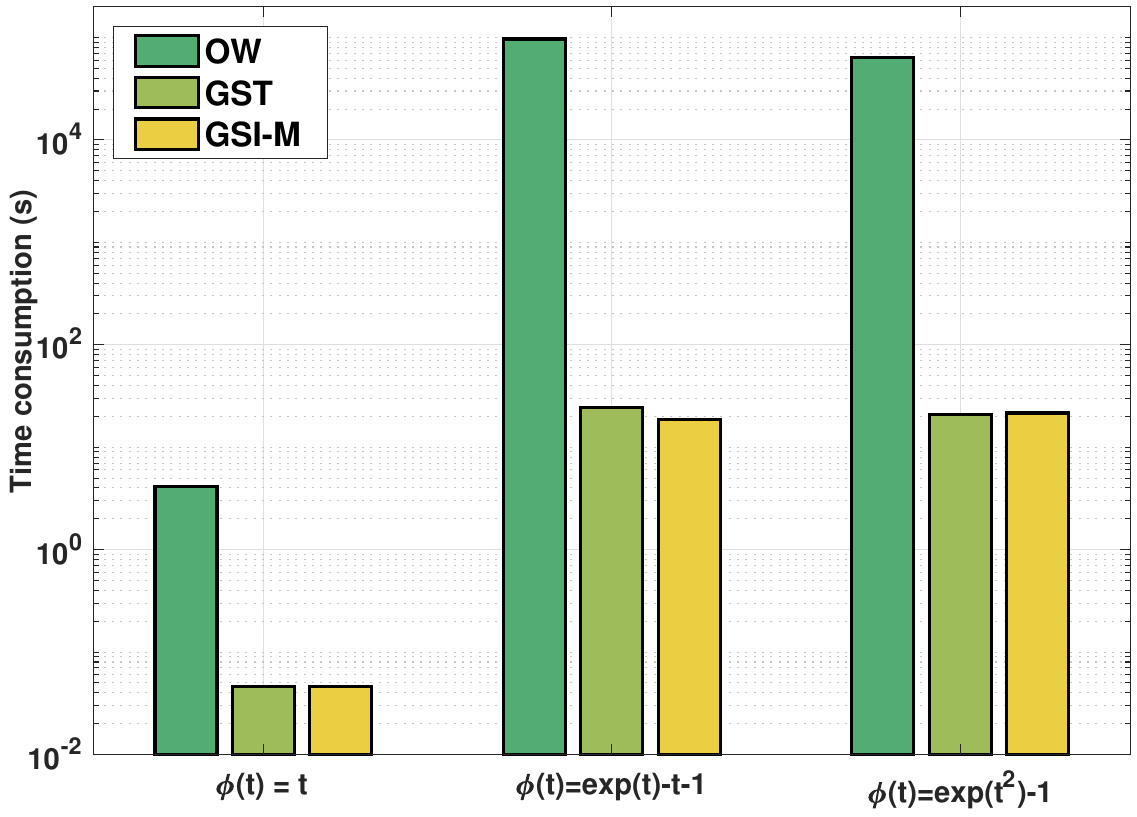}
  \end{center}
  \vspace{-6pt}
  \caption{Time consumption for GSI-M, GST and OW on $\G_{\text{Log}}$ with 1K nodes and about 7K edges.}
  \label{fg:Time_GSIM_GST_OW_10K_LLE_ALL3}
\end{figure}

\begin{figure}[h]
  \begin{center}
    \includegraphics[width=0.5\textwidth]{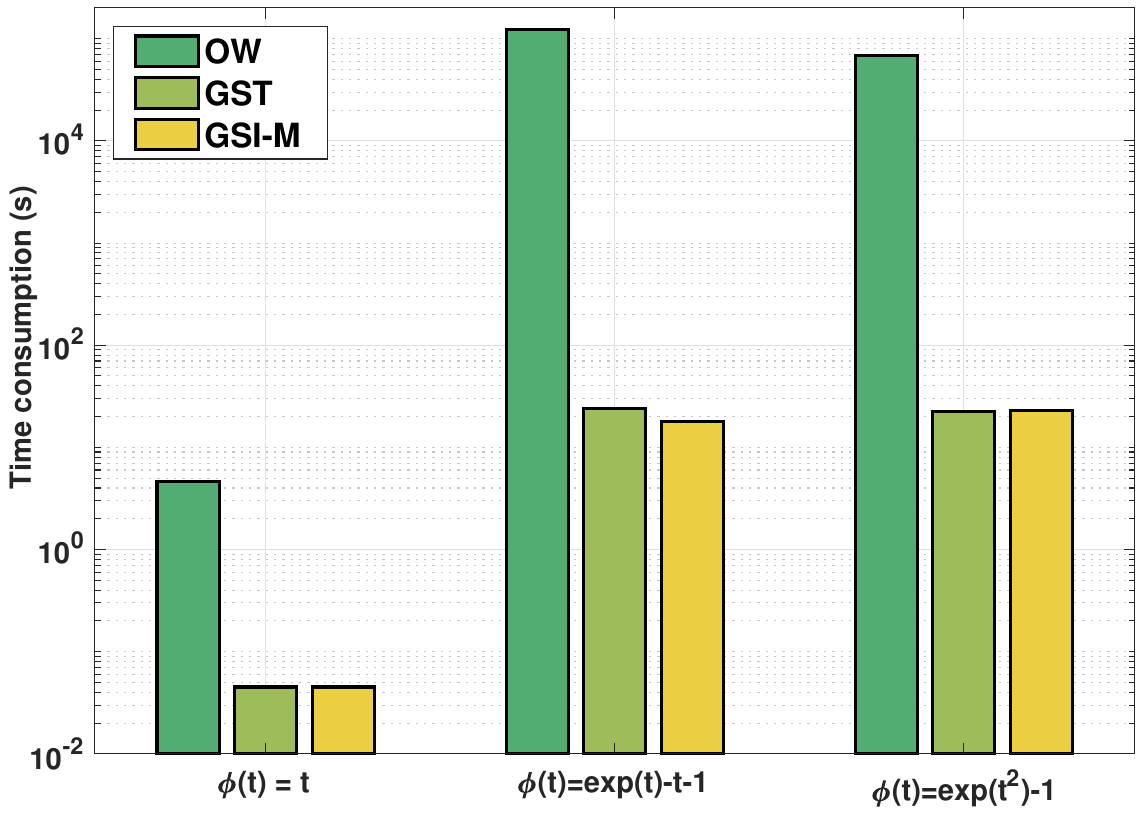}
  \end{center}
  \vspace{-6pt}
  \caption{Time consumption for GSI-M, GST and OW on $\G_{\text{Sqrt}}$ with 1K nodes and about 32K edges.}
  \label{fg:Time_GSIM_GST_OW_10K_SLE_ALL3}
\end{figure}

Additionally, we also include generalized Sobolev transport (GST)~\citep{le2024generalized}, which is a scalable alternative of OW in practical applications, in the computational comparison, and illustrated in Figures~\ref{fg:Time_GSIM_GST_OW_10K_LLE_ALL3} and~\ref{fg:Time_GSIM_GST_OW_10K_SLE_ALL3} for graphs $\G_{\text{Log}}$ and $\G_{\text{Sqrt}}$ respectively. We observe that the computations of the regularized GSI (GSI-M) are comparable to GST for all $\Phi$ functions.  

\begin{figure*}[h]
  \begin{center}
    \includegraphics[width=0.9\textwidth]{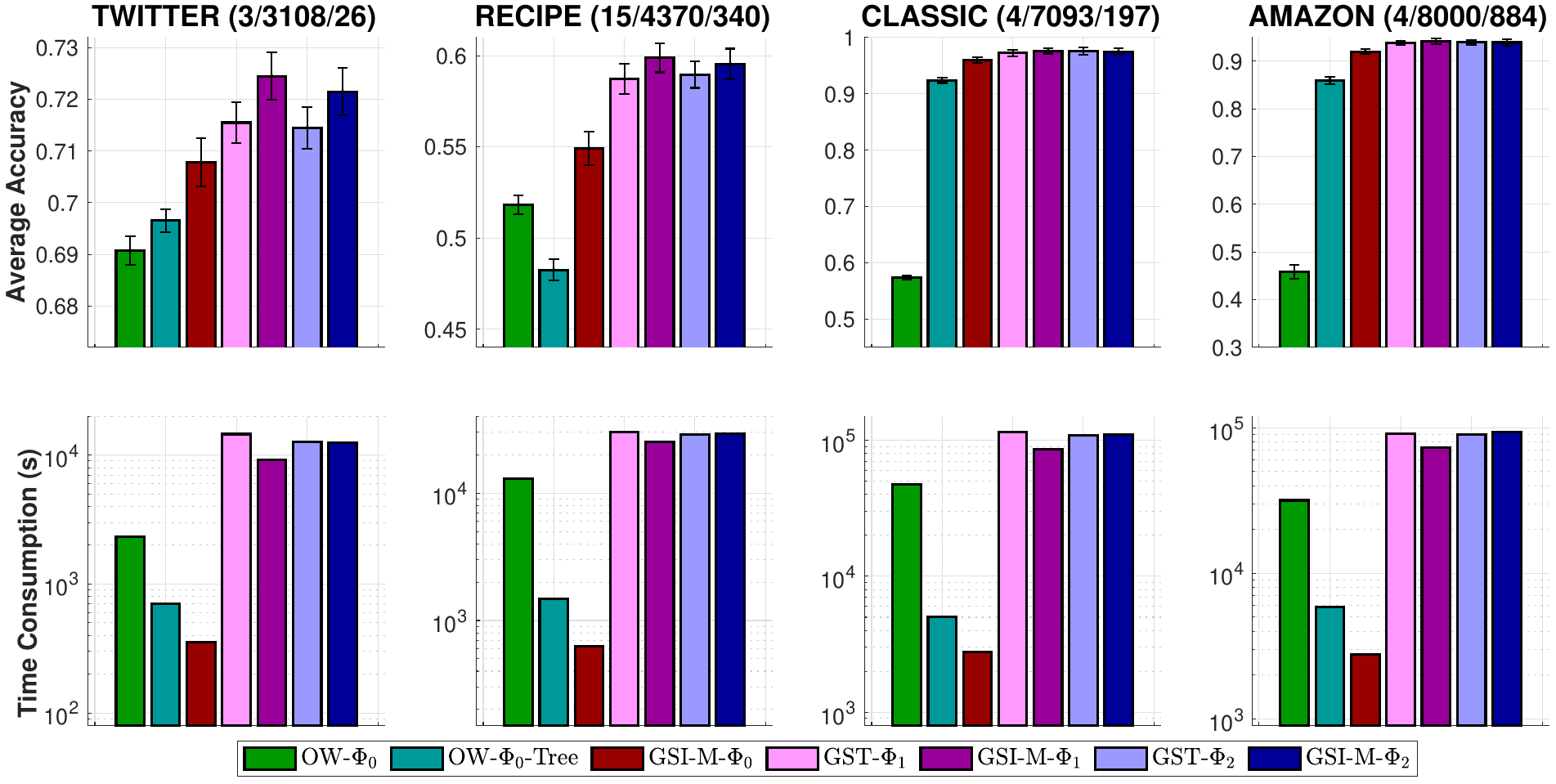}
  \end{center}
  \vspace{-10pt}
  \caption{SVM results and time consumption for kernel matrices with graph $\G_{\text{Sqrt}}$.}
  \label{fg:DOC_10KSqrt_app}
\end{figure*}

\paragraph{Document classification.} Figure~\ref{fg:DOC_10KSqrt_app} illustrates corresponding results for document classification on graph $\G_{\text{Sqrt}}$ with 10K nodes and about 1M edges.

\begin{figure}[h]
  \begin{center}
    \includegraphics[width=0.6\textwidth]{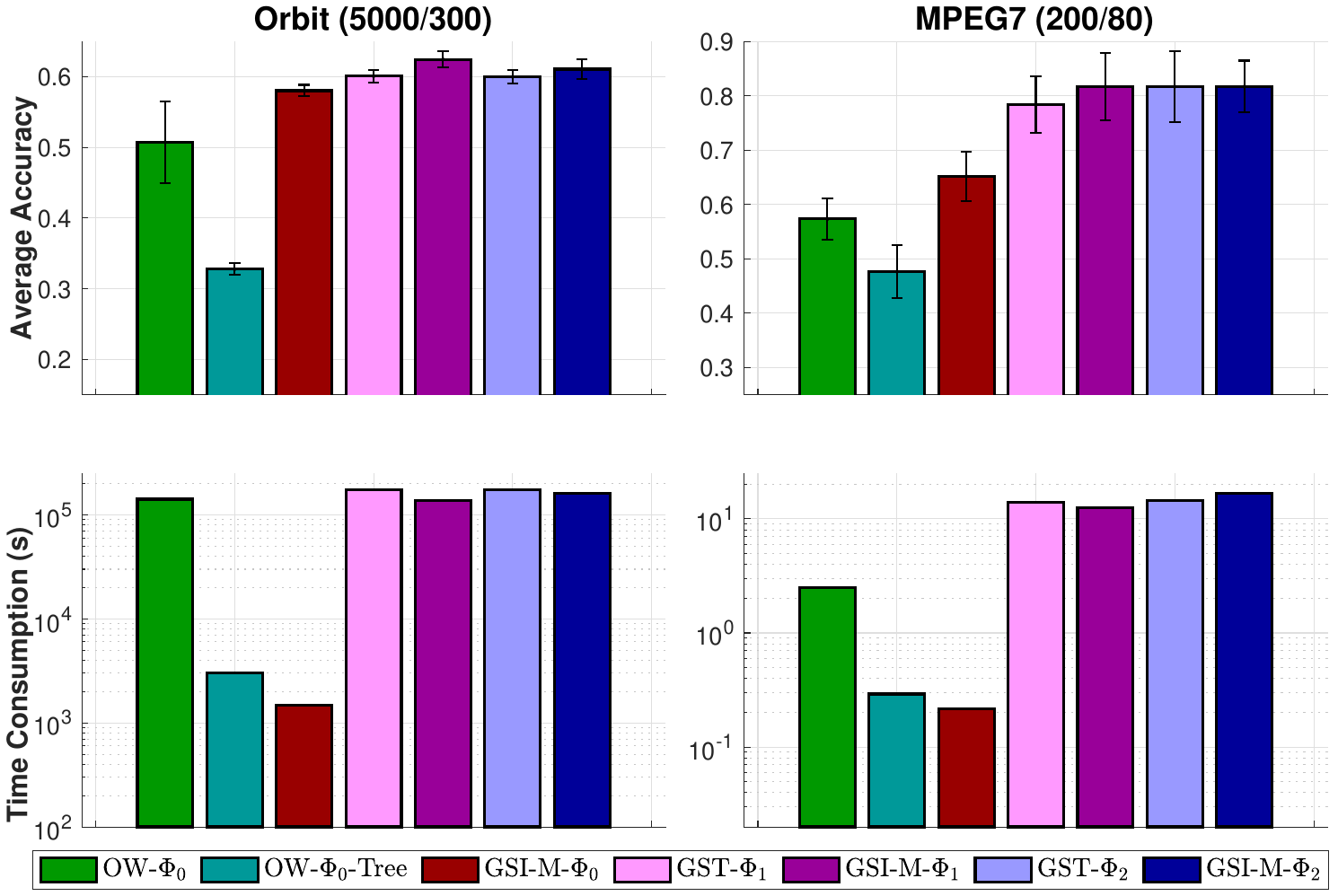}
  \end{center}
  \vspace{-6pt}
  \caption{SVM results and time consumption for kernel matrices with graph $\G_{\text{Sqrt}}$.}
  \label{fg:TDA_mix10KSqrt_app}
\end{figure}

\paragraph{TDA.} Figure~\ref{fg:TDA_mix10KSqrt_app} shows corresponding results for TDA on graph $\G_{\text{Sqrt}}$ with 10K nodes and about 1M edges.

\paragraph{Ablation study on different graphs and different $N$-functions.} Besides the graphs $\G_{\text{Log}}$(10K, 100K) and $\G_{\text{Sqrt}}$(10K, 1M) in our experiments where the numbers in the parenthesis, we run additional experiments on various graphs with different connectivities, and different number of nodes: $G_{\text{Log1K}}$(1K, 7K), $G_{\text{Sqrt1K}}$(1K, 32K), $G_{\text{Log100}}$(100, 500), and $G_{\text{Sqrt100}}$(100, 1K). Furthermore, following Proposition~\ref{prop:RGSI_closed_form_discrete} with $p = 2$, we consider an additional $N$-function $\Phi_3(t) = \frac{1}{4} t^2$, besides the two popular $\Phi$ functions $\Phi_1, \Phi_2$ in the literature, used in our experiments. We illustrate results for ablation study on different graphs and different $N$-functions for document classification on \texttt{TWITTER} dataset in Table~\ref{tb:ablation_study_graph_Nfunc_DOC} and for TDA on \texttt{MPEG7} dataset in Table~\ref{tb:ablation_study_graph_Nfunc_TDA}. We observe that for the same given graphs, performances of GSI-M-$\Phi_i$ ($i = 1, 2, 3$), consistently improve performances of GSI-M-$\Phi_0$.


\begin{table}
\caption{Ablation study on different graphs and different $N$-functions for document classification on \texttt{TWITTER} dataset.}
\label{tb:ablation_study_graph_Nfunc_DOC}
\vspace{+0.5 em}
\centering
\begin{tabular}{|l|c|c|c|c|}
\hline
                             & $\Phi_0$         & $\Phi_1$         & $\Phi_2$         & $\Phi_3$         \\ \hline
$\G_{\text{Log}}$(10K, 100K) & $70.55 \pm 0.41$ & $72.68 \pm 0.58$ & $71.85 \pm 0.61$ & $71.78 \pm 0.61$ \\ \hline
$\G_{\text{Sqrt}}$(10K, 1M)  & $70.78 \pm 0.47$ & $72.45 \pm 0.46$ & $72.15 \pm 0.46$ & $71.86 \pm 0.50$ \\ \hline
$\G_{\text{Log}}$(1K, 7K)    & $70.84 \pm 0.38$ & $72.52 \pm 0.36$ & $72.35 \pm 0.34$ & $72.03 \pm 0.40$ \\ \hline
$\G_{\text{Sqrt}}$(1K, 32K)  & $70.99 \pm 0.50$ & $72.25 \pm 0.42$ & $72.57 \pm 0.48$ & $72.05 \pm 0.44$ \\ \hline
$\G_{\text{Log}}$(100, 500)  & $70.05 \pm 0.33$ & $71.92 \pm 0.44$ & $71.36 \pm 0.42$ & $71.22 \pm 0.26$ \\ \hline
$\G_{\text{Sqrt}}$(100, 1K)  & $69.81 \pm 0.45$ & $71.71 \pm 0.30$ & $71.29 \pm 0.43$ & $70.96 \pm 0.43$ \\ \hline
\end{tabular}
\end{table}


\begin{table}
\caption{Ablation study on different graphs and different $N$-functions for TDA on \texttt{MPEG7} dataset.}
\label{tb:ablation_study_graph_Nfunc_TDA}
\vspace{+0.5 em}
\centering
\begin{tabular}{|l|c|c|c|c|}
\hline
                            & $\Phi_0$         & $\Phi_1$         & $\Phi_2$         & $\Phi_3$         \\ \hline
$\G_{\text{Log}}$(1K, 7K)   & $61.50 \pm 5.41$ & $75.00 \pm 4.16$ & $78.33 \pm 5.49$ & $72.67 \pm 4.66$ \\ \hline
$\G_{\text{Sqrt}}$(1K, 32K) & $65.17 \pm 4.54$ & $81.67 \pm 6.25$ & $81.67 \pm 4.78$ & $79.33 \pm 6.58$ \\ \hline
$\G_{\text{Log}}$(100, 500) & $68.17 \pm 5.00$ & $76.67 \pm 4.29$ & $75.67 \pm 5.44$ & $74.17 \pm 5.24$ \\ \hline
$\G_{\text{Sqrt}}$(100, 1K) & $72.17 \pm 5.67$ & $78.33 \pm 6.15$ & $77.67 \pm 6.35$ & $76.33 \pm 5.33$ \\ \hline
\end{tabular}
\end{table}

\paragraph{Ablation study on different root nodes.} In our experiments, we use cross validation to choose a root node from a random $10$-root-node subset of $V$ in the given graph $\G$. We further carry out additional experiments where the root node is randomly selected, and repeat it $50$ times. Tables|~\ref{tb:ablation_study_root_node_DOC} and~\ref{tb:ablation_study_root_node_TDA} illustrate ablation studies on different root nodes for document classification on \texttt{TWITTER} dataset, and for TDA on \texttt{MPEG7} dataset respectively. We observe that the root node may affect performances of GSI-M. The cross-validation can help to improve performances of GSI-M in our experiments.

\begin{table}
\caption{Ablation study on different root nodes for document classification on \texttt{TWITTER} dataset.}
\label{tb:ablation_study_root_node_DOC}
\vspace{+0.5 em}
\centering
\begin{tabular}{|l|c|c|}
\hline
                   & $\Phi_1$         & $\Phi_2$         \\ \hline
$\G_{\text{Log}}$  & $72.42 \pm 0.41$ & $71.53 \pm 0.44$ \\ \hline
$\G_{\text{Sqrt}}$ & $72.31 \pm 0.28$ & $72.04 \pm 0.25$ \\ \hline
\end{tabular}
\end{table}

\begin{table}
\caption{Ablation study on different root nodes for TDA on \texttt{MPEG7} dataset.}
\label{tb:ablation_study_root_node_TDA}
\vspace{+0.5 em}
\centering
\begin{tabular}{|l|c|c|}
\hline
                   & $\Phi_1$         & $\Phi_2$         \\ \hline
$\G_{\text{Log}}$  & $74.50 \pm 1.39$ & $77.23 \pm 2.02$ \\ \hline
$\G_{\text{Sqrt}}$ & $79.92 \pm 2.65$ & $80.03 \pm 2.61$ \\ \hline
\end{tabular}
\end{table}

\paragraph{For given graphs with a minor perturbation of edge weights.} We further consider a minor perturbation of edge weights on given graphs by adding random deviations in the range $(0, 0.1)$.
\begin{itemize}
\item In Tables~\ref{tb:ablation_study_perturbation_DOC_Log} and~\ref{tb:ablation_study_perturbation_DOC_Sqrt}, we illustrate results for document classification on \texttt{TWITTER} dataset with graphs $\G_{\text{Log}}$ and $\G_{\text{Sqrt}}$ with a minor perturbation of their edge weights respectively.

\item In Tables~\ref{tb:ablation_study_perturbation_TDA_Log} and~\ref{tb:ablation_study_perturbation_TDA_Sqrt}, we illustrate results for TDA on \texttt{MPEG7} dataset with graphs $\G_{\text{Log}}$ and $\G_{\text{Sqrt}}$ with a minor perturbation of their edge weights respectively.

\end{itemize}
We observe that the perturbation on edge weights slightly affects performances of GSI-M.


\begin{table}
\caption{Ablation study on graph $\G_{\text{Log}}$ with a minor perturbation of its edge weights for document classification on \texttt{TWITTER} dataset.}
\label{tb:ablation_study_perturbation_DOC_Log}
\vspace{+0.5 em}
\centering
\begin{tabular}{|l|c|c|}
\hline
                  & $\Phi_1$         & $\Phi_2$         \\ \hline
No perturbation   & $72.68 \pm 0.58$ & $71.85 \pm 0.61$ \\ \hline
With perturbation & $72.14 \pm 0.46$ & $71.42 \pm 0.39$ \\ \hline
\end{tabular}
\end{table}

\begin{table}
\caption{Ablation study on graph $\G_{\text{Sqrt}}$ with a minor perturbation of its edge weights for document classification on \texttt{TWITTER} dataset.}
\label{tb:ablation_study_perturbation_DOC_Sqrt}
\vspace{+0.5 em}
\centering
\begin{tabular}{|l|c|c|}
\hline
                  & $\Phi_1$         & $\Phi_2$         \\ \hline
No perturbation   & $72.45 \pm 0.46$ & $72.15 \pm 0.46$ \\ \hline
With perturbation & $71.95 \pm 0.44$ & $71.87 \pm 0.48$ \\ \hline
\end{tabular}
\end{table}


\begin{table}
\caption{Ablation study on graph $\G_{\text{Log}}$ with a minor perturbation of its edge weights for TDA on \texttt{MPEG7} dataset.}
\label{tb:ablation_study_perturbation_TDA_Log}
\vspace{+0.5 em}
\centering
\begin{tabular}{|l|c|c|}
\hline
                  & $\Phi_1$         & $\Phi_2$         \\ \hline
No perturbation   & $75.00 \pm 4.16$ & $78.33 \pm 5.49$ \\ \hline
With perturbation & $73.33 \pm 4.67$ & $76.17 \pm 3.93$ \\ \hline
\end{tabular}
\end{table}

\begin{table}
\caption{Ablation study on graph $\G_{\text{Sqrt}}$ with a minor perturbation of its edge weights for TDA on \texttt{MPEG7} dataset.}
\label{tb:ablation_study_perturbation_TDA_Sqrt}
\vspace{+0.5 em}
\centering
\begin{tabular}{|l|c|c|}
\hline
                  & $\Phi_1$         & $\Phi_2$         \\ \hline
No perturbation   & $81.67 \pm 6.25$ & $81.67 \pm 4.78$ \\ \hline
With perturbation & $81.33 \pm 6.31$ & $81.17 \pm 6.06$ \\ \hline
\end{tabular}
\end{table}

\paragraph{Empirical gaps between GSI-M and GST/OW.}

To our knowledge, for general non-tree graph structures, it is unknown whether there is any connection between GSI-M and OW or not. Their relation in Proposition~\ref{prop:Connection_GSIMPMR_OW} is only held when the given graph is a tree. For the connection between GSI-M and GST, their relation in Proposition~\ref{prop:Connection_GSIMPMR_GST} is held for any given general non-tree graph structures.

Additionally, we clarify that the goal of GSI-M is to provide an efficient algorithmic approach which yields an equivalent metric to the GSI problem. We agree that GSI-M leverages Orlicz geometry for the \emph{Sobolev IPM} problem similar to GST and OW which leverage Orlicz geometry for \emph{Sobolev transport} and \emph{optimal transport} respectively. Moreover, we also clarify that the goal of GSI-M is not to approximate either GST or OW.

Furthermore, with the setting in \S\ref{subsec:computation}, we compute the gap between GSI-M and GST, and the gap between GSI-M and OW, illustrated in Tables~\ref{tb:empirical_gap_GSIM_GST} and~\ref{tb:empirical_gap_GSIM_OW} respectively. We observe that the empirical gap between GSI-M and GST is relatively small, while the empirical gap between GSI-M and OW is relatively large.

\begin{table}
\caption{The empirical gap between GSI-M and GST.}
\label{tb:empirical_gap_GSIM_GST}
\vspace{+0.5 em}
\centering
\begin{tabular}{|l|c|c|}
\hline
                   & $\Phi_1$          & $\Phi_2$          \\ \hline
$\G_{\text{Log}}$  & $0.019 \pm 0.011$ & $0.026 \pm 0.015$ \\ \hline
$\G_{\text{Sqrt}}$ & $0.004 \pm 0.002$ & $0.005 \pm 0.002$ \\ \hline
\end{tabular}
\end{table}

\begin{table}
\caption{The empirical gap between GSI-M and OW.}
\label{tb:empirical_gap_GSIM_OW}
\vspace{+0.5 em}
\centering
\begin{tabular}{|l|c|c|}
\hline
                   & $\Phi_1$           & $\Phi_2$           \\ \hline
$\G_{\text{Log}}$  & $293.06 \pm 19.43$ & $403.47 \pm 26.75$ \\ \hline
$\G_{\text{Sqrt}}$ & $165.45 \pm 28.88$ & $227.79 \pm 39.75$ \\ \hline
\end{tabular}
\end{table}

\section{Further Discussions}\label{sec:app:further_discussions}

For completeness, we recall important discussions on the underlying graph in~\citep{le2022st, le2024generalized, le2025scalable}, since they are also applied and/or adapted to the proposed generalized Sobolev IPM with Musielak regularization.

\paragraph{Measures on a graph~\citep{le2025scalable}.} We reemphasize that in this work we consider the Sobolev IPM problem between \emph{two input probability measures} supported on the \emph{same} graph, which is also explored in~\citep{le2025scalable}. Such measures supported on a graph metric space are also considered in OT problem, explored in~\citep{le2022st, le2024generalized}. Our work generalizes Sobolev IPM, and we also derive a novel regularization for the generalized Sobolev IPM, which admits an efficient algorithmic approach (i.e., simply solving a univariate optimization problem for its computation).  

The generalized Sobolev IPM with Musielak regularization (GSI-M) is for \emph{input probability measures}, i.e., to compute distance between two probability measures, on the \emph{same} graph. We further distinguish the considered problem to the following related problems: 

\textbf{$\bullet$ Compute distance between two (different) input graphs.}~\citet{petric2019got, dong2020copt} compute OT problem between \emph{two input graphs}, where their goals are to compute distance between such two input graphs. Therefore, they are essentially different to our considered problem which computes distance between \emph{two input probability measures} supported on the \emph{same} graph.

\textbf{$\bullet$ Graph kernels between two (different) input graphs.} Graph kernels are functions between \emph{two input graphs} to measure their similarity. See~\citet{borgwardt2020graph} for a comprehensive review on graph kernels. Obviously, this research direction is different to our considered kernels for SVM which are built upon the GSI-M used for measuring similarity between \emph{two input probability measures} on the \emph{same} graph.

\paragraph{Applications with graph-based measures.} One of motivations of considering the graph-based measures in applications is from the manifold perspective for supports of measures. For such applications, one of popular approaches is to use graph structure as a representation for data manifolds (e.g., k-nearest-neighbor graphs). Additionally, some applications may have graph-based structure within themselves, e.g., one example is the popular transportation problem with a given map (regarded as a graph-structure over locations). A concrete example is to transport minerals from a distribution of mines to a distribution of factories, where locations of mines and factories are physically connected in a graph-based map, and the transportation (e.g., by vehicles) follows such a real physical map.

Additionally, please see~\citep{choromanski2026near} for further application settings of optimal transport for measures supported on a graph, and also Gaussian process on a given graph~\citep{zhang2026graph}, or graph field integrator~\citep{choromanski2023efficient}. Therefore, leveraging graph structure for machine learning is a promising research, not only for optimal transport and IPM, but also several other machine learning problems.

\paragraph{Path length for points in graph $\G$~\citep{le2022st}.} We can canonically measure a path length connecting any two points  $x, y \in \G$ where $x, y$ are not necessary to be nodes in $V$ of graph $\G$. 

Consider the edge $e= \langle u, v\rangle$ connecting two nodes $u, v \in V$, for $x, y \in \R^n$ and $x, y \in e$, we have 
\begin{eqnarray*}
& x = (1-s) u + s v, \\
& y = (1-t)u + t v,
\end{eqnarray*}
for some scalars $t,s\in [0,1]$. Therefore, the length of the path $[x, y]$ along edge $e$ (i.e., the line segment $\langle x, y\rangle$) is equal to $|t-s| w_e$. As a result, the length for an arbitrary path in $\G$ can be similarly defined by breaking down into pieces over edges and summing over their corresponding lengths~\citep{le2022st}.

\paragraph{Extension to measures supported on $\G$.} Similar to the regularized Sobolev IPM~\citep{le2025scalable}, the discrete case of the GSI-M in Equation~\eqref{equ:Amemiya_RGSI_1d_optimization_discrete} can be easily extended for measures with finite supports on $\G$ (i.e., supports of the input measures may not be nodes in $V$, but possibly points on edges in $E$) by using the same strategy to measure a path length for support data points in graph $\G$. More precisely, we break down edges containing supports into pieces and sum over their corresponding values instead of the sum over edges.

\paragraph{The assumption of uniqueness property of the shortest paths on $\G$.} As discussed in \citep{le2022st, le2024generalized, le2025scalable}, note that $w_e \in \R_{+}$ for any edge $e \in E$ in graph $\G$., it is almost surely that every node in $V$ can be regarded as unique-path root node since with a high probability, lengths of paths connecting any two nodes in graph $\G$ are different.

Additionally, for some special graph, e.g., a grid of nodes, there is \emph{no} unique-path root node for such graph. However, by perturbing each node, and/or perturbing lengths of edges if $\G$ is a non-physical graph, with a small deviation, we can obtain a graph satisfying the unique-path root node assumption.

Besides that, for input probability measures with full supports in graph $\G$, or at least full supports in any cycle in graph $\G$, then it exists a special support data point where there are multiple shortest paths from the root node to it. In this case, we simply choose one fixed shortest path among them for this support data point (or we can add a virtual edge from the root node to this support data point where the edge length is deducted by a small deviation). In many practical applications (e.g., document classification and TDA in our experiments), one can neglect this special case since input probability measures have a finite number of supports.

\paragraph{Recent advantages of Orlicz geometric structure.} In this work, we propose generalized Sobolev IPM (GSI) which leverages Orlicz geometric structure for Sobolev IPM for going beyond the $L^p$ paradigm, which is mainly motivated from recent advances and breakthroughs of Orlicz-Wasserstein (OW)~\citep{sturm2011generalized}.

As discussed in~\S\ref{sec:introduction}, the two breakthroughs results on OW~\citep{altschuler2023faster, GuhaHN23} illustrate that OW possesses unique implicit essences, which may not exist in the classic optimal transport (OT) with $L^p$
geometric structure cost.

\begin{itemize}
    \item \citet{altschuler2023faster} leverage OW to facilitate the development of differential privacy-inspired methodologies that address long-standing convergence challenges in hypocoercive differential equations.

    \item \citet{GuhaHN23} demonstrate that OW substantially enhances Bayesian contraction rates, effectively circumventing limitations inherent to traditional OT with Euclidean ground cost.
\end{itemize}

Additionally, we also summarize some recent advantages of Orlicz geometric structure as follows:

\begin{itemize}
    \item \citet{andoni2018subspace} and~\citet{song2019efficient} used Orlicz norm as a loss for regression to generalize both the least-square (squared $L^2$-norm loss) and least absolute deviation ($L^1$-norm loss) regressions, and also provide a scale-invariant version of $M$-estimator function loss (e.g., the Huber, Cauchy, Welsh, Tukey, Geman-McClure functions)

    \item \citet{andoni2018subspace} also illustrated that regression with Orlicz norm loss improves performances of linear regression with $L^1$/$L^2$ norms on Gaussian/sparse noise respectively.

    \item \citet{deng2022fast} leveraged Orlicz structure to scale up many machine learning problems including reinforcement learning, kernelized support vector machine, and clustering.

    \item \citet{chamakh2020orlicz} leveraged Orlicz structure to propose finite-sample uniform guarantee for random Fourier features which is almost surely convergence to approximate high-order derivatives for arbitrary kernels.

    \item \citet{lorenz2022orlicz} studied Orlicz norm as a regularization for OT problem with continuous probability measures.

    \item \citet{le2024generalized} illustrated that Orlicz geometric structure is useful for the generalize Sobolev transport in applications, including document classification, orbit recognition, image classification. These applications are also included for our experimental setups (\S\ref{sec:experiments}).

\end{itemize}

Advantages of Orlicz geometric structures in applications are a promising on-going research directions. For comprehensive studies of Orlicz geometry, see~\citep{adams2003sobolev, rao1991theory}.

\paragraph{The generalized Sobolev IPM with Musielak regularization (GSI-M).} Similar to regularized Sobolev IPM~\citep{le2025scalable}, we assume that the graph metric space is given. The question of adaptively learning an optimal graph metric structure from given data is left for future work for further investigation.

\paragraph{Further computational discussion on Sobolev IPM, generalized Sobolev IPM, and regularization approaches.}

Additionally, we clarify that for the original Sobolev IPM, and original generalized Sobolev IPM (GSI), to our knowledge, we have not been aware of any algorithmic approaches to compute them efficiently yet (i.e., without regularization).

The regularized Sobolev IPM has a closed-form expression for computation~\citep[Theorem 3.5]{le2025scalable}. Consequently, there is no need for any optimization in the computation of the regularized Sobolev IPM.

Our proposed GSI-M is an efficient approach which yields an equivalent metric to GSI only. We agree that it is interesting to consider whether it is possible to derive efficient algorithmic approaches for the original Sobolev IPM and original GSI in future work as described in \S\ref{sec:conclusion}.

Following Theorem~\ref{equ:Amemiya_RGSI_1d_optimization_discrete}, we simply need to solve the univariate optimization problem for the computation of GSI-M. In practice, we use a second-order solver (i.e., \emph{fmincon} with Trust Region Reflective solver) in our experiments 

\paragraph{The graphs $\G_{\text{Log}}$ and $\G_{\text{Sqrt}}$~\citep{le2022st, le2025scalable}.} For an efficient and fast computation, we apply the farthest-point clustering method to cluster supports of measures into at most $M$ clusters.\footnote{$M$ is the input number of clusters for the clustering method. Consequently, the clustering result has at most $M$ clusters, depending on input data.} Then, let the set of vertices $V$ be the set of centroids of these clusters, i.e., graph vertices. For edges, in graph $\G_{\text{Log}}$, we randomly choose $(M\log{M})$ edges; and $M^{3/2}$ edges for graph $\G_{\text{Sqrt}}$. We further denote the set of those randomly sampled edges as $\tilde{E}$.  

For each edge $e$, its corresponding edge length (i.e., edge weight) $w_e$ is computed by the Euclidean distance between the two corresponding nodes of edge $e$. Let $n_c$ be the number of connected components in the graph $\tilde{\G}(V, \tilde{E})$. Then, we randomly add $(n_c - 1)$ more edges between these $n_c$ connected components to construct a connected graph $\G$ from $\tilde{\G}$. Let $E_c$ be the set of these $(n_c - 1)$ added edges and denote set $E = \tilde{E} \cup E_c$, then $\G(V, E)$ is the constructed graph. 

$\bullet$ For document classification, following~\citep{le2025scalable}, we use word2vec to map words into vectors in~$\R^{300}$, and apply the above procedures for the word embeddings to construct graphs $\G_{\text{Log}}$ and $\G_{\text{Sqrt}}$. Therefore, we regard each document as a probability distribution of words supported on these given graphs.

$\bullet$ For TDA, we use the above procedure for topological features, which are $2$-dimensional data points summarizing its life span. Thus, we regard each PD as a probability measure of topological features supported on these given graphs.

\paragraph{Datasets.} For the datasets in our experiments (i.e., \texttt{TWITTER, RECIPE, CLASSIC, AMAZON} for document datasets, and \texttt{Orbit, MPEG7} for TDA), one can contact the authors of Sobolev transport~\citep{le2022st} to access to them. 

\paragraph{Computational devices.} We run all of our experiments on commodity hardware.

\paragraph{Hyperparamter validation.} We use the same strategy as in~\citep{le2025scalable}. For validation, we further randomly split \emph{the training set} into $70\%/30\%$ for validation-training and validation with $10$ repeats to choose hyper-parameters in experiments.

\paragraph{Further discussion on hyperparameters.} The performance of the generalized Sobolev IPM with Musielak regularization (GSI-M) basically depends on the choice of the $N$-function $\Phi$, which is much similar to how kernel functions impact performance in kernel-dependent machine learning framework. In our experiments with $N$-functions $\Phi_1$ and $\Phi_2$, performances with $N$-function $\Phi_1$ is slightly more favorable than those with $N$-function $\Phi_2$, except in \texttt{MPEG7} dataset with graph $\G_{\text{Sqrt}}$.

Determining the optimal $N$-function $\Phi$ for the generalized Sobolev IPM with Musielak regularization (GSI-M) in a given task is an open problem that warrants further investigation. We leave it for future work. As an interim solution, cross-validation can be used to select $N$-function $\Phi$ from a set of candidate functions.

\paragraph{The number of pairs in training and test for kernelized SVM~\citep{le2025scalable}.} Let $N_{tr}, N_{te}$ be the number of measures for training and test respectively. For the kernelized SVM training, the number of pairs which we need to evaluate distances is $(N_{tr}-1) \times \frac{N_{tr}}{2}$. On the test phase, the number of pairs which we need to evaluate distances is $N_{tr} \times N_{te}$. Therefore, for each run, the number of pairs which we require to evaluate distances for both training and test is totally $N_{tr} \times (\frac{N_{tr}-1}{2} +  N_{te})$. See Table~\ref{tb:numpairs} for the number of pairs we need to evaluate distances for kernelized SVM in experiments.


\paragraph{Graph structure construction.}
In our experiments, we consider the given graphs, which are constructed from supports of all measures in the training set. Additionally, one may consider the tree system for the generalized Sobolev IPM (GSI) problem which may by-pass a bias of graph structure construction from finite data points due to the unboundedness of the tree system, i.e., with an infinite number of nodes.

\paragraph{Limitations and discussions.} We study Sobolev IPM for probability measures supported on a \emph{given} graph metric, as in~\citep{le2022st, le2024generalized, le2025scalable, choromanski2026near}. The approach of graph-based probability measures extends those involving the tree-based measures and 1d-projected measures, since graph structure provides more flexibility and degrees of freedom compared to the tree structure and 1d-line structure (also see~\citep{le2022st}). Some examples using tree-structure and 1d-line structure include tree-sliced Wasserstein~\citep{LYFC, tran2025spherical, tran2025distancebased, tran2025treesliced, tran2025nonlinear, tran2026revisiting}, tree-sliced entropy partial transport (EPT)~\citep{le2021ept, tran2026treeEPT}, tree-sliced Sobolev IPM~\citep{tran2026treesliced}, and the popular sliced-Wasserstein~\citep{rabin2011wasserstein, bonneel2015sliced, nguyen2024sliced}. As in discussed in \S\ref{sec:introduction},~\citep{tran2026treesliced} leveraged tree system~\citep{tran2025spherical, tran2025distancebased, tran2025treesliced, tran2025nonlinear, tran2026revisiting} and proposed efficient Radon transforms to extend finding results for graph-based measures (e.g., regularized Sobolev IPM~\citep{le2025scalable}) to applications with probability measures on the popular Euclidean spaces and hyperspheres. 

Therefore, the tree system framework provides a mechanism to bridge finding results for graph-based measures and applications with probability measures on the popular Euclidean spaces and hyperspheres. Since we focus on Sobolev IPM for graph-based measures in this work, we leave this application research direction as future works for in-depth investigations.




\section{Brief Reviews}\label{sec:app:brief_reviews}

In this section, to ease the readers, we give a brief review for related notions which we have used in the development of our proposal. For completeness, we also summarize some notions reviewed in~\citep{le2024generalized, le2025scalable}.

\subsection{Graph Illustration}

\begin{figure}
  \begin{center}
    \includegraphics[width=0.3\textwidth]{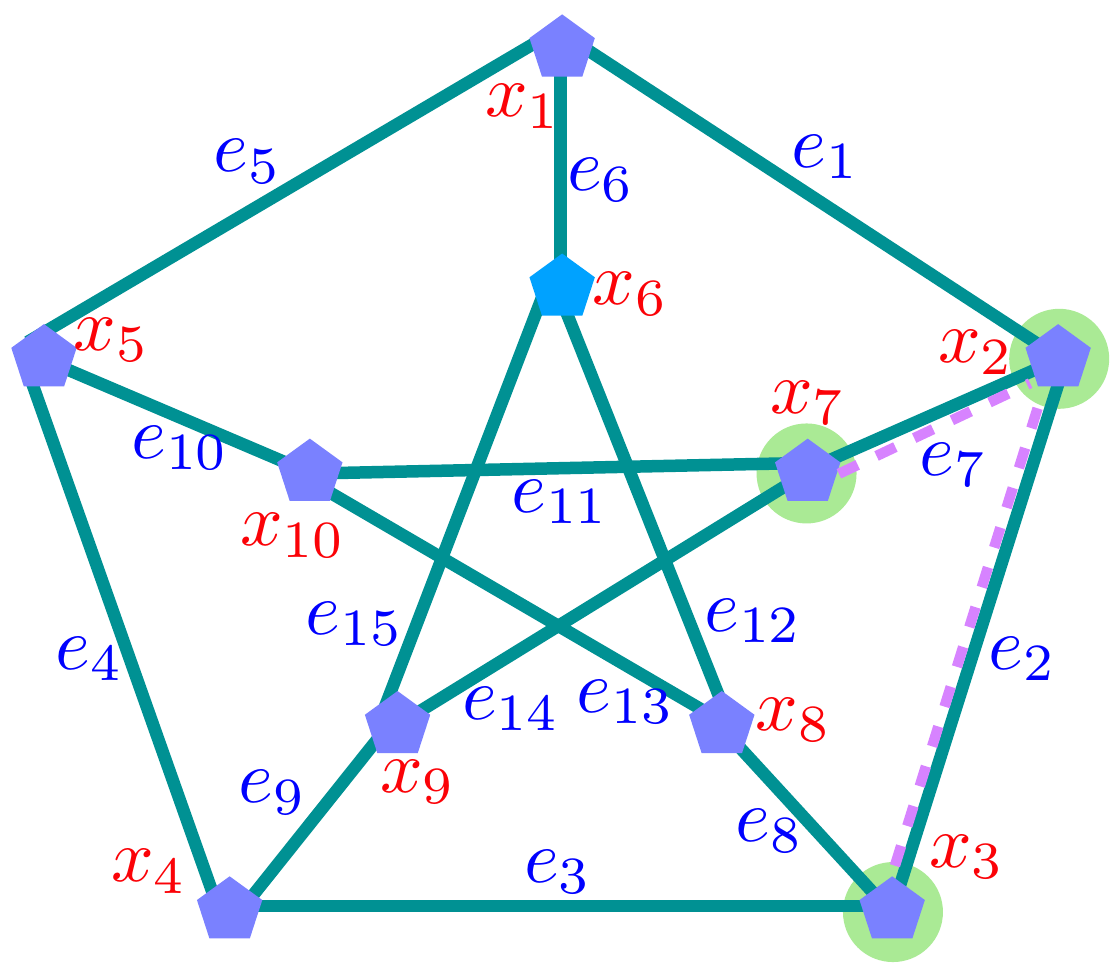}
  \end{center}
  \vspace{-6pt}
  \caption{A geodetic graph with $10$ nodes $\left\{x_1, x_2, \dotsc, x_{10}\right\}$ and $15$ edges $\left\{e_1, e_2, \dotsc, e_{15}\right\}$, and each edge length equals to $1$, i.e., $w_{e_j}=1, \forall j$. For any $x_i, x_j$, there is a unique shortest path between them, with a length $2$. Therefore, it satisfies the uniqueness property of the shortest paths. Let $x_1$ be the unique-path root node (i.e., $z_0 = x_1$), and subgraph $\widetilde{\G}$ containing $3$ nodes $\left\{x_2, x_3, x_{7}\right\}$ and $2$ edges $\left\{e_{2}, e_{7}\right\}$, then we have $\Lambda(x_2) = \gamma(e_1) = \widetilde{\G}$.}
  \label{fg:GeodeticGraph}
\end{figure}

We illustrate graph notions, reviewed in \S\ref{sec:preliminaries}, in Figure~\ref{fg:GeodeticGraph}.

\subsection{A Review on Functional Spaces}\label{app:sec:FuncSpaces}

We describe a short review on the $L^p$ space, the weighted $L^p$ space.

\paragraph{$L^{p}$ space.} For a nonnegative Borel measure $\lambda$ on $\G$, let $L^p( \G, \lambda)$ be the space of all Borel measurable functions $f:\G\to \R$ s.t. $\int_\G |f(y)|^p \lambda(\mathrm{d}y) <\infty$. For $p=\infty$, we instead assume that $f$ is bounded $\lambda$-a.e.
Then, $L^p( \G, \lambda)$ is a normed space with the norm being defined as follows:
\[
\|f\|_{L^p} := \left(\int_\G |f(y)|^p \lambda(\dd y)\right)^\frac{1}{p} \text{ for } 1\leq p < \infty.
\]
On the other hand, for $p = \infty$, then we have
\[
\|f\|_{L^{\infty}} := \inf\left\{t \in \R:\, |f(x)|\leq t \mbox{ for $\lambda$-a.e. } x\in\G\right\}.
\]
Additionally, functions $f_1, f_2 \in L^p( \G, \lambda)$ are considered to be the same if $f_1(x) =f_2(x)$ for $\lambda$-a.e. $x\in\G$.

\paragraph{${L^{p}_{\hat{w}}}$ space.} For a nonnegative Borel measure $\lambda$ on $\G$, and a positive weight function $\hat{w}$ on $G$, let $L^p_{\hat{w}}( \G, \lambda)$ be the space of all Borel measurable functions $f:\G\to \R$ s.t. $\int_\G \hat{w}(x) |f(x)|^p \lambda(\mathrm{d}x) <\infty$. For $p=\infty$, we instead assume that $f$ is bounded $\hat{w}\lambda$-a.e. Then, $L^p_{\hat{w}}( \G, \lambda)$ is a normed space with the norm being defined as follows:
\[
\|f\|_{L^p_{\hat{w}}} := \left(\int_\G \hat{w}(x) |f(x)|^p \lambda(\dd x)\right)^\frac{1}{p} \text{ for } 1\leq p < \infty.
\]
For the case $p = \infty$, as $\hat{w}(x) > 0$ for every $x \in \G$, we have
\begin{align*}
\|f\|_{L^\infty_{\hat{w}}} 
&:= \inf\left\{t \in \R:\, |f(x)|\leq t \mbox{ for $(\hat{w}\lambda)$-a.e. } x\in\G\right\}\\
&=\inf\left\{t \in \R:\, |f(x)|\leq t \mbox{ for $\lambda$-a.e. } x\in\G\right\}\\
&=\|f\|_{L^{\infty}} .
\end{align*}


\subsection{A Review on Sobolev Transport}\label{app:subsec:review-ST}

In this section, we provide a brief review on the Sobolev transport (ST)~\citep{le2022st} for graph-based measures, and the length measure on a graph.

\begin{definition}[Graph-based Sobolev space~\citep{le2022st}]\label{def:Sobolev}
 Let $\lambda$ be a nonnegative Borel measure on $\G$, and let  $1\leq p\leq \infty$. A continuous function $f: \G \to \R$ is said to belong to the Sobolev space $W^{1,p}(\G, \lambda)$ if there exists a  function $h\in L^p( \G, \lambda) $ satisfying 
\begin{equation}\label{FTC}
f(x) -f(z_0) =\int_{[z_0,x]} h(y) \, \lambda(\mathrm{d}y)  \quad \forall x\in \G.
\end{equation}
Such function  $h$ is unique in $L^p(\G, \lambda) $ and is called the graph derivative of $f$ w.r.t.~the measure $\lambda$. The graph derivative of $f \in W^{1,p}(\G, \lambda)$ is denoted as $f' \in L^p( \G, \lambda)$.
\end{definition}

\textbf{Sobolev transport (ST)~\citep{le2022st}.} For probability measures $\mu, \nu \in \calP(\G)$, and $1 \le p \le \infty$, the $p$-order Sobolev transport (ST)~\citep[Definition 3.2]{le2022st} is defined as 
\begin{equation} \label{eq:distance}
\mathcal{ST}_p(\mu,\nu ) \coloneqq  \left\{
\begin{array}{cl}
 \sup \Big[\int_\G f(x) \mu(\mathrm{d}x) - \int_\G f(x) \nu(\mathrm{d}x)\Big] \\
 \hspace{0.8em} \mathrm{s.t.} \, f  \in W^{1,p'} \hspace{-0.3em} (\G, \lambda),  \, \|f'\|_{L^{p'} (\G, \lambda)}\leq 1,
\end{array}
\right.
\end{equation}
where we write $f'$ for the generalized graph derivative of $f$, and $W^{1,p'} \hspace{-0.3em} (\G, \lambda)$ for the graph-based Sobolev space on $\G$.

ST is a scalable variant of optimal transport (OT) for probability measures supported on a graph (i.e., not OT itself) since it relaxes the Lipschitz constriant for the critic function as in $1$-order Wasserstein by considering this Lipschitz constraint in a Sobolev space. Additionally,~\citet[Corollary 4.3]{le2022st} showed that $1$-order Sobolev transport is identical to $1$-order Wasserstein \emph{when the graph is a tree}. Furthermore,~\citet[\S 4.1]{le2022st} admitted that the exact relationship between $p$-order Sobolev transport and $p$-order Wasserstein when $p > 1$ is \emph{unknown}.

Furthermore, for input measures possibly having different total mass, ST can be extended into its unbalanced variant, e.g., see~\citep{le2023scalable} for rigorous details of the unbalanced Sobolev transport. 

\begin{proposition}[Closed-form expression of Sobolev transport~\citep{le2022st}]\label{app:prop:closed-form_ST}
Let $\lambda$ be any nonnegative Borel measure on $\G$, and let  $1\leq p < \infty$. Then, we have 
\[
\mathcal{ST}_{p}(\mu,\nu )
= \left( \int_{\G} | \mu(\Lambda(x)) -  \nu(\Lambda(x))|^p \, \lambda(\mathrm{d}x) \right)^{\frac{1}{p}},
\]
where $\Lambda(x)$ is the subset of $\G$  defined by Equation~\eqref{sub-graph}. 
\end{proposition}

\begin{definition}[Length measure~\citep{le2022st}] \label{def:measure} 
Let $ \lambda^*$ be the unique Borel measure on $\G$ s.t. the restriction of $\lambda^*$ on any edge is the length measure of that edge. That is, $\lambda^*$  satisfies:
\begin{enumerate}
\item[i)] For  any edge $e$ connecting two nodes $u$ and $v$, we have 
 $\lambda^*(\langle x,y\rangle) = (t-s) w_e$ 
 whenever $x = (1-s) u + s v$ and $y = (1-t)u + t v$ for $s,t \in [0,1)$ with $s \leq t$. Here, recall that $\langle x,y\rangle$ is the line segment in $e$ connecting $x$ and $y$.
 \item[ii)] For any Borel set $F \subset \G$, we have
 \[
 \lambda^*(F) = \sum_{e\in E} \lambda^*(F\cap e).
 \]
\end{enumerate}
\end{definition}

\begin{lemma}[$\lambda^*$ is the length measure on graph~\citep{le2022st}] \label{lem:length-measure}
Suppose that $\G$ has no short cuts, namely, any edge $e$ is a shortest path connecting its two end-points. Then, $\lambda^*$ is a length measure in the sense that
\[
\lambda^*([x,y]) = d_\G(x,y)
\]
for  any  shortest path   $[x,y]$ connecting $x, y$. Particularly, $\lambda^*$ has no atom in the sense that $\lambda^*(\{x\})=0$ for every $x \in \G$. 
\end{lemma}

\subsection{A Review on Sobolev IPM and Its Scalable Regularized Approach}\label{app:sec:review_SIPM}

We give a brief review on the Sobolev norm~\citep{adams2003sobolev}, Sobolev IPM, and its scalable regularized approach~\citep{le2025scalable} for graph-based measures.

\paragraph{Sobolev norm.} $W^{1,p}(\G, \lambda)$ is a normed space (reviewed in \S\ref{app:subsec:review-ST}), with the Sobolev norm~\citep[\S3.1]{adams2003sobolev} being defined as
\begin{equation}\label{eq:SobolevNorm}
    \norm{f}_{W^{1,p}} := \left( \norm{f}_{L^p}^p + \norm{f'}_{L^p}^p \right)^{\frac{1}{p}}.
\end{equation}
Additionally, let $W^{1, p}_0(\G, \lambda)$ be the subspace consisting of all functions $f$ in $W^{1, p}(\G, \lambda)$ satisfying $f(z_0) = 0$. Denote $\calB_p := \left\{ f \in W^{1, p}_0(\G, \lambda): \norm{f}_{W^{1, p}} \le 1 \right\}$ as the unit ball in the Sobolev space.

\paragraph{Sobolev IPM.} Sobolev IPM for graph-based measures is an instance of the IPM where its critic function belongs to the graph-based Sobolev space, and is constrained within the unit ball of that space. More concretely, given a nonnegative Borel measure $\lambda$ on $\G$, an exponent
$1 \le p \le \infty$ and its conjugate $p'$, the Sobolev IPM between measures $\mu, \nu \in \calP(\G)$ is defined as follows:
\begin{equation}\label{eq:SobolevIPM_app}
\calS_{p}(\mu, \nu) := \hspace{-0.4em} \sup_{f \in \calB_{p'}} \hspace{-0.2em} \left| \int_{\G} f(x) \mu(\dd x) - \int_{\G} f(y) \nu(\dd y)\right|.
\end{equation}

Note that one should distinguish Sobolev IPM problem from Sobolev transport~\citep{le2022st}, generalized Sobolev transport (GST)~\citep{le2024generalized}, and Wasserstein problems. In fact, for Sobolev IPM, the critic function is constrained within a unit ball of Sobolev norm, which involves both critic function and its gradient while Sobolev transport, GST and Wasserstein only constraint on gradient of critic function (i.e., Lipschitz constraint).

\paragraph{Scalable regularized Sobolev IPM~\citep{le2025scalable}.} For weight function $\hat{w}_{\hat \calS}(x) = 1 + \lambda(\Lambda(x))$, for all $x \in \G$. Let $\calB(p', \hat{w}_{\hat \calS})$ be defined as follows:
\begin{equation}\label{eq:weightedLp_ball_regularized}
\calB(p', \hat{w}_{\hat \calS}) := \left\{f \in W^{1, p'}_0 \! (\G, \lambda) : \norm{f'}_{L^{p'}_{\hat{w}_{\hat \calS}}} \leq 1 \right\}.
\end{equation}
Then, the regularized Sobolev IPM is defined as
\begin{definition}[Regularized Sobolev IPM on graph~\citep{le2025scalable}]\label{def:regSobolevIPM_app}
Let $\lambda$ be a nonnegative Borel measure on $\G$ and $1 \le p \le \infty$.  Then for any given probability measures $\mu, \nu \in \calP(\G)$, the regularized Sobolev IPM is defined as
\begin{equation}\label{eq:regSobolevIPM}
\hat{\calS}_{p}(\mu, \nu) \hspace{-0.1em} := \hspace{-1.3em} \sup_{f \in \calB(p', \hat{w}_{\hat \calS})} \hspace{-0.2em} \left| \int_{\G} f(x) \mu(\text{d}x) - \int_{\G} f(y) \nu(\text{d}y)\right|.\hspace{-0.25em}
\end{equation}
\end{definition}

Note that it is \emph{unknown} whether the regularization approach in~\citet{le2025scalable} with weight function $\hat{w}_{\hat \calS}$ can be extended for general cases beyond the $L^p$ geometric structure, e.g., Orlicz geometric structure.

\subsection{A Review on IPM and Wasserstein Distance}\label{app:subsec:review-IPM-OT}

We provide a short review on IPM and Wasserstein distance for probability measures.

\paragraph{IPM.} Integral probability metrics (IPM) for probability measures $\mu, \nu$ are defined as follows:

\begin{equation}
    \gamma_{\calF}(\mu, \nu) = \sup_{f \in \calF} \left| \int_{\lambda} f(x) \mu(\text{d}x) - \int_{\lambda} f(y) \nu(\text{d}y)\right|.
\end{equation}

\paragraph{Special case: $1$-Wasserstein distance (dual formulation).} The $1$-Wasserstein distance is a special case of IPM. In particular, for $\calF = \calF_W := \left\{f : \left| f(x) - f(y) \right| \le d_{\G}(x, y) \right\}$ where $d_{\G}$ is the graph metric on graph $\G$, then IPM is equal to the $1$-Wasserstein distance with ground cost $d_{\G}$
\begin{equation}
    \calW(\mu, \nu) = \sup_{f \in \calF_W} \left| \int_{\G} f(x) \mu(\text{d}x) - \int_{\G} f(y) \nu(\text{d}y)\right|.
\end{equation}


\paragraph{$p$-Wasserstein distance (primal formulation).}
Let $1\leq p <\infty$, for probability measures $\mu$ and $\nu$ on $\G$, then, the $p$-Wasserstein distance is defined as follows:
\begin{align*}
\calW_p(\mu,\nu)^p 
&= \inf_{\pi \in \Pi(\mu,\nu)}\int_{\G\times\G} d_\G(x,y)^p \pi(\dd x, \dd y),
\end{align*}
where $\Pi(\mu,\nu) := \Big\{ \pi \in \calP(\G \times \G): \, \pi_1= \mu, \, \pi_2= \nu \Big\}$; $\pi_1, \pi_2$ are the first and second marginals of $\pi$ respectively.

We would like to remark that to our knowledge, there is \emph{no} closed-form expression for optimal transport (OT) problem for probability measures on a graph in general. However, when the graph is a tree, OT admits closed-form expression, a.k.a., tree-Wasserstein~\citep{LYFC}.




\subsection{Orlicz functions}\label{appsubsec:Orlicz_functions}

We provide a brief review on Orlicz functions as summarized in~\citep{le2024generalized} for completeness. For comprehensive studies on Orlicz functions, see~\citep{adams2003sobolev, rao1991theory}.


\paragraph{Popular examples of $N$-functions.} Some popular examples for $N$-functions~\citep[\S8.2]{adams2003sobolev} in the literature are as follows: 
\begin{enumerate}
\item $\Phi(t) = t^p$ with $1 < p < \infty$.
\item $\Phi(t) = \exp(t) - t- 1$.
\item $\Phi(t) = \exp(t^p) - 1$ with $1 < p < \infty$.
\item $\Phi(t) = (1+t) \log(1+t) - t$.
\end{enumerate}


\paragraph{Complementary function.}
For $N$-function $\Phi$, its complementary function $\Psi : \R_+ \to \R_+$~\citep[\S8.3]{adams2003sobolev} is the $N$-function, defined as follows:
\begin{equation}\label{eq:complementary_func}
\Psi(t) = \sup \left[at - \Phi(a) \, \mid \, a \ge 0  \right], \quad\mbox{for}\,\,\, t\geq 0.
\end{equation}

\paragraph{Popular examples of complementary pairs of $N$-functions.} Some popular complementary pairs of $N$-functions~\citep[\S8.3]{adams2003sobolev},~\citep[\S 2.2]{rao1991theory} are as follows:
\begin{enumerate}
\item $\Phi(t) = \frac{t^p}{p}$ and $\Psi(t) = \frac{t^q}{q}$ where $q$ is the conjugate of $p$, i.e., $\frac{1}{p} + \frac{1}{q} = 1$ and $1 < p < \infty$.
\item $\Phi(t) = \exp(t) - t - 1$ and $\Psi(t) = (1+t)\log(1+t) - t$.
\item For the $N$-function $\Phi(t) = \exp(t^p) - 1$ with $1 < p < \infty$, its complementary $N$-function yields an explicit for, but not simple~\citep[\S 2.2]{rao1991theory}, see~\citep[\S A.8]{le2024generalized} for the details.
\end{enumerate}

\paragraph{Young inequality.} Let $\Phi, \Psi$ be a pair of complementary $N$-functions, then we have
\[
	st \le \Psi(s) + \Phi(t).
\]

\paragraph{Orlicz norm.} Together with the Luxemburg norm, the Orlicz norm~\citep[\S3.3, Definition 2]{rao1991theory} is a popular norm for $L_{\Phi}(\G, \lambda)$ in the literature, defined as
\begin{equation}\label{eq:OrliczNorm}
\|f\|_{\Phi} := \sup{\Big\{ \int_{\G} | f(x) g(x)| \lambda(\text{d}x) \, \mid \, \int_{\G} \Psi(|g(x)|) \lambda(\text{d}x) \leq 1\Big\} }, 
\end{equation}
where $\Psi$ is the complementary $N$-function of $\Phi$.

\paragraph{Computation for Orlicz norm.} Following~\citep[\S3.3, Theorem 13]{rao1991theory}, the Orlicz norm can be recasted as follows: 
\begin{align*}
\|f\|_{\Phi}  =  
\inf_{k > 0} \frac{1}{k}\left( 1 + \int_{\G} \Phi(k \left| f(x) \right|) \lambda(\text{d}x) \right).
\end{align*}
Therefore, one can use any second-order method, e.g., \texttt{fmincon} solver in MATLAB (with trust region reflective algorithm), to solve the \emph{univariate} optimization problem for Orlicz norm computation. 


\paragraph{Equivalence~\citep[\S8.17]{adams2003sobolev}~\citep[\S13.11]{musielak2006orlicz}.} The Luxemburg norm is equivalent to the Orlicz norm. In fact, we have
\begin{equation}\label{eq:LuxemburgOrlicz}
\norm{f}_{L_\Phi} \le \norm{f}_{\Phi} \le 2 \norm{f}_{L_\Phi}.
\end{equation}


\paragraph{Connection between $\boldsymbol{L^{p}}$ and $\boldsymbol{L_{\Phi}}$ functional spaces.} When the convex function $\Phi(t) = t^p$, for $1 < p < \infty$, we have 
\[
L^{p}(\G, \lambda) = L_{\Phi}(\G, \lambda).
\]

\paragraph{Generalized H\"older inequality.} Let $\Phi, \Psi$ be a pair of complementary $N$-functions, then generalized H\"older inequality w.r.t. Luxemburg norm~\citep[\S8.11]{adams2003sobolev} is as follows:
\begin{equation}\label{eq:Holder-Luxemburg2}
\left| \int_{\G} f(x)g(x) \lambda(dx) \right| \le 2 \norm{f}_{L_\Phi} \norm{g}_{L_\Psi}.
\end{equation}
Additionally, we have the generalized H\"older inequality w.r.t. Luxemburg norm and Orlicz norm~\citep[\S13.13]{musielak2006orlicz} is as follows:
\begin{equation}\label{eq:Holder-LuxemburgOrlicz}
\left| \int_{\G} f(x)g(x) \lambda(dx) \right| \le \norm{f}_{L_\Phi} \norm{g}_{\Psi}.
\end{equation}

\subsection{Generalized Sobolev transport (GST)}\label{appsubsec:GST_app}

We briefly review generalized Sobolev transport (GST)~\citep{le2024generalized} for graph-based measures.


\paragraph{Generalized Sobolev transport (GST)~\citep{le2024generalized}.} Let $\Phi$ be an $N$-function and $\lambda$ be a nonnegative Borel measure on $\G$. For probability measures $\mu, \nu$ on a graph $\G$, the generalized Sobolev transport (GST) is defined as follows:
\begin{equation} \notag \label{eq:distance_gst}
\calGS_{\Phi}(\mu,\nu) \! \coloneqq \! \left\{
\begin{array}{cl}
\sup &  \Big| \int_\G f(x) \mu(\mathrm{d}x) - \int_\G f(x) \nu(\mathrm{d}x)  \Big| \\
\mathrm{s.t.} & f \in {\OrliczSobolevPsi}(\G, \lambda),  \, \|f'\|_{L_{\Psi}}\leq 1,
\end{array}
\right.
\end{equation}
where $\Psi$ is the complementary function of $\Phi$ (see Equation~\eqref{eq:complementary_func}).

Note that GST is a scalable variant of Orlicz-Wasserstein (OW) for graph-based probability measures (i.e., not OW itself). Moreover,~\citet[Proposition 4.6]{le2024generalized} showed that GST is equal to OW \emph{when the graph is a tree, and $\Phi(t) = t$ (i.e., the limite case of $N$-function).}

\subsection{Orlicz-Wasserstein (OW)}\label{appsubsec:OW_app}

We briefly review Orlicz-Wasserstein (OW)~\citep{sturm2011generalized} for graph-based measures. Following~\citet[Definition 3.2]{GuhaHN23}, the OW with the $N$-function $\Phi$ for measures $\mu, \nu \in \calP(\G)$ is defined as follows:
\begin{eqnarray}\label{eq:OrliczWasserstein}
W_{\Phi}(\mu, \nu) = \inf_{\pi \in \Pi(\mu, \nu)} \inf \left[ t > 0 : \int_{\G \times \G} \Phi\left(\frac{d_{\G}(x, z)}{t}\right) \text{d}\pi(x, z) \le 1\right],
\end{eqnarray}
where $\Pi(\mu, \nu)$ is the set of all  couplings between $\mu$ and $\nu$.

\section{Proofs}\label{app:sec:proofs}

In this section, we give detailed proofs for the theoretical results in the main manuscript and additional results in Appendix~\S\ref{app:sec:further_theoretical_results}.

\subsection{Proof for Theorem~\ref{thrm:equivalence_OrliczSobolev_Musielak}}\label{sec:proof:thrm:equivalence_OrliczSobolev_Musielak}

\begin{proof}
We first derive the lower bound as follows:
\begin{eqnarray}
    \norm{f}_{\OrliczSobolevPhi} &&= \norm{f}_{L_{\Phi}} +  \norm{f'}_{L_{\Phi}}\nonumber \\
    &&\ge \norm{f'}_{L_{\Phi}} \label{eq:tmp_lb01}.
\end{eqnarray}

Additionally, for any scalar $k \ge 1$, we have
\begin{eqnarray}
 \int_{\G} \Phi\!\left(k \frac{|f'(x)|}{t}\right)\lambda(\dd x) &&\ge \int_{\G} k \, \Phi\!\left( \frac{|f'(x)|}{t}\right) \lambda(\dd x) \label{eq:tmp01}\\
 &&\ge \int_{\G} 
 \frac{k}{2} \left( 1 + \frac{\lambda(\Lambda(x))}{\lambda(\G)} \right) \Phi\!\left(\frac{|f'(x)|}{t}\right)\lambda(\dd x) \label{eq:tmp02} \\
 &&= \int_{\G} 
 \frac{k}{2} \hat w(x) \, \Phi\!\left(\frac{|f'(x)|}{t}\right)\lambda(\dd x).\nonumber
\end{eqnarray}
The first inequality in \eqref{eq:tmp01} is due to Lemma \ref{lem:Nfunc_multiplicative}, and the second inequality in \eqref{eq:tmp02} comes from the property of the length measure $\lambda$, i.e., $\lambda(\Lambda(x)) \le \lambda(\G)$ for all $x \in \G$.

Therefore, we obtain
\begin{equation*}
\left\{ t>0 \mid \int_{\G} \, \Phi\!\left(k\frac{|f'(x)|}{t}\right)\lambda(\dd x) \le 1\right\} \subseteq \left\{ t>0 \mid \int_{\G} \frac{k}{2} \hat w(x) \, \Phi\!\left(\frac{|f'(x)|}{t}\right)\lambda(\dd x) \le 1\right\}.
\end{equation*}
Notice that the infimum of a set is smaller than or equal to the infimum of its subset. Consequently, we obtain
\begin{equation*}
k\norm{f'}_{L_{\Phi}} \ge \norm{f'}_{L_{\Phi}^{k {\hat w}/2}}.
\end{equation*}
Therefore, we have shown that
\begin{equation}\label{eq:tmp_lb02}
\norm{f'}_{L_{\Phi}} \ge \frac{1}{k} \norm{f'}_{L_{\Phi}^{k {\hat w}/2}}, \quad \mbox{for every } k \ge 1.
\end{equation}

By choosing $k=2$ in \eqref{eq:tmp_lb02} and combining with the inequality in \eqref{eq:tmp_lb01}, we obtain the lower bound as follows:
\begin{equation*}
\norm{f}_{\OrliczSobolevPhi} \ge \frac{1}{2} \norm{f'}_{L_{\Phi}^{\hat w}}.
\end{equation*}

Next, let weight function $w_1(x) := \frac{\lambda(\Lambda(x))}{\lambda(\G)}$ for all $x \in \G$, we then derive the upper bound as follows:
\begin{eqnarray}
    \norm{f}_{\OrliczSobolevPhi} &&= \norm{f}_{L_{\Phi}} +  \norm{f'}_{L_{\Phi}} \nonumber\\
    &&\le \lambda(\G) \norm{f'}_{L_{\Phi}^{w_1}} + \norm{f'}_{L_{\Phi}} \label{eq:tmp_ub01} \\
    &&\le \lambda(\G) \norm{f'}_{L_{\Phi}^{\hat w}} + \norm{f'}_{L_{\Phi}^{\hat w}} \label{eq:tmp_ub02} \\
    &&= \left( 1 + \lambda(\G) \right) \norm{f'}_{L_{\Phi}^{\hat w}},\nonumber
\end{eqnarray}
where the first inequality in \eqref{eq:tmp_ub01} is obtained by using Theorem~\ref{thrm:bound_f_gradf} with $w_0(x) := \lambda(\Lambda(x))$, and the second inequality \eqref{eq:tmp_ub02} is obtained by using Lemma~\ref{lem:different_weight_func}.

Hence, the proof is complete.
\end{proof}


\subsection{Proof for Theorem~\ref{thrm:RGSI_1d_optimization}}\label{sec:proof:thrm:RGSI_1d_optimization}

\begin{proof}
Consider a critic function $f\in {\OrliczSobolevPsiZero}(\G, \lambda)$. Then by Definition~\ref{def:OrliczSobolev}, we have
\begin{equation}\label{eq:OSF_representation}
f(x) = f(z_0) + \int_{[z_0,x]} f'(y) \lambda(\mathrm{d}y), \quad \forall x \in \G.
\end{equation}
Using \eqref{eq:OSF_representation}, leveraging the indicator function of the shortest path $[z_0, x]$ (denoted as ${\bf{1}}_{[z_0,x]}$), and notice that $\mu(\G) = 1$, we get
\begin{align*}
\int_\G f(x) \mu(\mathrm{d}x) &= \int_\G f(z_0) \mu(\mathrm{d}x) + \int_\G  \int_{[z_0,x]} f'(y) \lambda(\mathrm{d}y) \mu(\mathrm{d}x)\\
&= f(z_0) + \int_\G  \int_{\G}  {\bf{1}}_{[z_0,x]}(y) \, f'(y) \lambda(\mathrm{d}y) \mu(\mathrm{d}x).
\end{align*}
Then, applying Fubini's theorem to interchange the order of integration in the above last integral, we obtain
\begin{align*}
\int_\G f(x) \mu(\mathrm{d}x) &= f(z_0) + \int_\G  \int_{\G}  {\bf{1}}_{[z_0,x]}(y) \, f'(y)  \mu(\mathrm{d}x) \lambda(\mathrm{d}y)\\
&= f(z_0) + \int_\G  \Big(\int_{\G}  {\bf{1}}_{[z_0,x]}(y) \,  \mu(\mathrm{d}x)\Big) f'(y)  \lambda(\mathrm{d}y).
\end{align*}
Using the definition of $\Lambda(y)$ in Equation~\eqref{sub-graph}, we can rewrite it as 
\begin{align*}
\int_\G f(x) \mu(\mathrm{d}x) = f(z_0) + \int_{\G} f'(y)  \mu(\Lambda(y)) \, \lambda(\mathrm{d}y).
\end{align*}

By exactly the same arguments, we also have 
\begin{align*}
\int_\G f(x) \nu(\mathrm{d}x) 
 = f(z_0) + \int_{\G} f'(y)  \nu(\Lambda(y)) \, \lambda(\mathrm{d}y).
\end{align*}

Consequently, the regularized generalized Sobolev IPM in Equation~\eqref{eq:regGenSobolevIPM} can be reformulated as
\begin{align}\label{eq:reformulation}
\widehat{\calGS}_{\Phi}(\mu, \nu)  = 
\sup_{f \in \calB_{\Psi}^{\hat w}} \left| \int_{\G} f'(x) \big[ \mu(\Lambda(x)) -  \nu(\Lambda(x))\big] \, \lambda(\mathrm{d}x) \right|,
\end{align}
where we recall that $\calB_{\Psi}^{\hat w} = \left\{f \in {\OrliczSobolevPsiZero} (\G, \lambda) : \norm{f'}_{L_{\Psi}^{\hat{w}}} \leq 1 \right\}$ (see Equation~\eqref{eq:weightedLp_ball_regularized}). 

Observe that, we have on one hand
\[
\{ f': \, f \in \calB_{\Psi}^{\hat w} \} \subset \{g \in L_{\Psi}(\G, \lambda): \, \|g\|_{L_{\Psi}^{\hat{w}}}\leq 1  \}.
\]
On the other hand, for any $g\in L_{\Psi}(\G, \lambda)$, we have $g=f'$ with $f(x) \coloneqq  \int_{[z_0,x]} g(y) \lambda(\mathrm{d}y)\in {\OrliczSobolevPsiZero}(\G, \lambda)$. 

Therefore, we conclude that
\begin{equation}\label{app:eq:unitball_func_grad}
\{ f': \, f\in \calB_{\Psi}^{\hat w} \} = \{g\in L_{\Psi}(\G, \lambda): \, \|g\|_{L_{\Psi}^{\hat{w}}} \leq 1  \}.
\end{equation}


Consequently, if we let $\hat f(x) :=  \frac{\mu(\Lambda(x)) -  \nu(\Lambda(x))}{\hat{w}(x)}$ for  $ x \in \G$, then Equation~\eqref{eq:reformulation} can be recasted as 
\begin{align}\label{app:eq:dualnorm_regGenSobolevIPM}
\widehat{\calGS}_{\Phi}(\mu, \nu)  &= 
\sup_{g\in L_{\Psi}(\G, \lambda): \, \norm{g}_{L_{\Psi}^{\hat{w}}} \le 1} \left| \int_{\G} \hat{w}(x) \hat f(x) g(x) \lambda(\mathrm{d}x) \right| 
\end{align}




Additionally, $\Phi, \Psi$ are a pair of complementary $N$-functions. Therefore, we obtain from~\eqref{app:eq:dualnorm_regGenSobolevIPM}
and~\citep[\S13.20]{musielak2006orlicz}\footnote{Also see~\citep[Proposition 10, \S3.4]{rao1991theory}.} that 
\begin{align}\label{eq:same_as_MusielakOrlicz}
\widehat{\calGS}_{\Phi}(\mu, \nu)  =  \|\hat f\|_{\Phi, \hat w},
\end{align}
where $\|\hat f\|_{\Phi, \hat w}$ is the Musileak-Orlicz norm with weight function $\hat w$, 
define by (see~\citep[\S13.11]{musielak2006orlicz}\footnote{Also see ~\citep[Definition 2, \S3.3]{rao1991theory},~\citep[Definition 3.4.5]{harjulehto2019generalized}.})
\begin{equation}\label{eq:MusielakOrliczNorm}
\|\hat f\|_{\Phi, \hat w} := \left\{ \int_{\G} \hat w(x) \hat f(x) g(x) \lambda(\text{d}x):\, \int_{\G} \hat w(x) \Psi(|g(x)|) \lambda(\text{d}x) \leq 1 \right\}. 
\end{equation}



By applying~\citep[Theorem 10.5, \S10.8]{krasnoselskii1961convex}\footnote{See also~\citep[Theorem 13, \S3.3]{rao1991theory}.}, we have 
\begin{align*}
\|\hat f\|_{\Phi, \hat w}  =  
\inf_{k > 0} \frac{1}{k}\left( 1 + \int_{\G} \hat w(x) \, \Phi\!\left(k \left| \hat f(x) \right|\right) \lambda(\text{d}x) \right).
\end{align*}
This together with \eqref{eq:same_as_MusielakOrlicz} yields
\begin{eqnarray*}
\widehat \calGS_{\Phi}(\mu,\nu )  &&=  
\inf_{k > 0} \frac{1}{k}\left( 1 + \int_{\G} \hat w(x) \, \Phi\!\left(k \left| \hat f(x) \right|\right) \lambda(\text{d}x) \right) \\
&&= \inf_{k > 0} \frac{1}{k}\left( 1 + \int_{\G} \hat w(x) \, \Phi\!\left(k \left| \frac{\mu(\Lambda(x)) -  \nu(\Lambda(x))}{\hat{w}(x)} \right|\right) \lambda(\text{d}x) \right) \\
&&= \inf_{k > 0} \frac{1}{k}\left( 1 + \int_{\G} \hat w(x) \, \Phi\!\left( \frac{k}{\hat{w}(x)} \left| {\mu(\Lambda(x)) -  \nu(\Lambda(x))} \right|\right) \lambda(\text{d}x) \right).
\end{eqnarray*}
This completes the proof of the theorem.

\end{proof}

\subsection{Proof for Theorem~\ref{thrm:RGSI_1d_optimization_discrete}}\label{sec:proof:thrm:RGSI_1d_optimization_discrete}

\begin{proof}

    We consider the length measure on graph $\G$ for $\lambda$. Thus, we have $\lambda(\{x\}) = 0$ for all $x \in \G$. 
    
    Consequently, we have
    \begin{equation}\label{eq:edge_regGenSobolevIPM_tmp1}
    \widehat{\calGS}_{\Phi}(\mu,\nu ) = \inf_{k > 0} \frac{1}{k}\left( 1 + \sum_{e=\langle u,v\rangle\in E}   \int_{(u,v)} \hat w(x) \, \Phi\!\left( \frac{k}{\hat{w}(x)} \left|  \mu(\Lambda(x)) -  \nu(\Lambda(x)) \right|\right) \lambda(\text{d}x) \right).
    \end{equation}
    
    Additionally, we consider input probability measures $\mu, \nu$ supported on nodes in $V$ of graph $\G$. Thus, for all edge $e=\langle u,v\rangle\in E$, and any point $x \in (u, v)$, we have
    \begin{equation}\label{eq:remove_interior}
    \mu(\Lambda(x)) -  \nu(\Lambda(x))= \mu(\Lambda(x)\setminus (u,v)) -  \nu(\Lambda(x)\setminus (u,v)).
    \end{equation}
    

    Moreover, let us consider edge $e=\langle u,v\rangle \in E$.  Then for any $x\in (u,v)$, we have $y\in \G\setminus (u,v)$ belongs to $\Lambda(x)$ if and only if $y\in \gamma_e$ where we recall that $\Lambda(x)$ and $\gamma_e$ are defined in Equation~\eqref{sub-graph}. Thus, we have
    \begin{equation}\label{eq:GammaEdge_gammaEdge}
    \Lambda(x)\setminus (u,v) =\gamma_e, \qquad \forall x \in (u, v).
    \end{equation}
    Using Equations~\eqref{eq:remove_interior} and~\eqref{eq:GammaEdge_gammaEdge}, we can rewrite Equation~\eqref{eq:edge_regGenSobolevIPM_tmp1} as

    \begin{equation}\label{eq:remove_interior_edge_regGenSobolevIPM_tmp1}
    \widehat{\calGS}_{\Phi}(\mu,\nu ) = \inf_{k > 0} \frac{1}{k}\left( 1 + \sum_{e=\langle u,v\rangle\in E}   \int_{(u,v)} \hat w(x) \, \Phi\!\left( \frac{k}{\hat{w}(x)} \left|  \mu(\gamma_e) -  \nu(\gamma_e) \right|\right) \lambda(\text{d}x) \right).
    \end{equation}

    
    We next want to compute the integral in \eqref{eq:remove_interior_edge_regGenSobolevIPM_tmp1}
    for each edge $\langle u,v\rangle\in E$.

    For this, recall that $\hat w(x) = 1 + \frac{\lambda(\Lambda(x))}{\lambda(\G)}, \forall x \in \G$ (see Equation~\eqref{eq:weighting_func_wOrlicz}). Without loss of generality, assume that $d_{\G}(z_0, u) \le d_{\G}(z_0, v)$, i.e., among two nodes $u, v$ of the edge $e$, node $v$ is farther away from the root node $z_0$ than node $u$. 

    Notice that for $x \in (u, v)$, we can write $x = v + t (u-v)$ for $t \in (0, 1)$. With this change of variable, we have
    \begin{eqnarray*}
    && \hspace{-4em}\int_{(u, v)} \left[ 1 + \frac{\lambda(\Lambda(x))}{\lambda(\G)} \right] \Phi\!\left( \frac{k}{1 + \frac{\lambda(\Lambda(x))}{\lambda(\G)}} \left|  \mu(\gamma_e) -  \nu(\gamma_e) \right|\right) \lambda(\dd x) \\
    &&= \int_0^1 \left[ 1 + \frac{\lambda(\Lambda( v + t (u-v)))}{\lambda(\G)} \right] \Phi\!\left( \frac{k}{1 + \frac{\lambda(\Lambda( v + t (u-v)))}{\lambda(\G)}} \left|  \mu(\gamma_e) -  \nu(\gamma_e) \right|\right)   w_e\dd t.
    \end{eqnarray*}

    Moreover, we have
    \[
    \lambda(\Lambda( v + t (u-v))) = \lambda(\Lambda(v)) + \lambda([v,  v + t (u-v)]) = 
    \lambda(\gamma_e) + w_e t.
    \]
    
    Therefore, 
    \begin{eqnarray*}
    && \hspace{-4em}\int_{(u, v)} \left[ 1 + \frac{\lambda(\Lambda(x))}{\lambda(\G)} \right] \Phi\!\left( \frac{k}{1 + \frac{\lambda(\Lambda(x))}{\lambda(\G)}} \left|  \mu(\gamma_e) -  \nu(\gamma_e) \right|\right) \lambda(\dd x) \\
    &&= \int_0^1 \left[ 1 + \frac{\lambda(\gamma_e) + w_e t}{\lambda(\G)} \right] \Phi\!\left( \frac{k}{1 + \frac{\lambda(\gamma_e) + w_e t}{\lambda(\G)}} \left|  \mu(\gamma_e) -  \nu(\gamma_e) \right|\right)   w_e\dd t \\
    &&= \int_0^1 \left[ \frac{w_e}{\lambda(\G)}t + 1 + \frac{\lambda(\gamma_e)}{\lambda(\G)} \right] \Phi\!\left( \frac{k \left|  \mu(\gamma_e) -  \nu(\gamma_e) \right|}{\frac{w_e}{\lambda(\G)}t + 1 + \frac{\lambda(\gamma_e)}{\lambda(\G)}} \right)   w_e\dd t
    \end{eqnarray*}

 
Hence,  the proof is  complete. 

\end{proof}

\subsection{Proof for Proposition~\ref{prop:RGSI_closed_form_discrete}}\label{sec:proof:prop:RGSI_closed_form_discrete}

\begin{proof}

    For the length measure $\lambda$ on graph $\G$, we have $\lambda(\{x\}) = 0$ for all $x \in \G$. Consequently, for $N$-function $\Phi(t) = \frac{(p-1)^{p-1}}{p^p} t^p$ with $1 < p < \infty$, from the closed-form expression of the generalized Sobolev IPM with Musielak regularization in Proposition~\ref{prop:RGSI_closed_form}, we obtain
    \begin{equation}\label{eq:edge_regGenSobolevIPM}
    \widehat{\calGS}_{\Phi}(\mu, \nu)^p = \sum_{e=\langle u,v\rangle\in E}   \int_{(u,v)} \hat{w}(x)^{1-p} \left| \mu(\Lambda(x)) -  \nu(\Lambda(x)) \right|^p \lambda(\dd x).
    \end{equation}
    
    Additionally, for input measures $\mu, \nu$ supported on nodes in $V$ of graph $\G$, then for all edge $e=\langle u,v\rangle\in E$, and any point $x \in (u, v)$, we have
    \[
    \mu(\Lambda(x)) -  \nu(\Lambda(x))= \mu(\Lambda(x)\setminus (u,v)) -  \nu(\Lambda(x)\setminus (u,v)).
    \]
    
    Therefore, from Equation~\eqref{eq:edge_regGenSobolevIPM}, we obtain
    \begin{equation}\label{eq:remove_interior_regGenSobolevIPM}
    \widehat{\calGS}_{\Phi}(\mu, \nu)^p = \sum_{e=\langle u,v\rangle\in E}   \int_{(u,v)} \hat{w}(x)^{1-p} \left| \mu(\Lambda(x)\setminus (u,v)) -  \nu(\Lambda(x)\setminus (u,v)) \right|^p \lambda(\dd x).
    \end{equation}

    Moreover, consider edge $e=\langle u,v\rangle \in E$, for any $x\in (u,v)$, then we have $y\in \G\setminus (u,v)$ belongs to $\Lambda(x)$ if and only if $y\in \gamma_e$.\footnote{See Equation~\eqref{sub-graph} for the definitions of $\Lambda(x)$ and $\gamma_e$.} Thus, we have
    \[
    \Lambda(x)\setminus (u,v) =\gamma_e, \qquad \forall x \in (u, v).
    \]
    Thus, from Equation~\eqref{eq:remove_interior_regGenSobolevIPM}, we obtain
    \begin{equation}\label{eq:remove_interior_edge_regGenSobolevIPM}
    \widehat{\calGS}_{\Phi}(\mu, \nu)^p = \sum_{e=\langle u,v\rangle\in E} \left| \mu(\gamma_e) -  \nu(\gamma_e) \right|^p  \int_{(u,v)} \hat{w}(x)^{1-p} \lambda(\dd x).
    \end{equation}

    From Equation~\eqref{eq:weighted_Orlicz_norm}, for any $x$ in $\G$, we have
    \begin{equation*}
        \hat{w}(x) = 1 + \frac{\lambda(\Lambda(x))}{\lambda(\G)}.
    \end{equation*}
    Without loss of generality, for any edge $e=\langle u,v\rangle\in E$, assume that $d_{\G}(z_0, u) \le d_{\G}(z_0, v)$, i.e., among two nodes $u, v$ of the edge $e$, node $v$ is farther away from the root node $z_0$ than node $u$. 

    Observe that for $x \in (u, v)$, then $x = v + t (u-v)$ for $t \in (0, 1)$. Using this change of variable, we obtain
    \begin{equation*}
    \int_{(u, v)} \left[ 1 + \frac{\lambda(\Lambda(x))}{\lambda(\G)} \right]^{1-p} \lambda(\dd x) = \int_0^1 \left[ 1 + \lambda(\G)^{-1}\lambda(\Lambda( v + t (u-v))) \right]^{1-p}   w_e\dd t
    \end{equation*}
    
    Additionally, notice that
    \[
    \lambda(\Lambda( v + t (u-v))) = \lambda(\Lambda(v)) + \lambda([v,  v + t (u-v)]) = \lambda(\gamma_e) + w_e t.
    \]
    
    Thus, we obtain 
    \begin{equation*}
    \int_{(u, v)} \left[ 1 + \frac{\lambda(\Lambda(x))}{\lambda(\G)} \right]^{1-p} \lambda(\dd x) = \int_0^1 \left[ 1 + \lambda(\G)^{-1}\lambda(\gamma_e) + \lambda(\G)^{-1}w_e t \right]^{1-p} w_e\dd t.
    \end{equation*}
    
    Furthermore, the last integral can be computed easily depending on the case $p=2$ or $p\neq 2$.
    Consequently, we obtain 
\begin{equation*}
    \int_{(u, v)} \left[ 1 +  \frac{\lambda(\Lambda(x))}{\lambda(\G)} \right]^{1-p} \lambda(\dd x) = \left\{
        \begin{array}{ll}
              \lambda(\G)\log{\left(1 + \frac{w_e}{\lambda(\G) + \lambda(\gamma_e)} \right)} & \mbox{if } p = 2, \hspace{-1.1em}\\
              \frac{\left(\lambda(\G) + \lambda(\gamma_e)+ w_e \right)^{2-p} - \left(\lambda(\G) + \lambda(\gamma_e) \right)^{2-p} }{(2-p)\lambda(\G)^{1-p}} & \text{otherwise.}\hspace{-1.1em}
        \end{array}  
    \right. 
    \end{equation*}
Thus, we have $\int_{(u, v)} \left[ 1 + \frac{\lambda(\Lambda(x))}{\lambda(\G)} \right]^{1-p} \lambda(\dd x) = \beta_e$ (see Equation~\eqref{eq:closed_form_edge_weight_regSobolevIPM}).

Consequently, from Equation~\eqref{eq:remove_interior_edge_regGenSobolevIPM}, we obtain
\begin{equation}\label{eq:closed_form_discrete_regGenSobolevIPM}
    \widehat{\calGS}_{\Phi}(\mu, \nu) = \left( \sum_{e\in E} \beta_e | \mu(\gamma_e) -  \nu(\gamma_e) |^p \right)^{\frac{1}{p}}.
\end{equation}
 
Hence,  the proof is  completed.

\end{proof}

\subsection{Proof for Lemma~\ref{thrm:metrization}}\label{app:proof:thrm:metrization}

\begin{proof}

We will prove that the regularized generalized Sobolev IPM $\widehat \calGS_{\Phi}$ satisfies: (i) nonnegativity, (ii) indiscernibility, (iii) symmetry, and (iv) triangle inequality.


\textbf{(i) Nonnegativity.} By choosing $f=0$ in Definition~\ref{def:regSobolevIPM}, we have that $\widehat \calGS_{\Phi}(\mu,\nu) \ge 0$ for every pair of probability measures $(\mu, \nu)$ in $\calP(\G) \times \calP(\G)$. Therefore, the regularized generalized Sobolev IPM $\widehat \calGS_{\Phi}$ is nonnegative.


\textbf{(ii) Indiscernibility.} Assume that $\widehat \calGS_{\Phi}(\mu,\nu) =0$, then we have 
\begin{align}\label{int_identity}
\int_\G f(x) \mu(\mathrm{d}x) - \int_\G f(x) \nu(\mathrm{d}x) = 0,
\end{align}
for all $f \in {\OrliczSobolevPsiZero} (\G, \lambda)$ satisfying the constraint $\norm{f'}_{L_{\Psi}^{\hat{w}}} \leq 1$.


Let $g \in {\OrliczSobolevPsiZero}(\G, \lambda)$ be any nonconstant function. Then $\alpha := \|g'\|_{L_{\Psi}} > 0$. Then by choosing $f : = \frac{g}{\alpha}$, we have 
$f\in {\OrliczSobolevPsiZero}(\G, \lambda)$ with  $ \|f'\|_{L_{\Psi}} =\|\frac{g'}{\alpha}\|_{L_{\Psi}}= \frac{1}{\alpha} \|g'\|_{L_{\Psi}} = 1$ where we use Lemma~\ref{lem:MusielakNorm_kPhi} for the second equation. Hence, it follows from \eqref{int_identity} that
\begin{align*}
\int_\G \frac{g(x)}{\alpha} \mu(\mathrm{d}x) - \int_\G \frac{g(x)}{\alpha} \nu(\mathrm{d}x) = 0,
\end{align*}
which implies that
\begin{align}\label{eq: Indiscernibility}
\int_\G g(x) \mu(\mathrm{d}x) =\int_\G g(x)\nu(\mathrm{d}x).
\end{align}
Thus, we have shown that \eqref{eq: Indiscernibility} holds true for every nonconstant function $g \in {\OrliczSobolevPsiZero}(\G, \lambda)$. Additionally, Equation~\eqref{eq: Indiscernibility} is also obviously true for any constant function $g$. Therefore, we obtain
\[
\int_\G g(x) \mu(\mathrm{d}x) = \int_\G g(x) \nu(\mathrm{d}x), 
\]
for every $g\in {\OrliczSobolevPsiZero}(\G, \lambda)$, which gives $\mu =\nu$ as desired. 

\textbf{(iii) Symmetry.} This property is obvious from  Definition~\ref{def:regSobolevIPM} as the value $\widehat \calGS_{\Phi}(\mu,\nu)$ is unchanged when the role of $\mu$ and $\nu$ is interchanged, i.e.,  
\[
\widehat \calGS_{\Phi}(\mu,\nu) = \widehat \calGS_{\Phi}(\nu,\mu).
\]


\textbf{(iv) Triangle inequality.} Let $\mu,\nu,\sigma$ be probability measures in $\calP(\G)$, then for any function $f \in {\OrliczSobolevPsiZero}(\G, \lambda)$ satisfying the constraint $\|f'\|_{L_{\Psi}^{\hat w}}\leq 1$, we have
\begin{eqnarray*}
&& \hspace{-5.5em} \left| \int_\G f(x) \mu(\mathrm{d}x) - \int_\G f(x) \nu(\mathrm{d}x) \right| \\
&&= \left| \left[ \int_\G f(x) \mu(\mathrm{d}x) - \int_\G f(x) \sigma(\mathrm{d}x) \right]  + \left[\int_\G f(x) \sigma(\mathrm{d}x) - \int_\G f(x) \nu(\mathrm{d}x)\right] \right| \\
&&\leq  \left|  \int_\G f(x) \mu(\mathrm{d}x) - \int_\G f(x) \sigma(\mathrm{d}x) \right|  + \left|\int_\G f(x) \sigma(\mathrm{d}x) - \int_\G f(x) \nu(\mathrm{d}x) \right|\\
&&\leq \widehat \calGS_{\Phi}(\mu,\sigma) + \widehat \calGS_{\Phi}(\sigma, \nu).
\end{eqnarray*}
By taking the supremum over $f$, this implies that 
\[
\widehat \calGS_{\Phi}(\mu, \nu) \leq \widehat \calGS_{\Phi}(\mu,\sigma) + \widehat \calGS_{\Phi}(\sigma, \nu).
\]

Due to these above properties, we conclude that the regularized generalized Sobolev IPM $\widehat \calGS_{\Phi}$ is a metric on the space $\calP(\G)$ of probability measures on graph $\G$.

\end{proof}

\subsection{Proof for Proposition~\ref{prop:strong_metric}}\label{app:proof:prop:strong_metric}

%
%

\begin{proof}

Let $\Psi_1$ and $\Psi_2$ be respectively the complementary functions of $\Phi_1$ and $\Phi_2$, consider an arbitrary positive scale $t\in \R_+$, and notice that $\Phi_1 \leq \Phi_2$, then we have:
\begin{eqnarray*}
   at - \Phi_1(a) &&\ge \quad at - \Phi_2(a) \, \, \text{ for every } {a \in \R_+}, \\
    \Rightarrow \sup_{a \ge 0} \left( at - \Phi_1(a) \right) &&\ge \quad \sup_{a \ge 0} \left( at - \Phi_2(a) \right).
\end{eqnarray*}
This implies that
\[
\Psi_1(t) \ge \Psi_2(t), \text{ for all } t \in \R_+.
\]
It follows that $L_{\Psi_1}^{\hat w}(\G, \lambda) \subset L_{\Psi_2}^{\hat w}(\G, \lambda)$ and ${W\!L}_{\Psi_1, 0}^{1}(\G, \lambda) \subset {W\!L}_{\Psi_2, 0}^{1}(\G, \lambda)$. Moreover, for any fixed Orlicz function $f'$ and any number $t>0$, we have
\[
\int_{\G} \hat w(x) \, \Psi_1\!\left(\frac{|f'(x)|}{t}\right)\lambda(\text{d}x) \ge \int_{\G} \hat w(x) \, \Psi_2\!\left(\frac{|f'(x)|}{t}\right)\lambda(\text{d}x).
\]
Consequently, we obtain
\[
\left\{ t >0  \mid \int_{\G} \hat w(x) \, \Psi_1\!\left(\frac{|f'(x)|}{t}\right)\lambda(\text{d}x) \le 1 \right\} \subset \left\{ t>0 \mid \int_{\G} \hat w(x) \, \Psi_2\!\left(\frac{|f'(x)|}{t}\right)\lambda(\text{d}x) \le 1 \right\}.
\]
Since the infimum of a set is smaller than or equal to the infimum of its subset, we deduce that
\[
\|f'\|_{L_{\Psi_1}^{\hat w}} \ge \|f'\|_{L_{\Psi_2}^{\hat w}}. 
\]
It follows from this and ${W\!L}_{\Psi_1, 0}^{1}(\G, \lambda) \subset {W\!L}_{\Psi_2, 0}^{1}(\G, \lambda)$ that 
\[
\left\{f \mid f \in {W\!L}_{\Psi_1, 0}^{1}(\G, \lambda),  \, \|f'\|_{L_{\Psi_1}^{\hat w}}\leq 1 \right\} \subset \left\{f \mid f \in {W\!L}_{\Psi_2, 0}^{1}(\G, \lambda),  \, \|f'\|_{L_{\Psi_2}^{\hat w}}\leq 1 \right\}.
\]
Since the supremum of a set is larger than or equal to the supremum of its subset, we conclude that
\[
\widehat \calGS_{\Phi_1}(\mu,\nu ) \le \widehat \calGS_{\Phi_2}(\mu,\nu ),
\]
for any input measures $\mu, \nu$ in $\calP(\G)$. Therefore, the proof is complete.
    
\end{proof}

\subsection{Proof for Proposition~\ref{thrm:relationGenSobolevIPM}}\label{app:proof:thrm:relationGenSobolevIPM}

\begin{proof}

    Given a positive scalar $c > 0$, a pair of complementary $N$-functions $\Phi, \Psi$, and let $\calB_{\Psi, c}^{\hat w} := \left\{ f \in W\!L_{\Psi, 0}^{1}(\G, \lambda) : \norm{f'}_{L^{\hat{w}}_{\Psi}} \le \frac{1}{c} \right\}$.  We define the IPM distance w.r.t. $\calB_{\Psi, c}^{\hat w}$ as follows
\begin{equation}\label{eq:IPM_tmp_tmp1}
\widetilde{\calGS}_{\Phi, c}(\mu, \nu) := \sup_{f \in \calB_{\Psi, c}^{\hat w}} \left| \int_{\G} f(x) \mu(\text{d}x) - \int_{\G} f(y) \nu(\text{d}y) \right|.
\end{equation}
By exploiting the graph structure for the IPM objective function and applying a similar reasoning as in the proof of identity \eqref{eq:reformulation} in \S\ref{sec:proof:thrm:RGSI_1d_optimization}, we can rewrite Equation~\eqref{eq:IPM_tmp_tmp1} as 
\begin{align}\label{eq:reformulation_IMP_tmp}
\widetilde{\calGS}_{\Phi, c}(\mu, \nu)  = 
\sup_{f \in \calB_{\Psi, c}^{\hat w}} \left| \int_{\G} f'(x) \big[ \mu(\Lambda(x)) -  \nu(\Lambda(x))\big] \, \lambda(\mathrm{d}x) \right|.
\end{align}
Additionally, by using a similar reasoning as in the proof of \eqref{app:eq:unitball_func_grad} in \S\ref{sec:proof:thrm:RGSI_1d_optimization},  we have
\begin{equation}\label{app:eq:unitball_func_grad_IPM_tmp}
\left\{ f': \, f\in \calB_{\Psi, c}^{\hat w} \right\} = \left\{g\in L_{\Psi}(\G, \lambda): \, \|g\|_{L^{\hat w}_{\Psi}} \leq \frac{1}{c}  \right\}.
\end{equation}

Let $\tilde f(x) :=  \frac{\mu(\Lambda(x)) -  \nu(\Lambda(x))}{c \, \hat{w}(x)}$ for $ x \in \G$. Then by using \eqref{app:eq:unitball_func_grad_IPM_tmp}, Equation~\eqref{eq:reformulation_IMP_tmp} can be recasted as 
\begin{align}
\widetilde{\calGS}_{\Phi, c}(\mu, \nu)  &= 
\sup_{g\in L_{\Psi}(\G, \lambda):\, \norm{g}_{L^{\hat w}_{\Psi}} \le \frac{1}{c}} \left| \int_{\G} \hat{w}(x) \tilde f(x) [c \, g(x)] \lambda(\mathrm{d}x) \right| \nonumber \\
&= 
\sup_{\tilde g\in L_{\Psi}(\G, \lambda):\, \norm{\tilde g}_{L^{\hat w}_{\Psi}} \le 1} \left| \int_{\G} \hat{w}(x) \tilde f(x) \tilde g(x)\lambda(\mathrm{d}x) \right| \nonumber \\
&= \norm{\tilde f}_{\Phi, \hat w} \label{eq:app:MO_norm_tmp} \\
&= \inf_{k > 0} \frac{1}{k}\left( 1 + \int_{\G} \hat w(x) \, \Phi\!\left(k \left| \tilde f(x) \right|\right) \lambda(\text{d}x) \right) \label{eq:app:Amemiya_MO_norm_tmp} \\
&= \inf_{k > 0} \frac{1}{k}\left( 1 + \int_{\G} \hat w(x) \, \Phi\!\left(k \left| \frac{\mu(\Lambda(x)) -  \nu(\Lambda(x))}{c \, \hat{w}(x)} \right|\right) \lambda(\text{d}x) \right) \nonumber \\
&= \frac{1}{c} \inf_{(k/c) > 0} \frac{1}{(k/c)}\left( 1 + \int_{\G} \hat w(x) \, \Phi\!\left( \frac{(k/c)}{\hat w(x)} \left| {\mu(\Lambda(x)) -  \nu(\Lambda(x))} \right|\right) \lambda(\text{d}x) \right) \nonumber \\
&= \frac{1}{c} \, \widehat{\calGS}_{\Phi}(\mu, \nu), \label{eq:app:uni_opt_regGenSobolevIPM}
\end{align}
where the third equality in~\eqref{eq:app:MO_norm_tmp} following from~\citep[\S13.20]{musielak2006orlicz} and Equation~\eqref{eq:MusielakOrliczNorm}, the fourth equality in~\eqref{eq:app:Amemiya_MO_norm_tmp} following from~\citep[Theorem 10.5, \S10.8]{krasnoselskii1961convex}, and the last equality in~\eqref{eq:app:uni_opt_regGenSobolevIPM} following from Theorem~\ref{thrm:RGSI_1d_optimization}. 

Additionally, notice that from Theorem~\ref{thrm:equivalence_OrliczSobolev_Musielak}, we have 
\begin{equation*}
        c_1 \norm{f'}_{L_{\Psi}^{\hat w}} \le \norm{f}_{\OrliczSobolevPsi} \le c_2 \norm{f'}_{L_{\Psi}^{\hat w}},
\end{equation*}
where $c_1 = 1/2$, and $c_2 = 1 + \lambda(\G)$.

This implies that
\begin{equation*}
\calB_{\Psi, c_1}^{\hat w}  \supseteq \calB_{\Psi} \supseteq \calB_{\Psi, c_2}^{\hat w}.
\end{equation*}
Therefore, for probability measures $\mu, \nu \in \calP(\G)$, we have
\begin{equation}\label{eq:relation_SobolevIPM_IPMtmp}
\widetilde{\calGS}_{\Phi, c_1}(\mu, \nu) \ge \calGS_{\Phi}(\mu, \nu) \ge \widetilde{\calGS}_{\Phi, c_2}(\mu, \nu).
\end{equation}

It follows from Equations~\eqref{eq:app:uni_opt_regGenSobolevIPM} and~\eqref{eq:relation_SobolevIPM_IPMtmp} that
\begin{equation*}
\frac{1}{c_1} \, \widehat \calGS_{\Phi}(\mu, \nu) \ge \calGS_{\Phi}(\mu, \nu) \ge \frac{1}{c_2} \, \widehat \calGS_{\Phi}(\mu, \nu).
\end{equation*}
Consequently, we obtain
\begin{equation*}
{c_1} \, \calGS_{\Phi}(\mu, \nu) \le \widehat \calGS_{\Phi}(\mu, \nu) \le {c_2} \, \calGS_{\Phi}(\mu, \nu).
\end{equation*}
Hence, we have
\begin{equation*}
\frac{1}{2} \, \calGS_{\Phi}(\mu, \nu) \le \widehat \calGS_{\Phi}(\mu, \nu) \le (1 + \lambda(\G)) \, \calGS_{\Phi}(\mu, \nu).
\end{equation*}

The proof is complete.

\end{proof}

\subsection{Proof for Proposition~\ref{prop:Connection_GSIMPMR_RSIPM}}\label{app:proof:prop:Connection_GSIMPMR_RSIPM}

\begin{proof}

Following Equation~\eqref{app:eq:dualnorm_regGenSobolevIPM} in the proof of Theorem~\ref{thrm:RGSI_1d_optimization} in \S\ref{sec:proof:thrm:RGSI_1d_optimization}, we have
\begin{align}\label{eq:GSI_MR_tmp1}
\widehat{\calGS}_{\Phi}(\mu, \nu)  &= 
\sup_{g\in L_{\Psi}(\G, \lambda): \, \norm{g}_{L_{\Psi}^{\hat{w}}} \le 1} \left| \int_{\G} \hat{w}(x) \hat f(x) g(x) \lambda(\mathrm{d}x) \right|, 
\end{align}
where $\hat f(x) =  \frac{\mu(\Lambda(x)) -  \nu(\Lambda(x))}{\hat{w}(x)}$ for  $ x \in \G$.

Additionally, following~\citep[Remark A.1]{le2024generalized}, for $\Phi(t) = \frac{(p-1)^{p-1}}{p^p} t^p$ with $1 < p < \infty$, we have its complementary function $\Psi(t) = t^q$ where $q$ is the conjugate of $p$, i.e., $1/q + 1/p = 1$.

Consequently, we have $L_{\Psi} = L^q$, and $L_{\Psi}^{\hat w} = L_{\hat w}^q$, where $L_{\hat w}^q$ is the weighted $L^q$ space with weight function $\hat w$. Therefore, we can rewrite Equation~\eqref{eq:GSI_MR_tmp1} as follows:
\begin{align}
\widehat{\calGS}_{\Phi}(\mu, \nu)  &= 
\sup_{g\in L^q(\G, \lambda): \, \norm{g}_{L^q_{\hat{w}}} \le 1} \left| \int_{\G} \hat{w}(x) \hat f(x) g(x) \lambda(\mathrm{d}x) \right| \nonumber \\
&= \norm{\hat f}_{L^p_{\hat w}} \nonumber \\
&= \left[ \int_{\G} \hat w(x) \left| \hat f(x) \right|^p \lambda (\dd x) \right]^{\frac{1}{p}} \nonumber \\
&= \left[ \int_{\G} \hat w(x)^{1-p} \left| \mu(\Lambda(x)) - \nu(\Lambda(x)) \right|^p \lambda (\dd x) \right]^{\frac{1}{p}}. \label{eq:Connection_RSIPM_tmp1}
\end{align}

Moreover, following~\citep[Theorem 3.4]{le2025scalable}, the closed-form of the regularized $p$-order Sobolev IPM is as follows:
\begin{equation}\label{eq:Connection_RSIPM_tmp2}
    \hat \calS_p(\mu, \nu) = \left[ \int_{\G} \hat w_{\hat\calS}(x)^{1-p} \left| \mu(\Lambda(x)) - \nu(\Lambda(x)) \right|^p \lambda (\dd x) \right]^{\frac{1}{p}},
\end{equation}
where the weight function $\hat w_{\hat \calS}(x) = 1 + \lambda(\Lambda(x))$ for all $x \in \G$~\citep[Equation (5)]{le2025scalable}.

Recall that $\hat w(x) = 1 + \frac{\lambda(\Lambda(x))}{\lambda(\G)}$ for all $x \in \G$. For any $x \in \G$, we have
\begin{eqnarray*}
    \min(1, \lambda(\G)^{-1}) (1 + \lambda(\Lambda(x))) \le 1 + \frac{\lambda(\Lambda(x))}{\lambda(\G)} \le \max(1, \lambda(\G)^{-1}) (1 + \lambda(\Lambda(x))).
\end{eqnarray*}
Then, for $1 < p < \infty$, we obtain
\begin{eqnarray}\label{eq:Connection_RSIPM_tmp3}
    \min(1, \lambda(\G)^{-1})^{1-p} \hat w_{\calS}(x)^{1-p} \ge \hat w(x)^{1-p} \ge \max(1, \lambda(\G)^{-1})^{1-p} \hat w_{\calS}(x).
\end{eqnarray}
Therefore, from Equations~\eqref{eq:Connection_RSIPM_tmp1},~\eqref{eq:Connection_RSIPM_tmp2},~\eqref{eq:Connection_RSIPM_tmp3}, we get
\begin{equation*}
    \hat c_1 \, \hat \calS_p(\mu, \nu) \le \widehat \calGS_{\Phi}(\mu, \nu) \le \hat c_2 \, \hat \calS_p(\mu, \nu),
\end{equation*}
where $\hat c_1 = \max(1, \lambda(\G)^{-1})^{\frac{1-p}{p}}$ and $\hat c_2 = \min(1, \lambda(\G)^{-1})^{\frac{1-p}{p}}$.

Hence, the proof is complete.
\end{proof}





\subsection{Proof for Proposition~\ref{prop:Connection_GSIMPMR_GST}}\label{app:proof:prop:Connection_GSIMPMR_GST}

\begin{proof}
    From Equation~\eqref{eq:weighting_func_wOrlicz}, for any $x \in \G$, we have
    \begin{equation*}
        1 \le \hat w(x) \le 2.
    \end{equation*}
    Then, for any $t > 0$, we have
    \begin{equation*}
        \int_{\G} \Psi\!\left(\frac{|f'(x)|}{t} \right) \lambda(\dd x) \le         \int_{\G} \hat w(x) \, \Psi\!\left(\frac{|f'(x)|}{t} \right) \lambda(\dd x) \le         2 \int_{\G} \Psi\!\left(\frac{|f'(x)|}{t} \right) \lambda(\dd x).
    \end{equation*}
    Consequently, we obtain
    \begin{eqnarray}\label{eq:subset_GST_tmp1}
       \left\{ t > 0 \mid \int_{\G} \Psi\!\left(\frac{|f'(x)|}{t} \right) \lambda(\dd x) \le 1 \right\} \hspace{15em}  \\ \nonumber
       \supseteq \left\{ t > 0 \mid         \int_{\G} \hat w(x) \, \Psi\!\left(\frac{|f'(x)|}{t} \right) \lambda(\dd x) \le 1 \right\} \hspace{5em} \\ \nonumber 
       \supseteq   \left\{ t > 0 \mid      2 \int_{\G} \Psi\!\left(\frac{|f'(x)|}{t} \right) \lambda(\dd x) \le 1\right\}.
    \end{eqnarray}
    Additionally, observe that by following Lemma~\ref{lem:Nfunc_multiplicative}, for any $t > 0$, we have
    \begin{equation*}
        2 \int_{\G} \Psi\!\left(\frac{|f'(x)|}{t} \right) \lambda(\dd x) \le \int_{\G} \Psi\!\left(2 \, \frac{|f'(x)|}{t} \right) \lambda(\dd x)
    \end{equation*}
    Consequently, we have
    \begin{equation}\label{eq:subset_GST_tmp2}
        \left\{ t > 0 \mid 2 \int_{\G} \Psi\!\left(\frac{|f'(x)|}{t} \right) \lambda(\dd x) \le 1 \right\} \supseteq \left\{ t > 0 \mid \int_{\G} \Psi\!\left(2 \, \frac{|f'(x)|}{t} \right) \lambda(\dd x) \le 1 \right\}.
    \end{equation}
    
    Notice that the infimum of a set is smaller than or equal to the infimum of its subset. Consequently, from Equations~\eqref{eq:subset_GST_tmp1}, and~\eqref{eq:subset_GST_tmp2}, we obtain
    \begin{equation*}
        \norm{f'}_{L_{\Psi}} \le \norm{f'}_{L_{\Psi}^{\hat w}} \le \inf \left\{ t > 0 \mid \int_{\G} \Psi\!\left(2 \, \frac{|f'(x)|}{t} \right) \lambda(\dd x) \le 1 \right\}.
    \end{equation*}
    Moreover, we also have
    \begin{eqnarray*}
        \inf \left\{ t > 0 \mid \int_{\G} \Psi\!\left(2 \, \frac{|f'(x)|}{t} \right) \lambda(\dd x) \le 1 \right\} \hspace{-1.8em} &&= 2\inf \left\{ (t/2) > 0 \mid \int_{\G} \Psi\!\left( \, \frac{|f'(x)|}{(t/2)} \right) \lambda(\dd x) \le 1 \right\} \nonumber \\ 
        &&= 2 \norm{f'}_{L_{\Psi}}.
    \end{eqnarray*}
    Thus, we obtain
    \begin{equation*}
        \norm{f'}_{L_{\Psi}} \le \norm{f'}_{L_{\Psi}^{\hat w}} \le 2 \norm{f'}_{L_{\Psi}}.
    \end{equation*}
    
    With the technical assumption $f(z_0) = 0$, then $\OrliczSobolevPsi$ is equal to $\OrliczSobolevPsiZero$ (as assumed for GSI-M throughout our work). Thus, we have
    \begin{equation}\label{eq:Connection_GST_p1}
        \left\{f \in \OrliczSobolevPsi, \norm{f'}_{L_{\Psi}} \le 1 \right\} \supseteq \left\{f \in \OrliczSobolevPsiZero, \norm{f'}_{L_{\Psi}^{\hat w}} \le 1 \right\} \supseteq \left\{f \in \OrliczSobolevPsi, \norm{f'}_{L_{\Psi}} \le 1/2 \right\}.
    \end{equation}

    Additionally, for a positive scalar $c > 0$, a pair of complementary $N$-function $\Phi, \Psi$, let consider
    \begin{equation}\label{eq:Connection_GST_p2}
    \widetilde \calGST_{\Phi, c}(\mu, \nu) := \sup_{f \in \OrliczSobolevPsi, \norm{f'}_{L_{\Psi}} \le 1/c} \left| \int_{\G} f(x) \mu(\dd x) - \int_{\G} f(x) \nu(\dd x)  \right|.  
    \end{equation}
    Following~\citep[Equations (12), (13) in \S A.1]{le2024generalized}, we can rewrite $\widetilde \calGST_{\Phi, c}$ as follows:
    \begin{equation*}
        \widetilde \calGST_{\Phi, c}(\mu, \nu) = \sup \limits_{g\in L_{\Psi}(\G, \lambda): \, \|g\|_{L_\Psi}\leq 1/c} \left| \int_{\G} g(x) h(x) \, \lambda(\mathrm{d}x) \right|,
    \end{equation*}
    where $h(x) :=  \mu(\Lambda(x)) -  \nu(\Lambda(x))$ for all $x \in \G$. Consequently, let $\tilde g = c g$, we have
    \begin{equation*}
        \widetilde \calGST_{\Phi, c}(\mu, \nu) = \sup \limits_{\tilde g\in L_{\Psi}(\G, \lambda): \, \|\tilde g\|_{L_\Psi}\leq 1} \left| \int_{\G} \tilde g(x) \frac{h(x)}{c} \, \lambda(\mathrm{d}x) \right|
    \end{equation*}

    By applying \citep[Proposition 10, pp.81]{rao1991theory}, we obtain
    \begin{align}\label{eq:same_as_Orlicz}
    \widetilde \calGST_{\Phi, c}(\mu, \nu)  =  \norm{\frac{h}{c}}_{\Phi}
    \end{align}
    where $\norm{\frac{h}{c}}_{\Phi}$ is the Orlicz norm~\citep[Definition~2, pp.58]{rao1991theory}, defined as
    \begin{equation}\label{eq:OrliczNorm}
    \norm{\frac{h}{c}}_{\Phi} := \sup{\Big\{ \int_{\G} \left| \frac{1}{c} h(x) g(x) \right| \lambda(\text{d}x):\, \int_{\G} \Psi(|g(x)|) \lambda(\text{d}x) \leq 1\Big\} }. 
    \end{equation}
    
    Then, by applying~\citep[Theorem 13, pp.69]{rao1991theory}, we have 
    \begin{align*}
    \norm{\frac{h}{c}}_{\Phi}  =  
    \inf_{k > 0} \frac{1}{k}\left( 1 + \int_{\G} \Phi\!\left(\frac{k}{c} \left| h(x) \right| \right) \lambda(\text{d}x) \right).
    \end{align*}
    Thus, together with \eqref{eq:same_as_Orlicz}, we obtain
    \begin{align}\label{eq:Connection_GST_p3}
    \widetilde \calGST_{\Phi, c}(\mu, \nu)  =  \frac{1}{c}
    \inf_{(k/c) > 0} \frac{1}{(k/c)}\left( 1 + \int_{\G} \Phi\!\left((k/c) \left| h(x) \right| \right) \lambda(\text{d}x) \right) = \frac{1}{c} \, \calGST_{\Phi}(\mu, \nu).
    \end{align}

    Since the supremum of a set is larger than or equal to the supremum of its subset, then for $\mu, \nu \in \calP(\G)$, from Equations~\eqref{eq:Connection_GST_p1},~\eqref{eq:Connection_GST_p2},~\eqref{eq:Connection_GST_p3}, and consider $c = 2$, then we obtain
    \begin{equation*}
        \calGST_{\Phi}(\mu, \nu) \ge \widehat \calGS(\mu, \nu) \ge \frac{1}{2}\,\calGST_{\Phi}(\mu, \nu).    
    \end{equation*}
    Hence, the proof is complete.
\end{proof}



\subsection{Proof for Proposition~\ref{prop:Connection_GSIMPMR_OW}}\label{app:proof:prop:Connection_GSIMPMR_OW}

\begin{proof}
     For the limit case $\Phi(t) := t$, and graph $\G$ is a tree, then following~\citep[Remark 4.5 and Proposition 4.6]{le2024generalized}, we have
     \begin{equation}\label{eq:GSIMR_OW_tmp1}
         \calGST_{\Phi}(\mu, \nu) = \calOW(\mu, \nu).
     \end{equation}

     Additionally, following Proposition~\ref{prop:Connection_GSIMPMR_GST}, we have
    \begin{equation}\label{eq:GSIMR_OW_tmp2}
    \frac{1}{2} \, \calGST_{\Phi}(\mu, \nu) \le \widehat \calGS_{\Phi}(\mu, \nu) \le \calGST_{\Phi}(\mu, \nu).
    \end{equation}
    Thus, from Equations~\eqref{eq:GSIMR_OW_tmp1}, and~\eqref{eq:GSIMR_OW_tmp2}, we obtain
    \begin{equation*}
    \frac{1}{2} \, \calOW(\mu, \nu) \le \widehat \calGS_{\Phi}(\mu, \nu) \le \calOW(\mu, \nu).
    \end{equation*}
    Hence, the proof is complete.
    
\end{proof}

\subsection{Proof for Lemma~\ref{lem:ineq_f_gradf}}\label{app:proof:lem:ineq_f_gradf}

\begin{proof}
Consider an $N$-function $\Phi$, $f \in {\OrliczSobolevPhiZero}(\G, \lambda)$, a nonnegative weight function $\hat w$, and $t > 0$. Then we have
\begin{eqnarray}
 \hspace{-2em} \int_{\G} \hat w(x) \, \Phi \! \left( \frac{1}{t} \left|f(x)\right| \right) \lambda(\dd x) \hspace{-1.8em} &&= \int_{\G} \hat w(x) \, \Phi\!\left( \frac{1}{t} \left| \int_{\G} \mathbf{1}_{[z_0, x]}(y) f'(y) \lambda(\dd y) \right| \right) \lambda(\dd x) \nonumber\\
&&= \int_{\G} \hat w(x) \, \Phi\!\left( \frac{\lambda(\G)}{t} \left| \frac{1}{\lambda(\G)} \int_{\G} \mathbf{1}_{[z_0, x]}(y) f'(y) \lambda(\dd y) \right| \right) \lambda(\dd x). \label{eq:int_phi}
\end{eqnarray}

For $\alpha >0$, let $\bar{\Phi}_\alpha(s) := \Phi(\alpha s)$ for $s > 0$. Then $\bar{\Phi}_\alpha$ is also a convex, non-decreasing function. Using this function, we can rewrite Equation~\eqref{eq:int_phi} as follows:

\begin{equation*}
\int_{\G} \hat w(x) \, \Phi\!\left( \frac{1}{t} \left|f(x)\right| \right) \lambda(\dd x) = \int_{\G} \hat w(x) \, \bar\Phi_{\frac{\lambda(\G)}{t}} \! \left( \left| \frac{1}{\lambda(\G)} \int_{\G} \mathbf{1}_{[z_0, x]}(y) f'(y) \lambda(\dd y) \right| \right) \lambda(\dd x).
\end{equation*}

Then, by applying Jensen's inequality, we have

\begin{eqnarray*}
\hspace{-2em} \int_{\G} \hat w(x) \, \Phi\!\left( \frac{1}{t} \left|f(x)\right| \right) \lambda(\dd x) \hspace{-1.8em} &&\le \frac{1}{\lambda(\G)} \int_{\G} \int_{\G} \hat w(x) \, \bar\Phi_{\frac{\lambda(\G)}{t}} \! \left( \left| \mathbf{1}_{[z_0, x]}(y) f'(y)\right| \right) \lambda(\dd y) \lambda(\dd x) \\
&&= \frac{1}{\lambda(\G)} \int_{\G} \int_{\G} \hat w(x) \, \Phi \! \left( \frac{\lambda(\G)}{t} \left| \mathbf{1}_{[z_0, x]}(y) f'(y)\right| \right) \lambda(\dd y) \lambda(\dd x).
\end{eqnarray*}

Due to $\Phi(0) = 0$, we can rewrite the above expression as 
\begin{eqnarray*}
\hspace{-1.9em} \int_{\G} \! \hat w(x) \, \Phi\!\left( \frac{1}{t} \left|f(x)\right| \right) \! \lambda(\dd x) \hspace{-1.9em} &&\le \frac{1}{\lambda(\G)} \! \int_{\G} \! \hat w(x) \! \left( \int_{[z_0, x]} \Phi \left( \frac{\lambda(\G)}{t} \left|f'(y)\right| \right) \lambda(\dd y) \right) \! \lambda(\dd x) \\
&&= \frac{1}{\lambda(\G)} \! \int_{\G} \! \hat w(x) \! \left( \int_{\G} \mathbf{1}_{[z_0, x]}(y) \, \Phi \! \left( \frac{\lambda(\G)}{t} \left|f'(y)\right| \right) \lambda(\dd y) \right) \! \lambda(\dd x).
\end{eqnarray*}

Applying Fubini's theorem, we can interchange the order of the integrations. As a consequence, we obtain

\begin{eqnarray*}
\hspace{-1.9em} \int_{\G} \! \hat w(x) \, \Phi\!\left( \frac{1}{t} \left|f(x)\right| \right) \!\lambda(\dd x) 
\hspace{-1.9em} &&\le \frac{1}{\lambda(\G)} \! \int_{\G} \! \left(  \int_{\G} \! \hat w(x) \mathbf{1}_{[z_0, x]}(y) \lambda(\dd x) \right) \Phi \! \left( \frac{\lambda(\G)}{t} \left|f'(y)\right| \right) \! \lambda(\dd y) \\
&&= \frac{1}{\lambda(\G)} \int_{\G} \! \Phi \! \left( \frac{\lambda(\G)}{t} \left|f'(y)\right| \right) \! \bar{\lambda}_{\hat w}(\Lambda(y)) \lambda(\dd y).
\end{eqnarray*}

Hence, the proof is complete.
\end{proof}

\subsection{Proof for Lemma~\ref{lem:bound_fnorm}}\label{app:proof:lem:bound_fnorm}

\begin{proof} The conclusion of the lemma is trivial if $\norm{f}_{L_\Phi} = 0$. Thus we only need to consider the case $\norm{f}_{L_\Phi} > 0$ in the following proof. As a consequence and due to $f(z_0) = 0$, there must be $x\in \G$ such that $|f'(x)| > 0$, i.e. the set $\{x\in \G: |f'(x)|>0\}$ is nonempty.

We will prove the result by contradiction. Specifically, suppose by contradiction that $0< t < \frac{\norm{f}_{L_\Phi}}{\lambda(\G)}$. Then as  $|f'(x)| = 0$ implies $\Phi\!\left(\frac{|f'(x)|}{t} \right) = 0$ and as $\Phi$ is strictly increasing, we have

\begin{eqnarray}
    \int_{\G} w_0(x) \, \Phi\!\left(\frac{|f'(x)|}{t}\right)\lambda(\dd x) &&= \int_{\G, |f'|>0} w_0(x) \, \Phi\!\left(\frac{|f'(x)|}{t}\right)\lambda(\dd x) \nonumber\\
    && > \int_{\G, |f'|>0} w_0(x) \, \Phi\!\left(\frac{\lambda(\G)}{\norm{f}_{L_\Phi}} |f'(x)| \right)\lambda(\dd x) \nonumber\\
    &&= \int_{\G} w_0(x) \, \Phi\!\left(\frac{\lambda(\G)}{\norm{f}_{L_\Phi}} |f'(x)| \right)\lambda(\dd x). \label{eq:ineq_aux}
\end{eqnarray}    

On the other hand, since
\begin{equation*}
\norm{f}_{L_\Phi} = \inf \left\{t > 0 \mid \int_{\G} \Phi\!\left(\frac{|f(x)|}{t}\right)\lambda(\dd x) \le 1 \right\}
\end{equation*}
we deduce that
\begin{equation*}
\int_{\G} \Phi\!\left(\frac{|f(x)|}{ \norm{f}_{L_\Phi}} \right)\lambda(\dd x) = 1.
\end{equation*}

Applying Lemma~\ref{lem:ineq_f_gradf} for $\hat w =1$ and $t= \norm{f}_{L_\Phi}$ to the above left hand side, we have

\begin{eqnarray*}
1 = \int_{\G} \! \Phi \! \left( \frac{|f(x)|}{ \norm{f}_{L_\Phi}} \right) \! \lambda(\dd x) 
&&\le \frac{1}{\lambda(\G)} \int_{\G} \Phi \left( \frac{\lambda(\G)}{\norm{f}_{L_\Phi}} \left|f'(y)\right| \right) \lambda(\Lambda(y)) \lambda(\dd y) \\
&&\le \frac{1}{\lambda(\G)} \int_{\G} \Phi \left( \frac{\lambda(\G)}{\norm{f}_{L_\Phi}} \left|f'(y)\right| \right) w_0(y) \lambda(\dd y).
\end{eqnarray*}
Thus, we obtain
\begin{equation*}
\int_{\G} \Phi \left( \frac{\lambda(\G)}{\norm{f}_{L_\Phi}} \left|f'(x)\right| \right) w_0(x) \lambda(\dd y) \ge \lambda(\G).
\end{equation*}

This together with the inequality in~\eqref{eq:ineq_aux} yields

\begin{equation*}
    \int_{\G} w_0(x) \, \Phi\!\left(\frac{|f'(x)|}{t}\right)\lambda(\dd x) > \lambda(\G),
\end{equation*}
which contradicts the assumption. Hence, the proof is complete.
\end{proof}

\subsection{Proof for Theorem~\ref{thrm:bound_f_gradf}}\label{app:proof:thrm:bound_f_gradf}

\begin{proof}
We consider the Musielak norm with weight function $\frac{w_0}{\lambda(\G)}$ for gradient function $f'$
\begin{equation*}\label{eq:mo_norm}
\norm{f'}_{L_{\Phi}^{{w_0}/{\lambda(\G)}}} = \inf \left\{t > 0 \mid \int_{\G} \frac{w_0(x)}{\lambda(\G)} \, \Phi\!\left(\frac{|f'(x)|}{t}\right)\lambda(\dd x) \le 1 \right\}.
\end{equation*}
In particular,
\[
\int_{\G} w_0(x) \, \Phi\!\left(\frac{|f'(x)|}{\norm{f'}_{L_{\Phi}^{{w_0}/{\lambda(\G)}}}}\right)\lambda(\dd x) \leq  \lambda(\G).
\]

Therefore, by applying  Lemma~\ref{lem:bound_fnorm} with  $t := \norm{f}_{L_{\Phi}^{{w_0}/{\lambda(\G)}}}$, we conclude that
    \[
     \norm{f}_{L_{\Phi}} \le \lambda(\G) \norm{f'}_{L_{\Phi}^{{w_0}/{\lambda(\G)}}}.
    \]
    Hence, the proof is complete.
\end{proof}

\subsection{Proof for Lemma~\ref{lem:different_weight_func}}\label{app:proof:lem:different_weight_func}

\begin{proof}
    For any $N$-function $\Phi$, any Borel measurable function $f$ on $\G$, and scalar $t > 0$, we have
\begin{equation*}
 \int_{\G} w_1(x) \, \Phi\!\left(\frac{|f(x)|}{t}\right)\lambda(\dd x) \ge \int_{\G} w_2(x) \, \Phi\!\left(\frac{|f(x)|}{t}\right)\lambda(\dd x).
\end{equation*}
Therefore, we obtain
\begin{equation*}
\hspace{-1em} \left\{ t>0 \mid \int_{\G} w_1(x) \, \Phi\!\left(\frac{|f(x)|}{t}\right)\lambda(\dd x) \le 1\right\} \subseteq \left\{ t>0 \mid \int_{\G} w_2(x) \, \Phi\!\left(\frac{|f(x)|}{t}\right)\lambda(\dd x) \le 1\right\}.
\end{equation*}
Notice that the infimum of a set is smaller than or equal to the infimum of its subset. As a consequence, we obtain
\begin{equation*}
\norm{f}_{L_{\Phi}^{w_1}} \ge \norm{f}_{L_{\Phi}^{w_2}}.
\end{equation*}

The proof is complete.

\end{proof}

\subsection{Proof for Lemma~\ref{lem:Nfunc_multiplicative}}\label{app:proof:lem:Nfunc_multiplicative}

\begin{proof}
    Notice that $\Phi$ is a convex function and $\Phi(0) = 0$. Therefore, for any $0 < s \le t$, we have
    \begin{eqnarray*}
        \Phi(s) &&= \Phi\!\left(\frac{s}{t}t + \left(1 - \frac{s}{t}\right)0 \right) \\
        &&\le \frac{s}{t} \Phi(t) + \left(1 - \frac{s}{t}\right) \Phi(0) = \frac{s}{t} \Phi(t),
    \end{eqnarray*}
which yields $\frac{\Phi(s)}{s} \le \frac{\Phi(t)}{t}$. Thus, the function $t \mapsto \frac{\Phi(t)}{t}$ is nondecreasing on $(0, +\infty)$.

Consequently, since $k \ge 1$, we get 
\begin{equation*}
\frac{\Phi(t)}{t} \le \frac{\Phi(kt)}{kt}.
\end{equation*}

That is, $k\Phi(t) \le \Phi(kt)$. The proof is complete.
\end{proof}

\subsection{Proof for Lemma~\ref{lem:MusielakNorm_kPhi}}\label{app:proof:lem:MusielakNorm_kPhi}

Following the definition of Musielak norm (Equation~\eqref{eq:weighted_Orlicz_norm}), we have
\begin{eqnarray*}
    \norm{kf}_{L_{\Phi}^{\hat w}} &&= \inf \left\{t > 0 \mid \int_{\G} \hat w(x) \, \Phi\!\left(\frac{|kf(x)|}{t}\right)\lambda(\dd x) \le 1 \right\} \\
    &&= k \inf \left\{\frac{t}{k} > 0 \mid \int_{\G} \hat w(x) \, \Phi\!\left(\frac{|f(x)|}{t/k}\right)\lambda(\dd x) \le 1 \right\} \\
    &&= k \norm{f}_{L_{\Phi}^{\hat w}}.
\end{eqnarray*}
The proof is complete.

\subsection{Proof for Proposition~\ref{prop:RGSI_closed_form}}\label{sec:proof:prop:RGSI_closed_form}

\begin{proof}

Following~\citep[Remark A.1]{le2024generalized}, for $\Phi(t) = \frac{(p-1)^{p-1}}{p^p} t^p$ with $1 < p < \infty$, we have its complementary $N$-function $\Psi(t) = t^q$ where $q$ is the conjugate of $p$, i.e., $1/q + 1/p = 1$. Consequently, we have $L_{\Psi} = L^q$, and $L_{\Psi}^{\hat w} = L_{\hat w}^q$, where $L_{\hat w}^q$ is the weighted $L^q$ space with weight function $\hat w$. 

Additionally, let $\hat f(x) =  \frac{\mu(\Lambda(x)) -  \nu(\Lambda(x))}{\hat{w}(x)}$ for  $ x \in \G$, by following Equation~\eqref{app:eq:dualnorm_regGenSobolevIPM} in the proof of Theorem~\ref{thrm:RGSI_1d_optimization} in \S\ref{sec:proof:thrm:RGSI_1d_optimization}, we have
\begin{align*}
\widehat{\calGS}_{\Phi}(\mu, \nu)  &= 
\sup_{g\in L_{\Psi}(\G, \lambda): \, \norm{g}_{L_{\Psi}^{\hat{w}}} \le 1} \left| \int_{\G} \hat{w}(x) \hat f(x) g(x) \lambda(\mathrm{d}x) \right| \\
 &= 
\sup_{g\in L^q(\G, \lambda): \, \norm{g}_{L^q_{\hat{w}}} \le 1} \left| \int_{\G} \hat{w}(x) \hat f(x) g(x) \lambda(\mathrm{d}x) \right| \\
&= \norm{\hat f}_{L^p_{\hat w}} \\
&= \left[ \int_{\G} \hat w(x) \left| \hat f(x) \right|^p \lambda (\dd x) \right]^{\frac{1}{p}} \\
&= \left[ \int_{\G} \hat w(x)^{1-p} \left| \mu(\Lambda(x)) - \nu(\Lambda(x)) \right|^p \lambda (\dd x) \right]^{\frac{1}{p}}.
\end{align*}

Hence, the proof is complete.

\end{proof}

\subsection{Proof for Proposition~\ref{prop:Connection_GSIMPMR_SIPM}}\label{app:proof:prop:Connection_GSIMPMR_SIPM}

\begin{proof}
    Following Proposition~\ref{prop:Connection_GSIMPMR_RSIPM}, we have
    \begin{equation}\label{eq:GSIMPMR_RSIPM_tmp1}
    \max(1, \lambda(\G)^{-1})^{\frac{1-p}{p}} \, \hat \calS_p(\mu, \nu) \le \widehat \calGS_{\Phi}(\mu, \nu) \le \min(1, \lambda(\G)^{-1})^{\frac{1-p}{p}} \, \hat \calS_p(\mu, \nu).
\end{equation}
Additionally, following~\citep[Theorem 4.2]{le2025scalable}, we have
\begin{equation}\label{eq:RSIMP_SIMP_tmp2}
\left[ \frac{\min(1, \lambda(\G)^{p-1})}{1 + \lambda(\G)^{p}} \right]^{\frac{1}{p}} \, \calS_p(\mu, \nu) \le \hat{\calS}_p(\mu, \nu) \le \left[\max(1,  \lambda(\G)^{p-1})\right]^{\frac{1}{p}} \, \calS_p(\mu, \nu)
\end{equation}
Therefore, from Equations~\eqref{eq:GSIMPMR_RSIPM_tmp1},~\eqref{eq:RSIMP_SIMP_tmp2}, we obtain
\begin{equation*}
    c_1 \, \calS_p(\mu, \nu) \le \widehat \calGS_{\Phi}(\mu, \nu) \le c_2 \, \calS_p(\mu, \nu),
\end{equation*}
where $c_1 = \left[ \frac{\min(1, \lambda(\G)^{p-1}) \max(1, \lambda(\G)^{-1})^{1-p}}{1 + \lambda(\G)^{p}} \right]^{\frac{1}{p}}$; $c_2 = \left[\min(1, \lambda(\G)^{-1})^{1-p} \max(1,  \lambda(\G)^{p-1})\right]^{\frac{1}{p}}$.

Hence, the proof is complete.

\end{proof}

\subsection{Proof for Proposition~\ref{prop:Connection_GSIMPMR_ST}}\label{app:proof:prop:Connection_GSIMPMR_ST}

\begin{proof}
    For $N$-function $\Phi(t) = \frac{(p-1)^{p-1}}{p^p} t^p$ with $1 < p < \infty$, following~\citep[Proposition 4.4]{le2024generalized}, we have
    \begin{equation}\label{eq:GSIMR_ST_tmp1}
        \calGST_{\Phi}(\mu, \nu) = \calST_{p}(\mu, \nu).
    \end{equation}
    Additionally, following Proposition~\ref{prop:Connection_GSIMPMR_GST}, we have
    \begin{equation}\label{eq:GSIMR_ST_tmp2}
    \frac{1}{2} \, \calGST_{\Phi}(\mu, \nu) \le \widehat \calGS_{\Phi}(\mu, \nu) \le \calGST_{\Phi}(\mu, \nu).
    \end{equation}
    Thus, from Equations~\eqref{eq:GSIMR_ST_tmp1}, and~\eqref{eq:GSIMR_ST_tmp2}, we obtain
    \begin{equation*}
    \frac{1}{2} \, \calST_{p}(\mu, \nu) \le \widehat \calGS_{\Phi}(\mu, \nu) \le \calST_{p}(\mu, \nu).
    \end{equation*}
    Hence, the proof is complete.
\end{proof}

\subsection{Proof for Proposition~\ref{prop:Connection_GSIMPMR_OT}}\label{app:proof:prop:Connection_GSIMPMR_OT}

\begin{proof}
     For the limit case $\Phi(t) := t$, and notice that $\lim_{p \to 1^+} \frac{(p-1)^{p-1}}{p^p} t^p = t$~\citep[\S A.7]{le2024generalized}, then following~\citep[Proposition 4.4]{le2024generalized}, by taking the limit $p \to 1^+$, the closed-form of generalized Sobolev transport (GST) is equal to the $1$-order Sobolev transport (ST), i.e.,
     \begin{equation}\label{eq:OT_tmp1}
         \calGST_{\Phi}(\mu, \nu) = \calST_1(\mu, \nu).
     \end{equation}

     Additionally, suppose that graph $\G$ is a tree, then by following~\citet{le2022st}, the $1$-order ST is in turn equal to the $1$-order Wasserstein.
     \begin{equation}\label{eq:OT_tmp2}
         \calST_{1}(\mu, \nu) = \calW_1(\mu, \nu).
     \end{equation}
     
     Therefore, from Equations~\eqref{eq:OT_tmp1}, and~\eqref{eq:OT_tmp2}, we obtain
     \begin{equation}\label{eq:GSIMR_OT_tmp1}
         \calGST_{\Phi}(\mu, \nu) = \calW_1(\mu, \nu).
     \end{equation}

     Moreover, following Proposition~\ref{prop:Connection_GSIMPMR_GST}, we have
    \begin{equation}\label{eq:GSIMR_OT_tmp2}
    \frac{1}{2} \, \calGST_{\Phi}(\mu, \nu) \le \widehat \calGS_{\Phi}(\mu, \nu) \le \calGST_{\Phi}(\mu, \nu).
    \end{equation}
    Thus, from Equations~\eqref{eq:GSIMR_OT_tmp1}, and~\eqref{eq:GSIMR_OT_tmp2}, we obtain
    \begin{equation*}
    \frac{1}{2} \, \calW_1(\mu, \nu) \le \widehat \calGS_{\Phi}(\mu, \nu) \le \calW_1(\mu, \nu).
    \end{equation*}
    Hence, the proof is complete.
    
\end{proof}

\subsection{Compute $A_1$ for $\mathcal{A}_{\Phi_1}$ in $\widehat \calGS_{\Phi_1}$ (\S\ref{app:subsec:thrm:RGSI_1d_optimization_discrete})}\label{subsec:app:compute_A1}

For $a > 0, b > 0$, we consider the term
\begin{equation*}
A_1 := \beta \int_{0}^{1} (at + b) \, \Phi_1\!\left(\frac{\alpha}{at + b}\right)\dd t.
\end{equation*}
Set $u = at + b$, so $\dd u = a \dd t$ and $u \in [b, a+b]$, then we can rewrite
\begin{equation*}
    A_1 = \frac{\beta}{a} \int_{b}^{a+b} \left[u \exp(\alpha/u) - \alpha - u \right] \dd u.
\end{equation*}

Next, we want to compute
\begin{equation*}
    B_{\Phi_1} = \int_{b}^{a+b} u \exp(\alpha/u)\dd u.
\end{equation*}
First, we use the substitution by setting $v = \alpha/u$. Then $u = \alpha/v$ and $\dd u = - \alpha v^{-2} \dd v$.

The integrand becomes 
\begin{equation}\label{eq:app:Bint_p1_tmp1}
    \int u \exp(\alpha/u) \dd u = \int \frac{\alpha}{v} \exp(v) [- \alpha v^{-2} \dd v] = - \alpha^2 \int \exp(v) v^{-3} \dd v. 
\end{equation}

Second, we use two integrations by parts to compute:
\begin{equation}\label{eq:app:B1_p1}
    B_1 := \int \exp(v) v^{-3} \dd v.
\end{equation}

Applying the integration by part, we get
\begin{equation}\label{eq:app:B1_p2}
    B_1 = \frac{1}{2} \int \exp(v) v^{-2} \dd v - \frac{1}{2} \exp(v) v^{-2}.
\end{equation}

Applying the integration by part for the integral in the right hand side, we obtain
\begin{equation}\label{eq:app:B1_p3}
    \int \exp(v) v^{-2} \dd v = \int \exp(v) v^{-1} \dd v - \exp(v) v^{-1}.
\end{equation}

Then, from Equations~\eqref{eq:app:B1_p1},~\eqref{eq:app:B1_p2}, and~\eqref{eq:app:B1_p3}, we obtain
\begin{eqnarray}\label{eq:app:Bint_p2}
    B_1 &&= \frac{1}{2} \left(Ei(v) - \exp(v)v^{-1} \right) - \frac{1}{2} \exp(v) v^{-2} + C \\
    &&= \frac{1}{2} Ei(v) - \frac{1}{2} \exp(v)v^{-1} - \frac{1}{2} \exp(v) v^{-2} + C.
\end{eqnarray}

Thus, from Equations~\eqref{eq:app:Bint_p1_tmp1}, and~\eqref{eq:app:Bint_p2}, and recall that $v = \alpha/u$, we have
    \begin{eqnarray*}
    \int u \exp(\alpha/u) \dd u &&= - \alpha^2 B_1    \\
    &&= -\frac{\alpha^2}{2} Ei(\alpha/u) + \frac{\alpha^2}{2} \frac{\exp(\alpha/u)}{\alpha/u} + \frac{\alpha^2}{2} \frac{\exp(\alpha/u)}{(\alpha/u)^2} + C \\
    &&= \frac{1}{2}\left[u^2 \exp(\alpha/u) + \alpha u \exp(\alpha/u) - \alpha^2 Ei(\alpha/u) \right] + C.
    \end{eqnarray*}

Thus, an antiderivative for the integrand of $A_1$ is
\begin{eqnarray*}
    F_{A_1} &&:= \frac{\beta}{a} \int \left[u \exp(\alpha/u) - \alpha - u \right] \dd u \\
    &&= \frac{\beta}{2a}\left(u^2 \exp(\alpha/u) + \alpha u \exp(\alpha/u) - \alpha^2 Ei(\alpha/u) - 2\alpha u - u^2 \right) + C.
\end{eqnarray*}

Hence, we have
\begin{eqnarray*}
    A_1 = F_{A_1}(a + b) - F_{A_1}(b).
\end{eqnarray*}
Or, we obtain
\begin{eqnarray*}
    A_1 = \frac{\beta}{2a}\Bigg[ -\alpha^2 \left[ Ei\!\left(\frac{\alpha}{a+b}\right) - Ei\!\left(\frac{\alpha}{b}\right) \right] + \alpha\left[(a+b)\exp\!\left(\frac{\alpha}{a+b} \right) - b \exp\!\left(\frac{\alpha}{b} \right)\right] \hspace{2.7em} \nonumber \\
    + \left[(a+b)^2\exp\!\left(\frac{\alpha}{a+b} \right) - b^2 \exp\!\left(\frac{\alpha}{b} \right)\right] - 2\alpha a - \left[(a+b)^2 - b^2 \right] \Bigg].
\end{eqnarray*}
    
Therefore, we have
\begin{equation*}
    \mathcal{A}_{\Phi_1} = A_1,
\end{equation*}
where $\beta = w_e, a = \frac{w_e}{\lambda(\G)}, b = 1 + \frac{\lambda(\gamma_e)}{\lambda(\G)}, \alpha = k|\bar h(e) |$.

Additionally, we next compute the derivative of $A_1$ w.r.t. $k$.

For simplicity, let 
\begin{eqnarray*}
S_1(\alpha) := -\alpha^2 \left[ Ei\!\left(\frac{\alpha}{a+b}\right) - Ei\!\left(\frac{\alpha}{b}\right) \right] + \alpha\left[(a+b)\exp\!\left(\frac{\alpha}{a+b} \right) - b \exp\!\left(\frac{\alpha}{b} \right)\right] \hspace{2.7em} \nonumber \\
    + \left[(a+b)^2\exp\!\left(\frac{\alpha}{a+b} \right) - b^2 \exp\!\left(\frac{\alpha}{b} \right)\right] - 2\alpha a - \left[(a+b)^2 - b^2 \right],
\end{eqnarray*}
then we have
\begin{equation*}
    A_1 = \frac{\beta}{2a}S_1(\alpha).
\end{equation*}

Thus, we have
\begin{eqnarray*}
    \frac{\dd A_1}{\dd k} = \frac{\dd A_1}{\dd \alpha} \frac{\dd \alpha}{\dd k}. 
\end{eqnarray*}
Notice that $\frac{\dd}{\dd z}Ei(z) = \frac{\exp(z)}{z}$, and use the chain rule, we have

For the first term of $S_1$, let $S_{11}(\alpha) := -\alpha^2 \left[ Ei\!\left(\frac{\alpha}{a+b}\right) - Ei\!\left(\frac{\alpha}{b}\right) \right]$, we have
\begin{equation*}
\frac{\dd S_{11}}{\dd \alpha} = -2\alpha \left[ Ei\!\left(\frac{\alpha}{a+b}\right) - Ei\!\left(\frac{\alpha}{b}\right) \right] - \alpha \left[\exp(\alpha/(a+b)) - \exp(\alpha/b) \right].
\end{equation*}

For second term of $S_1$, let $S_{12}(\alpha) := \alpha\left[(a+b)\exp\!\left(\frac{\alpha}{a+b} \right) - b \exp\!\left(\frac{\alpha}{b} \right)\right]$, we have
\begin{equation*}
    \frac{\dd S_{12}}{\dd \alpha} = (a+b)\exp\!\left(\frac{\alpha}{a+b} \right) - b \exp\!\left(\frac{\alpha}{b} \right) + \alpha \left[\exp(\alpha/(a+b)) - \exp(\alpha/b) \right].
\end{equation*}

For the third term of $S_1$, denote $S_{13}(\alpha) := (a+b)^2\exp\!\left(\frac{\alpha}{a+b} \right) - b^2 \exp\!\left(\frac{\alpha}{b} \right)$, we have
\begin{equation*}
    \frac{\dd S_{13}}{\dd \alpha} = (a+b)\exp\!\left(\frac{\alpha}{a+b} \right) - b \exp\!\left(\frac{\alpha}{b} \right)
\end{equation*}

For the fourth term  of $S_1$, we have $\frac{\dd}{\dd \alpha}(-2\alpha a) = -2a$.

The last term of $S_1$ is constant w.r.t. $\alpha$.

Therefore, we obtain
\begin{eqnarray*}
    \frac{\dd}{\dd k}A_1 = \frac{\beta |\bar h(e)|}{a} \left\{ -\alpha \left[ Ei\!\left(\frac{\alpha}{a+b}\right) - Ei\!\left(\frac{\alpha}{b}\right) \right] +(a+b)\exp\!\left(\frac{\alpha}{a+b} \right) - b \exp\!\left(\frac{\alpha}{b} \right) - a \right\}.
\end{eqnarray*}

Furthermore, we next compute the second-order derivative of $A_1$ w.r.t. $k$.
\begin{equation*}
    \frac{\dd^2}{\dd k^2}A_1 = \frac{\beta |\bar h(e)|}{a} \frac{\dd}{\dd k}Q_1(\alpha),
\end{equation*}
where we denote
\begin{equation*}
    Q_1(\alpha) := -\alpha \left[ Ei\!\left(\frac{\alpha}{a+b}\right) - Ei\!\left(\frac{\alpha}{b}\right) \right] +(a+b)\exp\!\left(\frac{\alpha}{a+b} \right) - b \exp\!\left(\frac{\alpha}{b} \right) - a.
\end{equation*}
By using the chain rule, we have
\begin{equation*}
    \frac{\dd^2}{\dd k^2}A_1 = \frac{\beta |\bar h(e)|^2}{a}  \frac{\dd}{\dd \alpha}Q_1(\alpha),
\end{equation*}

For the first term of $Q_1$, denote $Q_{11}(\alpha) := -\alpha \left[ Ei\!\left(\frac{\alpha}{a+b}\right) - Ei\!\left(\frac{\alpha}{b}\right) \right]$, we have
\begin{eqnarray*}
\frac{\dd Q_{11}}{\dd \alpha} &&=  - \left[ Ei\!\left(\frac{\alpha}{a+b}\right) - Ei\!\left(\frac{\alpha}{b}\right) \right] - \alpha \left[\frac{\exp(\alpha/(a+b))}{\alpha} - \frac{\exp(\alpha/b}{\alpha} \right] \\
&&= - \left[ Ei\!\left(\frac{\alpha}{a+b}\right) - Ei\!\left(\frac{\alpha}{b}\right) \right] -  \left[{\exp(\alpha/(a+b))} - {\exp(\alpha/b} \right].
\end{eqnarray*}

For the second term of $Q_1$, we have
\begin{equation*}
    \frac{\dd}{\dd \alpha}\left[ (a+b)\exp\!\left(\frac{\alpha}{a+b} \right) \right] = \exp\!\left(\frac{\alpha}{a+b} \right).
\end{equation*}

For the third term of $Q_1$, we have
\begin{eqnarray*}
    \frac{\dd}{\dd \alpha}\left[ -b\exp(\alpha/b) \right] = - \exp(\alpha/b).
\end{eqnarray*}

The last term of $Q_1$ is a constant w.r.t. $\alpha$.

Therefore, we obtain
\begin{equation*}
    \frac{\dd^2}{\dd k^2}A_1 = - \frac{\beta |\bar h(e)|^2}{a}  \left[ Ei\!\left(\frac{\alpha}{a+b}\right) - Ei\!\left(\frac{\alpha}{b}\right) \right].
\end{equation*}

Hence, we complete the detailed derivation for $A_1$.

\subsection{Compute $A_2$ for $\mathcal{A}_{\Phi_2}$ in $\widehat \calGS_{\Phi_2}$ (\S\ref{app:subsec:thrm:RGSI_1d_optimization_discrete})}\label{subsec:app:compute_A2}

For $a > 0, b > 0$, we consider 
\begin{equation*}
A_2 := \beta \int_{0}^{1} (at + b) \, \Phi_2\!\left(\frac{\alpha}{at + b}\right)\dd t.
\end{equation*}

Set $u = at + b$, so $\dd u = a \dd t$ and $u \in [b, a+b]$, then we can rewrite
\begin{equation*}
    A_2 = \frac{\beta}{a} \int_{b}^{a+b} \left[u \exp(\alpha^2/u^2) - u \right] \dd u.
\end{equation*}

Next, we want to compute
\begin{equation*}
    B_{\Phi_2} = \int_{b}^{a+b} u \exp(\alpha^2/u^2)\dd u.
\end{equation*}
First, we use the substitution by setting $v = \alpha^2/u^2$, so $v > 0$ on the interval. Then $\dd v = - 2 \alpha^2 u^{-3} \dd u$, or $\dd u = - \frac{u^3}{2\alpha^2} \dd v$.

The integrand becomes 
\begin{equation*}
    \int u \exp(\alpha^2/u^2) \dd u = - \int \frac{1}{2\alpha^2} u^4 \exp(v) \dd v = - \frac{\alpha^2}{2} \int  \exp(v) v^{-2} \dd v,
\end{equation*}
since we have $u^4 = \alpha^4 v^{-2}$.

Second, following Equation~\eqref{eq:app:B1_p3}, we have
\begin{equation*}
    B_2 := \int \exp(v) v^{-2} \dd v = Ei(v) - \exp(v)v^{-1} + C.
\end{equation*}
Thus, we obtain
\begin{eqnarray*}
    \int u \exp(\alpha^2/u^2) \dd u &&= -\frac{\alpha^2}{2} \left( Ei(v) - \exp(v)v^{-1} \right) + C \\
    &&= \frac{1}{2} \left[ u^2 \exp(\alpha^2/u^2) - \alpha^2Ei(\alpha^2 / u^2)\right] + C,    
\end{eqnarray*}
since $v = \alpha^2/u^2$, and consequently $\exp(v)v^{-1} = (u^2 / \alpha^2) \exp(\alpha^2/u^2) $.

Thus, an antiderivative for the integrand of $A_2$ is
\begin{eqnarray*}
    F_{A_2} &&:= \frac{\beta}{a} \int \left[u \exp(\alpha^2/u^2) - u \right] \dd u \\
    &&= \frac{\beta}{a}\left(\frac{1}{2} \left[ u^2 \exp(\alpha^2/u^2) - \alpha^2Ei(\alpha^2 / u^2)\right] - \frac{u^2}{2}\right) + C.
\end{eqnarray*}

Hence, we have
\begin{eqnarray*}
    A_2 = F_{A_2}(a + b) - F_{A_2}(b).
\end{eqnarray*}
Or, we obtain
\begin{eqnarray*}
    A_2 = \frac{\beta}{2a}\Bigg[ -\alpha^2 \left[ Ei\!\left(\frac{\alpha^2}{(a+b)^2}\right) - Ei\!\left(\frac{\alpha^2}{b^2}\right) \right] \hspace{14em} \nonumber \\
    + \left[(a+b)^2\exp\!\left(\frac{\alpha^2}{(a+b)^2} \right) - b^2 \exp\!\left(\frac{\alpha^2}{b^2} \right)\right] - \left[(a+b)^2 - b^2 \right] \Bigg].
\end{eqnarray*}
    
Therefore, we have
\begin{equation*}
    \mathcal{A}_{\Phi_2} = A_2,
\end{equation*}
where $\beta = w_e, a = \frac{w_e}{\lambda(\G)}, b = 1 + \frac{\lambda(\gamma_e)}{\lambda(\G)}, \alpha = k|\bar h(e) |$.

Additionally, we next compute the derivative of $A_2$ w.r.t. $k$.

For simplicity, let 
\begin{eqnarray*}
S_2(\alpha) := -\alpha^2 \left[ Ei\!\left(\frac{\alpha^2}{(a+b)^2}\right) - Ei\!\left(\frac{\alpha^2}{b^2}\right) \right] \hspace{14em} \nonumber \\
    + \left[(a+b)^2\exp\!\left(\frac{\alpha^2}{(a+b)^2} \right) - b^2 \exp\!\left(\frac{\alpha^2}{b^2} \right)\right] - \left[(a+b)^2 - b^2 \right],
\end{eqnarray*}
then we have
\begin{equation*}
    A_2 = \frac{\beta}{2a}S_2(\alpha).
\end{equation*}

Thus, we have
\begin{eqnarray*}
    \frac{\dd A_2}{\dd k} = \frac{\dd A_2}{\dd \alpha} \frac{\dd \alpha}{\dd k}. 
\end{eqnarray*}
Notice that $\frac{\dd}{\dd z}Ei(z) = \frac{\exp(z)}{z}$, and use the chain rule, we have

\begin{eqnarray*}
    \frac{\dd S_2}{\dd \alpha} &&= -2\alpha\left[ Ei\!\left(\frac{\alpha^2}{(a+b)^2}\right) - Ei\!\left(\frac{\alpha^2}{b^2}\right) \right] - \alpha^2 \left[ \frac{2}{\alpha}\exp(\alpha^2/(a+b)^2) - \frac{2}{\alpha}\exp(\alpha^2/b^2)\right] \nonumber \\
    && \hspace{8em} + 2\alpha\left[ \exp(\alpha^2/(a+b)^2) - \exp(\alpha^2/b^2)\right] \nonumber \\
    &&= -2\alpha\left[ Ei\!\left(\frac{\alpha^2}{(a+b)^2}\right) - Ei\!\left(\frac{\alpha^2}{b^2}\right) \right].
\end{eqnarray*}
Therefore, we obtain
\begin{eqnarray*}
    \frac{\dd A_2}{\dd k} = -\frac{\beta\alpha|\bar{h}(e)|}{a}\left[ Ei\!\left(\frac{\alpha^2}{(a+b)^2}\right) - Ei\!\left(\frac{\alpha^2}{b^2}\right) \right].
\end{eqnarray*}

Furthermore, we next compute the second-order derivative of $A_2$ w.r.t. $k$.
\begin{equation*}
    \frac{\dd^2}{\dd k^2}A_2 = -\frac{\beta |\bar h(e)|}{a} \frac{\dd}{\dd k}Q_2(\alpha),
\end{equation*}
where we denote
\begin{equation*}
    Q_2(\alpha) := \alpha \left[ Ei\!\left(\frac{\alpha^2}{(a+b)^2}\right) - Ei\!\left(\frac{\alpha^2}{b^2}\right) \right].
\end{equation*}
By using the chain rule, we have
\begin{eqnarray*}
    \frac{\dd^2}{\dd k^2}A_2 &&= - \frac{\beta |\bar h(e)|}{a}  \Bigg\{ |\bar h(e)| \left[ Ei\!\left(\frac{\alpha^2}{(a+b)^2}\right) - Ei\!\left(\frac{\alpha^2}{b^2}\right) \right] \nonumber \\
    && \hspace{5em} + \alpha  \left[ \frac{2}{\alpha} \exp(\alpha^2/(a+b)^2) - \frac{2}{\alpha} \exp(\alpha^2/b^2)  \right] |\bar h(e)| \Bigg\}.
\end{eqnarray*}
Therefore, we obtain
\begin{eqnarray*}
    \frac{\dd^2}{\dd k^2}A_2 = - \frac{\beta |\bar h(e)|^2}{a}  \left[ Ei\!\left(\frac{\alpha^2}{(a+b)^2}\right) \! - Ei\!\left(\frac{\alpha^2}{b^2}\right) \! + 2 \exp(\alpha^2/(a+b)^2) - 2 \exp(\alpha^2/b^2) \right].
\end{eqnarray*}

Hence, we complete the detailed derivation for $A_2$.


\end{document}

%% file: comments.tex
\makeatletter
\def\munderbar#1{\underline{\sbox\tw@{$#1$}\dp\tw@\z@\box\tw@}}
\makeatother

\newcommand{\be}{\begin{equation}}
\newcommand{\ee}{\end{equation}}
\newcommand{\bea}{\begin{equation*}\begin{aligned}}
\newcommand{\eea}{\end{aligned}\end{equation*}}

\newcommand{\R}{\mathbb{R}}

\newcommand{\G}{{\mathbb G}}

\newcommand{\calP}{{\mathcal P}}

\newcommand{\calS}{{\mathcal S}}
\newcommand{\calB}{{\mathcal B}}

\newcommand{\calW}{{\mathcal W}}

\newcommand{\calF}{{\mathcal F}}

\newcommand{\dd}{\mathrm{d}}

\newcommand{\norm}[1]{\left\lVert#1\right\rVert}

%% file: checklist.tex
\section*{NeurIPS Paper Checklist}

\begin{enumerate}

\item {\bf Claims}
    \item[] Question: Do the main claims made in the abstract and introduction accurately reflect the paper's contributions and scope?
    \item[] Answer: \answerYes{} 
    \item[] Justification: Please see \S\ref{sec:introduction} for the main claims of the paper's contributions and scope. 

\item {\bf Limitations}
    \item[] Question: Does the paper discuss the limitations of the work performed by the authors?
    \item[] Answer: \answerYes{} 
    \item[] Justification: Please see the discussion in \S\ref{sec:conclusion} and further discussion in Appendix \S\ref{sec:app:further_discussions}.

\item {\bf Theory assumptions and proofs}
    \item[] Question: For each theoretical result, does the paper provide the full set of assumptions and a complete (and correct) proof?
    \item[] Answer: \answerYes{} 
    \item[] Justification: Please see theoretical results in \S\ref{sec:GenSobolevIPM} and \S\ref{sec:Properties_GenSobolevIPM}; and further theoretical results in Appendix~\S\ref{app:sec:further_theoretical_results}. We provide full proofs in Appendix~\S\ref{app:sec:proofs}. 

    \item {\bf Experimental result reproducibility}
    \item[] Question: Does the paper fully disclose all the information needed to reproduce the main experimental results of the paper to the extent that it affects the main claims and/or conclusions of the paper (regardless of whether the code and data are provided or not)?
    \item[] Answer: \answerYes{} 
    \item[] Justification: We provide code in the supplementary. Our experiments are described in \S\ref{sec:experiments}, and further details are provided in Appendix \S\ref{sec:app:further_exp_details_empirical_results}. 

\item {\bf Open access to data and code}
    \item[] Question: Does the paper provide open access to the data and code, with sufficient instructions to faithfully reproduce the main experimental results, as described in supplemental material?
    \item[] Answer: \answerYes{} 
    \item[] Justification: We provide code in the supplementary. We use available data for the experiments in \S\ref{sec:experiments}.

\item {\bf Experimental setting/details}
    \item[] Question: Does the paper specify all the training and test details (e.g., data splits, hyperparameters, how they were chosen, type of optimizer) necessary to understand the results?
    \item[] Answer: \answerYes{} 
    \item[] Justification: Our experiments are described in \S\ref{sec:experiments}, and further details are provided in Appendix \S\ref{sec:app:further_exp_details_empirical_results}. 

\item {\bf Experiment statistical significance}
    \item[] Question: Does the paper report error bars suitably and correctly defined or other appropriate information about the statistical significance of the experiments?
    \item[] Answer: \answerYes{} 
    \item[] Justification: Please see empirical results presented in \S\ref{sec:experiments}, and further results in Appendix \S\ref{sec:app:further_exp_details_empirical_results}. 

\item {\bf Experiments compute resources}
    \item[] Question: For each experiment, does the paper provide sufficient information on the computer resources (type of compute workers, memory, time of execution) needed to reproduce the experiments?
    \item[] Answer: \answerYes{} 
    \item[] Justification: Please see \S\ref{sec:experiments}, and further details in Appendix \S\ref{sec:app:further_exp_details_empirical_results} and \S\ref{sec:app:further_discussions}. 
    
\item {\bf Code of ethics}
    \item[] Question: Does the research conducted in the paper conform, in every respect, with the NeurIPS Code of Ethics \url{https://neurips.cc/public/EthicsGuidelines}?
    \item[] Answer:  \answerYes{} 
    \item[] Justification: This paper presents work whose goal is to advance the field of Machine Learning. 

\item {\bf Broader impacts}
    \item[] Question: Does the paper discuss both potential positive societal impacts and negative societal impacts of the work performed?
    \item[] Answer: \answerYes{} 
    \item[] Justification: Please see Appendix~\S\ref{sec:app:further_discussions}. 
    
\item {\bf Safeguards}
    \item[] Question: Does the paper describe safeguards that have been put in place for responsible release of data or models that have a high risk for misuse (e.g., pre-trained language models, image generators, or scraped datasets)?
    \item[] Answer: \answerNA{} 
    \item[] Justification: There are no such risks. 

\item {\bf Licenses for existing assets}
    \item[] Question: Are the creators or original owners of assets (e.g., code, data, models), used in the paper, properly credited and are the license and terms of use explicitly mentioned and properly respected?
    \item[] Answer: \answerYes{} 
    \item[] Justification: We cite original papers for codes used in our work. 

\item {\bf New assets}
    \item[] Question: Are new assets introduced in the paper well documented and is the documentation provided alongside the assets?
    \item[] Answer: \answerNA{} 
    \item[] Justification: \answerNA{} 

\item {\bf Crowdsourcing and research with human subjects}
    \item[] Question: For crowdsourcing experiments and research with human subjects, does the paper include the full text of instructions given to participants and screenshots, if applicable, as well as details about compensation (if any)? 
    \item[] Answer: \answerNA{} 
    \item[] Justification: \answerNA{} 

\item {\bf Institutional review board (IRB) approvals or equivalent for research with human subjects}
    \item[] Question: Does the paper describe potential risks incurred by study participants, whether such risks were disclosed to the subjects, and whether Institutional Review Board (IRB) approvals (or an equivalent approval/review based on the requirements of your country or institution) were obtained?
    \item[] Answer: \answerNA{} 
    \item[] Justification: \answerNA{} 

\item {\bf Declaration of LLM usage}
    \item[] Question: Does the paper describe the usage of LLMs if it is an important, original, or non-standard component of the core methods in this research? Note that if the LLM is used only for writing, editing, or formatting purposes and does \emph{not} impact the core methodology, scientific rigor, or originality of the research, declaration is not required.
    \item[] Answer: \answerNA{} 
    \item[] Justification: We only use LLM for editing. 

\end{enumerate}

%% file: bibSobolev22.bib
@inproceedings{choromanski2023taming,
  title={Taming graph kernels with random features},
  author={Choromanski, Krzysztof Marcin},
  booktitle={International Conference on Machine Learning},
  pages={5964--5977},
  year={2023},
}

@inproceedings{choromanski2023efficient,
  title={Efficient graph field integrators meet point clouds},
  author={Choromanski, Krzysztof Marcin and Sehanobish, Arijit and Lin, Han and Zhao, Yunfan and Berger, Eli and Parshakova, Tetiana and Pan, Alvin and Watkins, David and Zhang, Tianyi and Likhosherstov, Valerii and others},
  booktitle={International Conference on Machine Learning},
  pages={5978--6004},
  year={2023},
}

@inproceedings{
le2025scalable,
title={Scalable {S}obolev {IPM} for Probability Measures on a Graph},
author={Tam Le and Truyen Nguyen and Hideitsu Hino and Kenji Fukumizu},
booktitle={International Conference on Machine Learning},
year={2025},
}

@article{tran2026treeEPT,
  title={Tree-sliced entropy partial transport},
  author={Tran, Viet-Hoang and Tran, Thanh and Chu, Thanh and Le, Tam and Nguyen, Tan},
  journal={Advances in Neural Information Processing Systems},
  volume={38},
  pages={16503--16560},
  year={2026}
}

@inproceedings{tran2025nonlinear,
title={Tree-Sliced {W}asserstein Distance with Nonlinear Projection},
author={Tran, Thanh and Tran, Hoang and Chu, Thanh and Pham, Trang and Ghaoui, Laurent El and Le, Tam and Nguyen, Tan},
booktitle={Forty-second International Conference on Machine Learning},
year={2025},
}

@inproceedings{tran2025spherical,
title={Spherical Tree-Sliced {W}asserstein Distance},
author={Tran, Hoang and Chu, Thanh and Nguyen, Khoi and Pham, Trang and Le, Tam and Nguyen, Tan},
booktitle={The Thirteenth International Conference on Learning Representations},
year={2025},
}

@book{krasnoselskii1961convex,
  title={Convex functions and {O}rlicz spaces},
  author={Krasnoselskii, M. A. and Rutickii, Ya. B.},
  year={1961},
  publisher={P. Noordhoff Ltd., Groningen}
}

@book{harjulehto2019generalized,
  title={Orlicz Spaces and Generalized {O}rlicz Spaces},
  author={Harjulehto, Petteri and H{\"a}st{\"o}, Peter},
  year={2019},
  publisher={Springer}
}


%% file: bibSobolev25.bib
@article{Sriperumbudur2012OnTE,
  title={On the empirical estimation of integral probability metrics},
  author={Sriperumbudur, Bharath K. and Fukumizu, Kenji and Gretton, Arthur and Sch{\"o}lkopf, Bernhard and Lanckriet, Gert R. G.},
  journal={Electronic Journal of Statistics},
  volume={6},
  pages={1550--1599},
  year={2012},
}

@article{sando2025tree,
  title={Tree structure for the categorical {W}asserstein {W}eisfeiler-{L}ehman graph kernel},
  author={Sando, Keishi and Le, Tam and Hino, Hideitsu},
  journal={Transactions on Machine Learning Research},
  year={2025}
}

@inproceedings{tran2026revisiting,
  title={Revisiting tree-sliced {W}asserstein distance through the lens of the {F}ermat--{W}eber problem},
  author={Tran, Viet-Hoang and Tran, Thanh Q and Chu, Thanh and Tran, Trung-Khang and Pham, Duy-Tung and Le, Tam and Nguyen, Tan},
  booktitle={International Conference on Learning Representations},
  volume={2026},
  pages={62831--62860},
  year={2026}
}

@article{le2026efficient,
  title={An Efficient {O}rlicz-{S}obolev Approach for Transporting Unbalanced Measures on a Graph},
  author={Le, Tam and Nguyen, Truyen and Hino, Hideitsu and Fukumizu, Kenji},
  journal={Advances in Neural Information Processing Systems},
  volume={38},
  pages={81586--81627},
  year={2026}
}

@inproceedings{zhang2026graph,
  title={Graph random features for scalable Gaussian processes},
  author={Zhang, Matthew and Lin, Jihao Andreas and Choromanski, Krzysztof and Weller, Adrian and Turner, Richard E and Reid, Isaac},
  booktitle={International Conference on Learning Representations},
  volume={2026},
  pages={45157--45180},
  year={2026}
}

@article{choromanski2026near,
  title={Near-Linear Time Generalized Sinkhorn Algorithms for Bounded Genus Graphs},
  author={Choromanski, Krzysztof and Long, Derek and Parashar, Ananya and Saha, Dwaipayan},
  journal={arXiv preprint arXiv:2605.09782},
  year={2026}
}

@inproceedings{nguyen2024sliced,
title={Sliced {W}asserstein with Random-Path Projecting Directions},
author={Khai Nguyen and Shujian Zhang and Tam Le and Nhat Ho},
booktitle={Forty-first International Conference on Machine Learning},
year={2024},
}

@inproceedings{tran2025treesliced,
title={Tree-Sliced Wasserstein Distance: A Geometric Perspective},
author={Hoang V. Tran and Huyen Trang Pham and Tho Tran Huu and Minh-Khoi Nguyen-Nhat and Thanh Chu and Tam Le and Tan Minh Nguyen},
booktitle={Forty-second International Conference on Machine Learning},
year={2025},
}

@inproceedings{tran2025distancebased,
title={Distance-Based Tree-Sliced Wasserstein Distance},
author={Hoang V. Tran and Minh-Khoi Nguyen-Nhat and Huyen Trang Pham and Thanh Chu and Tam Le and Tan Minh Nguyen},
booktitle={The Thirteenth International Conference on Learning Representations},
year={2025},
}

@inproceedings{tran2026treesliced,
title={Tree-sliced {S}obolev {IPM}},
author={Viet-Hoang Tran and Thanh Tran and Thanh Chu and Duy-Tung Pham and Trung-Khang Tran and Tam Le and Tan Minh Nguyen},
booktitle={The Fourteenth International Conference on Learning Representations},
year={2026},
}

@article{bai2025fused,
  title={Fused Partial Gromov-Wasserstein for Structured Objects},
  author={Bai, Yikun and Tran, Huy and Du, Hengrong and Liu, Xinran and Kolouri, Soheil},
  journal={arXiv preprint arXiv:2502.09934},
  year={2025}
}

@inproceedings{brogat2022learning,
  title={Learning to predict graphs with fused Gromov-Wasserstein barycenters},
  author={Brogat-Motte, Luc and Flamary, R{\'e}mi and Brouard, C{\'e}line and Rousu, Juho and d’Alch{\'e}-Buc, Florence},
  booktitle={International Conference on Machine Learning},
  pages={2321--2335},
  year={2022},
  organization={PMLR}
}

@inproceedings{xu2019gromov,
  title={{G}romov-{W}asserstein learning for graph matching and node embedding},
  author={Xu, Hongteng and Luo, Dixin and Zha, Hongyuan and Duke, Lawrence Carin},
  booktitle={International conference on machine learning},
  pages={6932--6941},
  year={2019},
  organization={PMLR}
}

@inproceedings{le2024generalized,
title={Generalized {S}obolev Transport for Probability Measures on a Graph},
author={Tam Le and Truyen Nguyen and Kenji Fukumizu},
booktitle={Forty-first International Conference on Machine Learning},
year={2024},
}

@article{borgwardt2020graph,
  title={Graph kernels: State-of-the-art and future challenges},
  author={Borgwardt, Karsten and Ghisu, Elisabetta and Llinares-L{\'o}pez, Felipe and O’Bray, Leslie and Rieck, Bastian and others},
  journal={Foundations and Trends{\textregistered} in Machine Learning},
  volume={13},
  number={5-6},
  pages={531--712},
  year={2020},
  publisher={Now Publishers, Inc.}
}

@article{petric2019got,
  title={{GOT}: {A}n optimal transport framework for graph comparison},
  author={Petric Maretic, Hermina and El Gheche, Mireille and Chierchia, Giovanni and Frossard, Pascal},
  journal={Advances in Neural Information Processing Systems},
  volume={32},
  year={2019}
}

@article{dong2020copt,
  title={{COPT}: {C}oordinated optimal transport on graphs},
  author={Dong, Yihe and Sawin, Will},
  journal={Advances in Neural Information Processing Systems},
  volume={33},
  pages={19327--19338},
  year={2020}
}

@inproceedings{Kolouri2020Sliced,
title={Sliced {C}ramer Synaptic Consolidation for Preserving Deeply Learned Representations},
author={Soheil Kolouri and Nicholas A. Ketz and Andrea Soltoggio and Praveen K. Pilly},
booktitle={International Conference on Learning Representations},
year={2020},
}

@article{nadjahi2020statistical,
  title={Statistical and topological properties of sliced probability divergences},
  author={Nadjahi, Kimia and Durmus, Alain and Chizat, L{\'e}na{\"\i}c and Kolouri, Soheil and Shahrampour, Shahin and Simsekli, Umut},
  journal={Advances in Neural Information Processing Systems},
  volume={33},
  pages={20802--20812},
  year={2020}
}

@article{gretton2012kernel,
  title={A kernel two-sample test},
  author={Gretton, Arthur and Borgwardt, Karsten M and Rasch, Malte J and Sch{\"o}lkopf, Bernhard and Smola, Alexander},
  journal={The Journal of Machine Learning Research},
  volume={13},
  number={1},
  pages={723--773},
  year={2012},
}

@inproceedings{hua2018pointwise,
  title={Pointwise convolutional neural networks},
  author={Hua, Binh-Son and Tran, Minh-Khoi and Yeung, Sai-Kit},
  booktitle={Proceedings of the IEEE conference on computer vision and pattern recognition},
  pages={984--993},
  year={2018}
}

@article{wang2019dynamic,
  title={Dynamic graph {CNN} for learning on point clouds},
  author={Wang, Yue and Sun, Yongbin and Liu, Ziwei and Sarma, Sanjay E and Bronstein, Michael M and Solomon, Justin M},
  journal={ACM Transactions on Graphics (TOG)},
  volume={38},
  number={5},
  pages={1--12},
  year={2019},
}

@article{yurochkin2019hierarchical,
  title={Hierarchical optimal transport for document representation},
  author={Yurochkin, Mikhail and Claici, Sebastian and Chien, Edward and Mirzazadeh, Farzaneh and Solomon, Justin M},
  journal={Advances in neural information processing systems},
  volume={32},
  year={2019}
}

@article{uppal2020robust,
  title={Robust density estimation under {B}esov {IPM} losses},
  author={Uppal, Ananya and Singh, Shashank and P{\'o}czos, Barnab{\'a}s},
  journal={Advances in Neural Information Processing Systems},
  volume={33},
  pages={5345--5355},
  year={2020}
}

@article{uppal2019nonparametric,
  title={Nonparametric density estimation \& convergence rates for {GANs} under {B}esov {IPM} losses},
  author={Uppal, Ananya and Singh, Shashank and P{\'o}czos, Barnab{\'a}s},
  journal={Advances in neural information processing systems},
  volume={32},
  year={2019}
}

@article{liang2019estimating,
  title={Estimating certain integral probability metric ({IPM}) is as hard as estimating under the {IPM}},
  author={Liang, Tengyuan},
  journal={arXiv preprint},
  year={2019}
}

@article{singh2018nonparametric,
  title={Nonparametric density estimation under adversarial losses},
  author={Singh, Shashank and Uppal, Ananya and Li, Boyue and Li, Chun-Liang and Zaheer, Manzil and P{\'o}czos, Barnab{\'a}s},
  journal={Advances in Neural Information Processing Systems},
  volume={31},
  year={2018}
}

@article{liang2017well_arXiv,
  title={How well can generative adversarial networks learn densities: A nonparametric view},
  author={Liang, Tengyuan},
  journal={arXiv preprint},
  year={2017}
}

@article{liang2021well_jmlr,
  title={How well generative adversarial networks learn distributions},
  author={Liang, Tengyuan},
  journal={Journal of Machine Learning Research},
  volume={22},
  number={228},
  pages={1--41},
  year={2021}
}

@article{sturm2011generalized,
  title={Generalized {O}rlicz spaces and {W}asserstein distances for convex--concave scale functions},
  author={Sturm, Karl-Theodor},
  journal={Bulletin des sciences math{\'e}matiques},
  volume={135},
  number={6-7},
  pages={795--802},
  year={2011},
  publisher={Elsevier}
}

@book{vershynin2018high,
  title={High-dimensional probability: An introduction with applications in data science},
  author={Vershynin, Roman},
  volume={47},
  year={2018},
  publisher={Cambridge university press}
}

@phdthesis{chewi2023optimization,
  title={An optimization perspective on log-concave sampling and beyond},
  author={Chewi, Sinho},
  year={2023},
  school={Massachusetts Institute of Technology}
}

@article{altschuler2023faster,
  title={Faster high-accuracy log-concave sampling via algorithmic warm starts},
  author={Altschuler, Jason M and Chewi, Sinho},
  journal={Journal of the ACM},
  volume={71},
  number={3},
  pages={1--55},
  year={2024},
}

@inproceedings{le2023scalable,
  title={Scalable Unbalanced {S}obolev Transport for Measures on a Graph},
  author={Le, Tam and Nguyen, Truyen and Fukumizu, Kenji},
  booktitle={International Conference on Artificial Intelligence and Statistics},
  pages={8521--8560},
  year={2023},
  organization={PMLR}
}

@article{kell2017interpolation,
  title={On interpolation and curvature via {W}asserstein geodesics},
  author={Kell, Martin},
  journal={Advances in Calculus of Variations},
  volume={10},
  number={2},
  pages={125--167},
  year={2017},
  publisher={De Gruyter}
}

@article{chamakh2020orlicz,
  title={{O}rlicz random {F}ourier features},
  author={Chamakh, Linda and Gobet, Emmanuel and Szab{\'o}, Zolt{\'a}n},
  journal={The Journal of Machine Learning Research},
  volume={21},
  number={1},
  pages={5739--5775},
  year={2020},
  publisher={JMLRORG}
}

@inproceedings{andoni2018subspace,
  title={Subspace embedding and linear regression with {O}rlicz norm},
  author={Andoni, Alexandr and Lin, Chengyu and Sheng, Ying and Zhong, Peilin and Zhong, Ruiqi},
  booktitle={International Conference on Machine Learning},
  pages={224--233},
  year={2018},
  organization={PMLR}
}

@article{song2019efficient,
  title={Efficient symmetric norm regression via linear sketching},
  author={Song, Zhao and Wang, Ruosong and Yang, Lin and Zhang, Hongyang and Zhong, Peilin},
  journal={Advances in Neural Information Processing Systems},
  volume={32},
  year={2019}
}

@article{deng2022fast,
  title={Fast distance oracles for any symmetric norm},
  author={Deng, Yichuan and Song, Zhao and Weinstein, Omri and Zhang, Ruizhe},
  journal={Advances in Neural Information Processing Systems},
  volume={35},
  pages={7304--7317},
  year={2022}
}

@article{lorenz2022orlicz,
  title={{O}rlicz space regularization of continuous optimal transport problems},
  author={Lorenz, Dirk and Mahler, Hinrich},
  journal={Applied Mathematics \& Optimization},
  volume={85},
  number={2},
  pages={14},
  year={2022},
  publisher={Springer}
}

@inproceedings{GuhaHN23,
  author       = {Aritra Guha and
                  Nhat Ho and
                  XuanLong Nguyen},
  title        = {On Excess Mass Behavior in {G}aussian Mixture Models with {O}rlicz-{W}asserstein
                  Distances},
  booktitle    = {International Conference on Machine Learning, {ICML}},
  volume       = {202},
  pages        = {11847--11870},
  publisher    = {{PMLR}},
  year         = {2023},
}

@article{rao1991theory,
  title={Theory of {O}rlicz spaces},
  author={Rao, Malempati Madhusudana and Ren, Zhong Dao},
  journal={Marcel Dekker},
  year={1991}
}

@book{musielak2006orlicz,
  title={Orlicz spaces and modular spaces},
  author={Musielak, Julian},
  volume={1034},
  year={2006},
  publisher={Springer}
}

@book{adams2003sobolev,
  title={Sobolev spaces},
  author={Adams, Robert A and Fournier, John JF},
  year={2003},
  publisher={Elsevier}
}

@inproceedings{nguyen2021point,
  title={Point-set distances for learning representations of 3D point clouds},
  author={Nguyen, Trung and Pham, Quang-Hieu and Le, Tam and Pham, Tung and Ho, Nhat and Hua, Binh-Son},
  booktitle={Proceedings of the IEEE/CVF International Conference on Computer Vision (ICCV)},
  pages={10478--10487},
  year={2021}
}

@article{sriperumbudur2009integral,
  title={On integral probability metrics, $\phi$-divergences and binary classification},
  author={Sriperumbudur, Bharath K and Fukumizu, Kenji and Gretton, Arthur and Sch{\"o}lkopf, Bernhard and Lanckriet, Gert RG},
  journal={arXiv preprint},
  year={2009}
}

@article{muller1997integral,
  title={Integral probability metrics and their generating classes of functions},
  author={M{\"u}ller, Alfred},
  journal={Advances in Applied Probability},
  volume={29},
  number={2},
  pages={429--443},
  year={1997},
  publisher={Cambridge University Press}
}

@inproceedings{latecki2000shape,
  title={Shape descriptors for non-rigid shapes with a single closed contour},
  author={Latecki, Longin Jan and Lakamper, Rolf and Eckhardt, T},
  booktitle={Proceedings of the IEEE Conference on Computer Vision and Pattern Recognition (CVPR)},
  volume={1},
  pages={424--429},
  year={2000}
}

@article{edelsbrunner2008persistent,
  title={Persistent homology - {A} survey},
  author={Edelsbrunner, Herbert and Harer, John},
  journal={Contemporary Mathematics},
  volume={453},
  pages={257--282},
  year={2008}
  }

@article{adams2017persistence,
  title={Persistence images: A stable vector representation of persistent homology},
  author={Adams, Henry and Emerson, Tegan and Kirby, Michael and Neville, Rachel and Peterson, Chris and Shipman, Patrick and Chepushtanova, Sofya and Hanson, Eric and Motta, Francis and Ziegelmeier, Lori},
  journal={Journal of Machine Learning Research},
  volume={18},
  number={1},
  pages={218--252},
  year={2017}
}

@inproceedings{le2021ept,
  title={Entropy Partial Transport with Tree Metrics: Theory and Practice},
  author={Le, Tam and Nguyen, Truyen},
  booktitle={International Conference on Artificial Intelligence and Statistics (AISTATS)},
  pages={3835--3843},
  year={2021},
}

@article{bonneel2015sliced,
  title={Sliced and radon wasserstein barycenters of measures},
  author={Bonneel, Nicolas and Rabin, Julien and Peyr{\'e}, Gabriel and Pfister, Hanspeter},
  journal={Journal of Mathematical Imaging and Vision},
  volume={51},
  number={1},
  pages={22--45},
  year={2015},
}

@Inproceedings{Cuturi-2013-Sinkhorn,
  	Title 			  = {Sinkhorn distances: {L}ightspeed computation of optimal transport},
  	Author 			  = {Cuturi, M. },
  	Booktitle 		= {Advances in Neural Information Processing Systems},
  	Pages 			  = {2292-2300},
  	Year 			    = {2013}
}

@inproceedings{Mroueh-2018-Sobolev,
    Title         = {Sobolev {GAN}},
    Author        = {Youssef Mroueh and Chun-Liang Li and Tom Sercu and Anant Raj and and Yu Cheng},
    Booktitle       = {International Conference on Learning Representations},
  		Pages 			= {5767-5777},
  		Year            = {2018}
}

@article{hertzsch2007dna,
  title={DNA microarrays: design principles for maximizing ergodic, chaotic mixing},
  author={Hertzsch, Jan-Martin and Sturman, Rob and Wiggins, Stephen},
  journal={Small},
  volume={3},
  number={2},
  pages={202--218},
  year={2007}
  }

@inproceedings{le2022st,
  title={Sobolev Transport: A Scalable Metric for Probability Measures with Graph Metrics},
  author={Le, Tam and Nguyen, Truyen and Phung, Dinh and Nguyen, Viet Anh},
  booktitle={International Conference on Artificial Intelligence and Statistics},
  pages={9844--9868},
  year={2022},
  organization={PMLR}
}

@inproceedings{rabin2011wasserstein,
  title={Wasserstein barycenter and its application to texture mixing},
  author={Rabin, Julien and Peyr{\'e}, Gabriel and Delon, Julie and Bernot, Marc},
  booktitle={International Conference on Scale Space and Variational Methods in Computer Vision},
  pages={435--446},
  year={2011},
}

@inproceedings{kusner2015word,
  title={From word embeddings to document distances},
  author={Kusner, Matt and Sun, Yu and Kolkin, Nicholas and Weinberger, Kilian},
  booktitle={International conference on machine learning},
  pages={957--966},
  year={2015},
}

@article{peyre2019computational,
  title={Computational optimal transport},
  author={Peyr{\'e}, Gabriel and Cuturi, Marco},
  journal={Foundations and Trends{\textregistered} in Machine Learning},
  volume={11},
  number={5-6},
  pages={355--607},
  year={2019},
}

@inproceedings{LYFC,
  title={Tree-sliced variants of {W}asserstein distances},
  author={Le, Tam and Yamada, Makoto and Fukumizu, Kenji and Cuturi, Marco},
  booktitle={Advances in neural information processing systems},
  pages={12283--12294},
  year={2019},
}
